\documentclass[twoside,11pt]{article}
\input epsf 

\usepackage{epsfig}
\usepackage{amsfonts}
\usepackage{amssymb}
\usepackage{jair}
\usepackage{theapa}
\usepackage{url}

\jairheading{32}{2008}{37-94}{08/07}{05/08}
\ShortHeadings{Extended RDF as a Semantic Foundation of 
Rule Markup Languages} {Analyti, Antoniou, Dam\'asio, \& Wagner}
\firstpageno{37}

\long\def\comment#1{}

\newlength{\figtblfootnotemargin}
\newlength{\figtblfootnotewidth}

\newcommand{\SquareOfSize}[2]{
 \fbox{\hsize #1cm \hbox to #1cm{\vbox{#2}}}
}

\newcommand{\puttableSV}[4]{
    \begin{table}[htbp]
        \begin{center}
            #4
            \vspace*{0.1cm}
            \caption{#2}{\vspace*{0.01cm} \centerline{ \parbox[t]{10cm}{ #3}}}
            \label{#1}
        \end{center}
        	\vspace*{-1.2cm}
    \end{table}
}

\newsavebox{\wholeWidthLine}
\sbox{\wholeWidthLine} {\rule[0.1in]{\textwidth}{.01in}}

\newcommand{\bq}{\begin{quote}}
\newcommand{\eq}{\end{quote}}
\newcommand{\be}{\begin{enumerate}}
\newcommand{\ee}{\end{enumerate}}
\newcommand{\bi}{\begin{itemize}}
\newcommand{\ei}{\end{itemize}}
\newcommand{\bie}{\begin{itemize}\begin{enumerate}}
\newcommand{\eie}{\end{enumerate}\end{itemize}}
\newcommand{\ba}{\begin{array}}
\newcommand{\ea}{\end{array}}
\newcommand{\btbl}{\begin{tabular}}
\newcommand{\etbl}{\end{tabular}}
\newcommand{\bequ}{\begin{displaymath}}
\newcommand{\eequ}{\end{displaymath}}
\newcommand{\bequa}{\begin{eqnarray*}}
\newcommand{\eequa}{\end{eqnarray*}}
\newcommand{\bc}{\begin{center}}
\newcommand{\ec}{\end{center}}
\newcommand{\btab}{\begin{tabbing}}
\newcommand{\etab}{\end{tabbing}}

\newcommand{\godown}{ \vspace*{0.3cm}}

\def\mpr#1{\ifmmode #1 \else #1 \fi}

\newcommand{\R} 		{\mpr{R}}

\newcommand{\sneg}	 { \mbox{\={$\! \! \! s$}} }

\newcommand{\DPATTRNAME}{{\large{{\bf SC}$_{attr}$}}}
\newcommand{\DPISANAME}{{\large{{\bf SC}$_{isA}$}}}
\newcommand{\DPINNAME}{{\large{{\bf SC}$_{in}$}}}

\def\arrow{\leftarrow}
\def\wneg{\mbox{$\sim$}}
\def\sneg{\neg}
\def\<{\langle}
\def\L{\langle}
\def\>{\rangle}
\def\R{\rangle}
\def\c-{\mbox{$c$-}}

\def\mImpl{\supset}
\def\h{\hat{\ }}
\def\Var{\mathit{Var}}
\def\rdf{\mathit{rdf}}
\def\rdfs{\mathit{rdfs}}
\def\erdf{\mathit{erdf}}
\def\XMLLiteral{\mathit{XMLLiteral}}
\def\rdfXMLLiteral{\mathit{rdf\?XMLLiteral}}
\def\TProp{\mathit{TProp}}
\def\TCls{\mathit{TCls}}
\def\CF{\mathit{CF}}
\def\CT{\mathit{CT}}
\def\FVar{\mathit{FVar}}
\def\PT{\mathit{PT}}
\def\PF{\mathit{PF}}
\def\LV{\mathit{LV}}
\def\IL{\mathit{IL}}

\def\ERDF{\mathit{ERDF}}
\def\formula{\mathit{formula}}
\def\ans{\mathit{ans}}
\def\Ans{\mathit{Ans}}
\def\xsd{\mathit{xsd}}

\long\def\comment#1{}
\newcommand{\I}{{\cal I}}
\newcommand{\V}{{\cal V}}
\newcommand{\C}{{\cal C}}
\newcommand{\LL}{{\cal L}}

\newcommand{\T}{{\cal T}}
\newcommand{\D}{{\cal D}}
\newcommand{\POW}{{\mathcal P}}
\newcommand{\M}{{\cal M}}
\newcommand{\?} {\mbox{:}}
\newcommand{\PL}{{\cal {PL}}}
\newcommand{\TL}{{\cal {TL}}}
\newcommand{\Lit}{{\cal LIT}}
\newcommand{\URI}{{\cal{URI}}}

\newcommand{\OR}{{\vee}}
\newcommand{\AND}{{\wedge}}

\newcommand{\NN}{\mbox{$I\!\!N$}}

\newif\iflong

\newtheorem{definition}{Definition}[section]
\newtheorem{proposition}{Proposition}[section]
\newtheorem{lemma}{Lemma}[section]
\newtheorem{example}{Example}[section]

\begin{document}

\title{Extended RDF as a Semantic Foundation of \\
Rule Markup Languages}

\author{\name Anastasia Analyti \email analyti@ics.forth.gr \\
       \addr Institute of Computer Science,  FORTH-ICS, Crete, Greece
       \AND
       \name Grigoris Antoniou \email antoniou@ics.forth.gr \\
       \addr Institute of Computer Science,  FORTH-ICS, Crete, Greece\\
 			Department of Computer Science, University of Crete, Greece
       \AND
       \name Carlos Viegas Dam\'asio \email cd@di.fct.unl.pt \\
       \addr Centro de Intelig\^encia Artificial, Universidade Nova de Lisboa, \\
       Caparica, Portugal
         \AND
       \name Gerd Wagner \email G.Wagner@tu-cottbus.de\\
       \addr Institute of Informatics, Brandenburg University \\
       of Technology at Cottbus, Germany}

\maketitle

\begin{abstract}
Ontologies and automated reasoning are the building blocks of the Semantic Web initiative. Derivation rules can be included in an ontology to define derived concepts, based on base concepts. For example, rules allow to define the extension of a class or property, based on a complex relation between the extensions of the same or other classes
and properties. On the other hand, the inclusion of negative information both in the form of negation-as-failure and explicit negative information is also needed to enable various forms of reasoning. In this paper, we extend RDF graphs with weak and strong negation, as well as derivation rules. The {\em ERDF stable model semantics} of the extended framework ({\em Extended RDF}) is defined, extending RDF(S) semantics. A distinctive feature of our theory, which is based on Partial Logic, is that both truth and falsity extensions of properties and classes are considered, allowing for truth value gaps. Our framework supports both closed-world and open-world reasoning through the explicit representation of the particular closed-world assumptions and the ERDF ontological categories of total properties and total classes.\\
\end{abstract}

\section {Introduction}

The idea of the Semantic Web is to describe the meaning of web data in a way suitable for automated reasoning. This means that descriptive data (meta-data) in machine readable form are to be stored on the web and used for reasoning.
Due to its distributed and world-wide nature, the Web creates new
problems for knowledge representation research.
Berners-Lee  \citeyear{TBL:designissues} identifies the
following fundamental theoretical problems:
negation and contradictions, open-world versus closed-world assumptions, and
rule systems for the Semantic Web.
For the time being, the first two issues have been circumvented by
discarding the facilities to introduce them, namely negation and
closed-world assumptions. Though the web ontology language OWL \cite{owloverview},
which is based on Description Logics (DLs) \cite{DLhandbook}, includes a form of classical negation through class complements,
this form is limited. This is because, to achieve decidability, classes are formed
based on specific class constructors and negation on properties is not fully considered.
Rules constitute the next layer over the ontology languages of the Semantic Web
and, in contrast to DL, allow arbitrary interaction of variables in the body of the rules.
The widely
recognized need of having rules in the Semantic
Web, demonstrated by the Rule Markup Initiative\footnote{http://www.ruleml.org/},
 has restarted the discussion of the
fundamentals of closed-world reasoning and the appropriate
mechanisms to implement it in rule systems.

The RDF(S)\footnote{RDF(S) stands for {\em Resource Description
Framework (Schema)}.} recommendation~\cite{rdfconcepts,rdfsemantics} provides
the basic constructs for defining web ontologies and a solid
ground to discuss the above issues. RDF(S) is a special predicate
logical language that is restricted to existentially quantified
conjunctions of atomic formulas, involving binary predicates only.
Due to its purpose, RDF(S) has a number of special features that
distinguish it from traditional logic languages: \vspace{-0.2cm}
\begin{enumerate}
    \item It uses a special jargon, where the things of the universe of
discourse are called \emph{resources}, types are called
\emph{classes}, and binary predicates are called
\emph{properties}. Like binary relations in set theory, properties
have a \emph{domain} and a \emph{range}. Resources are classified
with the help of the property $\rdf\?type$ (for stating that a
resource is of type $c$, where $c$ is a class).

    \item It distinguishes a special sort of resources, called \emph{literal
values}, which are denotations of lexical representations of strings, numbers, dates, or
other basic datatypes.

    \item Properties are resources, that is, properties are also
    elements of the universe of discourse. Consequently, it is possible
    to state properties of properties, i.e., make statements about
    predicates.

    \item All resources, except anonymous ones and literal values, are named with the help
of a globally unique reference schema, called {\em Uniform Resource
Identifier} (URI)\footnote{http://gbiv.com/protocols/uri/rfc/rfc3986.html}, that has been developed for the Web.

    \item RDF(S) comes with a non-standard model-theoretic semantics developed by
Pat Hayes on the basis of an idea of Christopher Menzel, which
allows self-application without violating the axiom of foundation.
An example of this is the provable sentence stating that
$\mathit{rdfs}\?\mathit{Class}$, the class of all classes, is an instance
of itself.
\end{enumerate}

However, RDF(S) does not support negation and rules. Wagner
\citeyear{wag91:2k} argues that a database, as a knowledge
representation system, needs two kinds of negation, namely {\em
weak negation} $\wneg$ (expressing negation-as-failure or
non-truth) and {\em strong negation} $\sneg$ (expressing explicit
negative information or falsity) to be able to deal with partial
information. In a subsequent paper, Wagner \citeyear{WagnerPPSWR03} makes also this point  for the
Semantic Web, as a framework for knowledge representation in
general. In the present paper, we make the same argument for the
Semantic Web language RDF and show how it can be extended to
accommodate the two negations of Partial Logic \cite{HJW96}, as
well as derivation rules. We call the new language {\em
Extended RDF} and denote it by {\em ERDF}. The model-theoretic
semantics of ERDF, called {\em ERDF stable model semantics}, is
developed based on Partial Logic \cite{HJW96}.

In Partial Logic, relating strong and weak negation at the interpretation level allows to
distinguish four  categories of properties and classes.
{\em Partial properties}  are properties $p$ that may have truth-value gaps and truth-value clashes, that
is $p(x,y)$ is possibly neither true nor false, or both true and false. {\em Total properties} are properties $p$
that satisfy $totalness$, that is $p(x,y)$ is true or false (but possibly both).
{\em Coherent properties} are properties $p$ that satisfy $coherence$, that is $p(x,y)$ cannot be both
true and false. {\em Classical properties} are total and coherent properties. For
classical properties $p$, the {\em classical logic law} applies: $p(x,y)$ is either true or false.
Partial, total, coherent, and classical classes $c$ are defined similarly, by
replacing $p(x,y)$ by $\rdf\?type(x,c)$. 

Partial logic also allows to distinguish between properties (and classes)
that are completely represented in
a knowledge base and those that are not. The classification if a
property is completely represented or not is up to the owner of
the knowledge base: the owner must know for which properties there
is complete information and for which there is not. Clearly, in
the case of a completely represented (\emph{closed}) property
$p$, entailment of $\wneg p(x,y)$ allows to derive $\sneg p(x,y)$,
and the underlying \emph{completeness assumption} has also been
called \emph{Closed-World Assumption (CWA)} in the AI literature.

Such a completeness assumption for \emph{closing} a
partial property $p$ by default may be expressed in ERDF by means
of the rule $\sneg p(?x,?y) \leftarrow \wneg p(?x,?y)$ and for a
partial class $c$, by means of the rule $\sneg \rdf\?type(?x,c) \leftarrow
\wneg \rdf\?type(?x,c)$. These derivation rules are called {\em default closure rules}.
In the case of a total property $p$, default closure rules are not applicable. 
This is because, some of the considered interpretations will satisfy 
$p(x,y)$ and the rest $\sneg p(x,y)$\footnote{
On total properties $p$, the {\em Law of Excluded Middle}  $p(x,y) \OR \sneg p(x,y)$ applies.},
preventing the preferential entailment of $\wneg p(x,y)$. Thus, on total properties,
an \emph{Open-World Assumption (OWA)} applies.
Similarly to first-order-logic, in order to infer 
negated statements about total properties,
explicit negative information has to be supplied,
along with ordinary (positive) information.

As an example, consider an ERDF knowledge base $\mathit{KB}$ that contains the facts:
{
\begin{center}
$\mathit{interestedIn(Anastasia, SemanticWeb)}$ $\;\;\;\;$ 
$\mathit{interestedIn(Grigoris, Robotics)}$ 
\end{center}
}
\noindent
indicating that  $\mathit{Anastasia}$ is interested in the $\mathit{Semantic Web}$ area and
$\mathit{Grigoris}$ is interested in the $\mathit{Robotics}$ area.
Then, the statement $\mathit{interestedIn(Anastasia}$, $\mathit{Robotics)}$ is not satisfied in the single intended
model of $\mathit{KB}$. Thus, $\mathit{KB}$ entails $\wneg \mathit{interestedIn(Anastasia, Robotics)}$.

Assume now that the previous list of areas of interest is not complete for
$\mathit{Anastasia}$ or $\mathit{Grigoris}$. Then, we should add to knowledge base $\mathit{KB}$ the statement:
\begin{center}
$\rdf\?\mathit{type}(\mathit{interestedIn}$, $\mathit{\erdf\?TotalProperty})$ 
\end{center}
\noindent
indicating that
$\mathit{interestedIn}$ is a total property.
In this case, an open-world assumption is made for 
$\mathit{interestedIn}$ and
$\mathit{KB}$ does not 
entail $\wneg \mathit{interestedIn(Anastasia, Robotics)}$, any longer. In particular,
 there is an intended model of the revised $\mathit{KB}$ that satisfies 
$\mathit{interestedIn(Anastasia,} $ $\mathit{Robotics)}$.
Of course, if it is known that $\mathit{Anastasia}$ is not interested in $\mathit{Robotics}$ then
$\sneg \mathit{interestedIn(Anastasia,} $ $\mathit{Robotics)}$ should be added to $\mathit{KB}$.

Assume now that we add to $\mathit{KB}$ the following facts:
{
\begin{center}
$\mathit{hasCar(Anastasia, Suzuki)}$ $\;\;\;\;$ 
$\mathit{hasCar(Grigoris, Volvo)}$ 
\end{center}
}
\noindent
and assume that $\mathit{KB}$ has complete knowledge
on the property $\mathit{hasCar}$, as far as it concerns elements in the Herbrand Universe of  $\mathit{KB}$.
Then,
the default closure rule
$\sneg \mathit{hasCar(?x, ?y)} \leftarrow \wneg \mathit{hasCar(?x, ?y)}$
can be safely added to $\mathit{KB}$. As a result, $\sneg \mathit{hasCar(Anastasia, Volvo)}$
is satisfied in all intended models of  $\mathit{KB}$.
Thus, $\mathit{KB}$ entails $\sneg \mathit{hasCar(Anastasia, Volvo)}$.

The previous example shows  the need for
supporting both closed-world and open-world
reasoning in the same framework.
Damasio et al. \citeyear{DAAW_PPSWR06} and Analyti et al. \citeyear{AADW04} provide further examples and arguments for this need.
Unfortunately, classical logic and thus also OWL support only open-world reasoning.

Specifically, in this paper:

\begin{enumerate}
\item We extend RDF graphs to ERDF graphs with the inclusion of strong negation, and then to
ERDF ontologies (or ERDF knowledge bases) with the inclusion of
general derivation rules. ERDF graphs allow to express existential positive and negative information, whereas
general derivation rules allow inferences based on formulas built using the connectives $\wneg$, $\sneg$, $\mImpl$, $\AND$, $\OR$
and the quantifiers $\forall$,
$\exists$.

\item We extend the vocabulary of RDF(S) with the terms $\erdf\?\mathit{\mathit{TotalProperty}}$ and \\
$\erdf\?\mathit{TotalClass}$, representing the metaclasses of total properties and total classes, on which the open-world assumption applies.

\item We extend RDFS interpretations to ERDF interpretations including both truth and falsity extensions for
properties and classes. Particularly, we consider only {\em coherent} ERDF interpretations (imposing coherence on all properties).
Thus, in this paper, total properties and classes become synonymous to classical properties and classes.

\item We extend RDF graphs to ERDF formulas that are built from positive triples, using the connectives
$\wneg$, $\sneg$, $\mImpl$, $\AND$, $\OR$ and the quantifiers $\forall$, $\exists$. Then, we define ERDF 
entailment between two ERDF formulas, extending RDFS entailment between RDF graphs.

\item We define the ERDF models, the Herbrand interpretations, and the minimal Herbrand models of an
ERDF ontology. Since not all minimal Herbrand models of an
ERDF ontology are intended, we define the {\em stable models} of an
ERDF ontology. The definition of a stable model is based on the intuition that:
\begin{enumerate}
\item assertions stating that a property $p$ or class $c$ is total
should only be accepted, if the ontology contains some direct support for them in the form of an acceptable rule sequence, and

\item assertions $[\sneg]p(s,o)$ and $[\sneg]\rdf\?type(o,c)$ should only be accepted, if (i) the ontology 
contains some direct support for them in the form of an acceptable rule sequence, or (ii) property $p$ and class $c$ are total, respectively.

\end{enumerate}

\item We show that stable model entailment on ERDF ontologies extends ERDF entailment on ERDF 
graphs, and thus it also extends RDFS entailment on RDF graphs.
Moreover, we show that if all properties are total, (boolean) Herbrand model reasoning and stable model 
reasoning coincide. In this case, we make an open-world assumption for all properties and classes.

\end{enumerate}

A distinctive feature of the developed framework with respect to Partial Logic \cite{HJW96} is that properties and classes are declared as total on a selective basis, by extending RDF(S)
with new built-in classes and providing support for the respective ontological categories. In contrast, in Partial Logic \cite{HJW96}, the choice of partial or total should be taken
for the complete set of predicates.
Thus, the approach presented here is, in this respect, more flexible and general.

This work extends our conference paper \cite{AADW-ISWC05} by (i) considering the full RDFS model, (ii) providing a detailed characterization
of the properties of ERDF interpretations/models, Herbrand interpretations/models, and finally ERDF stable models, 
(iii) discussing decidability issues, and (iv) providing formal proofs of all lemmas
and propositions.

The rest of the paper is organized as follows:
In Section \ref{sec:ERDF_graphs}, we extend RDF graphs to ERDF graphs and ERDF formulas.
Section \ref{sec:ERDF_interpretations} defines ERDF interpretations and ERDF entailment.
We show that ERDF entailment extends RDFS entailment.
In Section \ref{sec:ERDF_ontologies}, we define ERDF ontologies and the Herbrand models of an ERDF ontology.
In Section \ref{sec:ERDF_stableModel}, we define the stable models of an ERDF ontology. Section \ref{sec:StableEntailment}
defines stable model entailment, showing that
it extends ERDF entailment.  In Section \ref{sec:ERDF/XMLsyntax}, we provide a brief sketch of the ERDF/XML  syntax. Decidability  issues for the ERDF stable model semantics are discussed in Section \ref{sec:Complexity}.
Section \ref{sec:ERDF-Tarski} shows that the developed ERDF model theory can be seen as a Tarski-style model theory.
Section \ref{sec:RelatedWork} reviews related work and Section
\ref{sec:Conclusions} concludes the paper, including future work. The main definitions of RDF(S) semantics are reviewed
in Appendix A. Appendix B includes the proofs of the lemmas and propositions, presented in the paper.

\section{Extending RDF Graphs with Negative Information}
\label{sec:ERDF_graphs}

 In this section, we extend RDF graphs to ERDF graphs, by adding strong negation. Moreover, we extend RDF graphs
 to ERDF formulas, which are built from positive ERDF triples, the connectives $\wneg$, $\sneg$,
 $\mImpl$, $\AND$, $\OR$, and the quantifiers $\forall$,
$\exists$.

According to RDF concepts \cite{rdfconcepts,rdfsemantics},
\emph{URI references} are used as globally unique names for web
resources. An RDF URI reference is a Unicode string that
represents an absolute URI (with an optional fragment identifier). It
may be represented as a \emph{qualified name}, that is a
colon-separated two-part string consisting of a \emph{namespace
prefix} (an abbreviated name for a namespace URI) and a local
name. For example, given the namespace prefix ``ex" defined to
stand for the namespace URI ``http://www.example.org/", the
qualified name ``ex:Riesling" (which stands for
``http://www.example.org/Riesling") is a URI reference.

A plain literal is a string ``$s$", where $s$ is a sequence of
Unicode characters, or a pair of a string ``$s$" and a language
tag $t$, denoted by ``$s$"@$t$. A typed literal is a pair of a
string ``$s$" and a datatype URI reference $d$, denoted by
``$s$"$\h\h d$. For example, ``$27$"$\h\h xsd\?integer$ is a typed
literal.

A (Web) {\em vocabulary} $V$ is a set of URI references and/or
literals (plain or typed). We denote the set of all URI references
by $\URI$, the set of all plain literals by $\PL$, the set of all
typed literals by $\TL$, and the set of all literals by $\Lit$.
It holds: $\URI \cap \Lit = \emptyset$.

In our formalization, we consider a set $\Var$ of variable symbols, such that the sets $\Var$, $\URI$, $\Lit$
are pairwise disjoint. In the main text, variable symbols are explicitly indicated, while in our examples,
variable symbols are prefixed by a question mark symbol ``?".

An RDF triple \cite{rdfconcepts,rdfsemantics} is a triple ``{\em s p o.}", where
$s \in \URI \cup \Var$, $\; p \in \URI$, and $o \in \URI \cup \Lit \cup \Var$,
expressing that the subject $s$ is related with the object $o$ through the property $p$.
An RDF graph is a set of RDF triples. The variable symbols appearing in an RDF graph are called {\em blank nodes},
and are, intuitively, {\em existentially quantified variables}.
In this paper, we denote an RDF triple ``{\em s p o.}" by $p(s,o)$.
Below we extend the notion of RDF triple to allow for both
positive and negative information.

\begin{definition}[ERDF triple]
{\em Let $V$ be a vocabulary. A {\em positive ERDF triple} over
$V$ (also called \emph{ERDF sentence atom}) is an expression of
the form $p(s,o)$, where $s, o \in V \cup \Var$ are called
\emph{subject}\footnote{Opposed to ``pure" RDF \cite{rdfconcepts}, 
we allow literals in the subject position of an ERDF triple.} and \emph{object}, respectively, and
$p \in V\cap \URI$ is called \emph{predicate} or \emph{property}.
A {\em negative ERDF triple} over $V$ is the strong negation $\neg
p(s,o)$ of a positive ERDF triple $p(s,o)$ over $V$. 
An {\em ERDF triple} over $V$ (also called \emph{ERDF sentence
literal}) is a positive or negative ERDF triple over $V$. $\Box$}
\end{definition}

For example, $ex\?likes(ex\?\mathit{Gerd}, ex\?\mathit{Riesling})$ is a positive ERDF triple, expressing that
$\mathit{Gerd}$ likes $\mathit{Riesling}$, and
$\sneg ex\?likes(ex\?\mathit{Carlos}, ex\?\mathit{Riesling})$ is a negative ERDF triple, expressing that
$\mathit{Carlos}$ dislikes $\mathit{Riesling}$.
Note that an RDF triple is a positive ERDF triple with
the constraint that the subject of the triple is not a literal.
For example, $ex\?\mathit{denotationOf}(``\mathit{Grigoris}", ex\?\mathit{Grigoris})$ is a valid
ERDF triple but not a valid RDF triple. Our choice of allowing literals appearing
in the subject position is based on our intuition that this case can naturally appear
in knowledge representation (as in the previous
example). Prud'hommeaux \& Seaborne \citeyear{SPARQL} and de Bruijn et al. 
\citeyear{bruijn05} also consider literals in the subject position of RDF triples.

\godown

Based on the notion of ERDF triple, we define ERDF graphs and
ERDF formulas, as follows:

\begin{definition}[ERDF graph]
{\em An {\em ERDF graph} $G$ is a set of ERDF triples over some
vocabulary $V$. We denote the variables appearing in $G$ by
$\Var(G)$, and the set of URI references and literals appearing in
$G$ by $V_G$. $\Box$}
\end{definition}

Note that as an RDF graph is a set of RDF triples \cite{rdfconcepts,rdfsemantics}, an RDF graph is
also an ERDF graph.

\begin{definition}[ERDF formula]
{\em Let $V$ be a vocabulary. We consider the logical factors
$\{\wneg, \sneg, $ $\AND, \OR, \mImpl, \exists, \forall \}$, where
$\neg$, $\sim$, and $\mImpl$ are called {\em strong negation},
{\em weak negation}, and {\em material implication}, respectively.
We denote by $L(V)$ the smallest set that contains the positive
ERDF triples over $V$ and is closed with respect to the following
conditions: if $F,G \in L(V)$ then $\{\wneg F, \; \sneg F, \; F \AND
G, \; F \OR G, \; $
$F \mImpl G, \; \exists x F, \; \forall x F\} \subseteq L(V)$, where $x \in \Var$. An
{\em ERDF formula} over $V$ is an element of $L(V)$. We denote the
set of variables appearing in $F$ by $\Var(F)$, and the set of
free variables\footnote{Without loss of generality, we assume that
a variable cannot have both free and bound occurrences in $F$, and more than one bound occurrence.}
appearing in $F$ by $\FVar(F)$. Moreover, we denote the set of URI references and literals appearing in
$F$ by $V_F$. $\Box$}
\end{definition}

For example, let:

\begin{center}
$F= \forall ?x \; \exists ?y \; (\rdf\?type(?x, ex\?\mathit{Person}) \mImpl  ex\?\mathit{hasChild}(?y,?x))\; \AND \; $ $\rdf\?type(?z, ex\?\mathit{Person})$ 
\end{center}

 Then, $F$ is an ERDF formula over the
vocabulary $V=\{\rdf\?type,$ $ex\?\mathit{Person},$ $ex\?\mathit{hasChild}\}$ with
$\Var(F)=\{?x,?y,?z\}$ and 
$\FVar(F)=\{?z\}.$

We will denote the sublanguages of $L(V)$ formed by means of a subset
$S$ of the logical factors, by $L(V|S)$. For example, $L(V|\{\sneg\})$ denotes the set of
(positive and negative) ERDF triples over $V$.

\section{ERDF Interpretations} \label{sec:ERDF_interpretations}

In this section, we extend RDF(S) semantics by allowing for
partial properties and classes. In particular, we define ERDF
interpretations and satisfaction of an ERDF formula, based on the notion of
{\em partial interpretation}.

\subsection{Partial Interpretations}

We define a partial interpretation as an extension of a
simple interpretation \cite{rdfsemantics}, where each property is
associated not only with a truth extension but also with a falsity
extension allowing for partial properties. 
The notation $\POW(S)$, where $S$ is a set, denotes the {\em powerset} of $S$.

\begin{definition}  [Partial interpretation] \label{def:PartialInterpretation}
{\em A {\em partial interpretation} $I$ of a vocabulary $V$
consists of:

{\small
 \begin{itemize}
\item A non-empty set of resources $Res_I$, called the {\em domain} or {\em universe} of $I$.

\item A set of properties $Prop_I$.

\item A vocabulary interpretation mapping $I_V$\footnote{In the symbol $I_V$, $V$ stands for {\em Vocabulary}.}$:V \cap \URI \rightarrow
Res_I \cup Prop_I$.

\item A property-truth extension mapping $\PT_I: Prop_I \rightarrow
\POW(Res_I \times Res_I)$.

\item A property-falsity extension mapping
$\PF_I: Prop_I \rightarrow \POW(Res_I \times Res_I)$.

\item A mapping $\IL_I: V \cap \TL \rightarrow Res_I$.

\item A set of literal values $\LV_I \subseteq Res_I$, which contains
$V \cap \PL$.
\end{itemize}

}
\noindent  We define the mapping: $\;I: V \rightarrow Res_I \cup Prop_I$, called {\em denotation}, such that:

{\small
 \begin{itemize}
 \item $I(x)=I_V(x)$,  $\; \forall x \in V \cap \URI$.
\item $I(x)=x$, $\; \forall \; x \in V \cap \PL.$

 \item $I(x)=\IL_I(x)$, $\; \forall \; x \in V \cap \TL.$ $\Box$
 \end{itemize}
 }
 
 }
\end{definition}

Note that the truth and falsity  extensions of a property $p$
according to a partial interpretation $I$, 
that is $PT_I(p)$ and $PF_I(p)$, are sets
of pairs $\<subject, object\>$ of resources.
As an example, let:
{\small \begin{center}
 $V=\{ex\?\mathit{Carlos},$ $ex\?\mathit{Grigoris},$  
$ex\?\mathit{Riesling},$ $ex\?likes$, $ex\?\mathit{denotationOf}$, $``\mathit{Grigoris}$"$\h\h xsd\?string \}$ 
\end{center}
}
and consider a structure $I$ that consists of:
 
 {\small
 \begin{itemize}
\item A set of resources $Res_I=\{C, G, R, l, d, ``\mathit{Grigoris}"\}$.

\item A set of properties $Prop_I=\{l, d\}$.

\item A vocabulary interpretation mapping $I_V :V \cap \URI \rightarrow
Res_I \cup Prop_I$ such that:\\
$I_V(ex\?\mathit{Carlos})=C$, $I_V(ex\?\mathit{Grigoris})=G$, $I_V(ex\?\mathit{Riesling})=R$,
$I_V(ex\?likes)=l$, and \\
$I_V(ex\?\mathit{denotationOf})=d$.

\item A property-truth extension mapping $\PT_I: Prop_I \rightarrow
\POW(Res_I \times Res_I)$ such that: \\
$PT_I(d)=\{\<``\mathit{Grigoris}",G\>\}$.

\item A property-falsity extension mapping
$\PF_I: Prop_I \rightarrow \POW(Res_I \times Res_I)$ such that: \\
$PF_I(l)=\{\<C,R\>\}$.

\item A mapping $\IL_I: V \cap \TL \rightarrow Res_I$ such that: 
$IL_I(``\mathit{Grigoris}$"$\h\h \xsd\?string)= ``\mathit{Grigoris}"$.

\item A set of literal values $\LV_I =\{``\mathit{Grigoris}"\}$.
\end{itemize}
}

It is easy to see that $I$ is a partial interpretation of $V$,
expressing that: (i) ``$\mathit{Grigoris}$" is the denotation of $\mathit{Grigoris}$
and (ii) $\mathit{Carlos}$ dislikes $\mathit{Riesling}$.

\begin{definition} [Coherent partial interpretation] \label{def:CoherentERDFinterpretation}
 {\em A partial interpretation $I$ of a
vocabulary $V$ is $coherent$ iff
for all $x \in Prop_I, \;$ $\PT_I(x) \cap \PF_I(x)=\emptyset$. $\Box$}

\end{definition}

Coherent partial interpretations enforce the constraint that a pair of
resources cannot belong to both the truth and falsity extensions of a property
(i.e., all properties are coherent).
Intuitively, this means that an ERDF triple cannot be both true and false.

Continuing our previous example, note that $I$ is a coherent partial interpretation.
Consider now a partial interpretation $J$ which is exactly as $I$, except that
it also holds:  $PT_J(l)=\{\<C,R\>\}$ (expressing that $\mathit{Carlos}$ likes $\mathit{Riesling}$).
Then, $\<C,R\>$ belongs to both the truth and falsity extension of $l$ 
(i.e., $\<C, R\> \in PT_J(l) \cap PF_J(l)$). Thus, $J$ is not coherent.

\godown
To define satisfaction of an ERDF formula w.r.t. a partial interpretation,
we need first the following auxiliary definition.
 
\begin{definition} [Composition of a partial interpretation and a valuation]
\label{def:Valuation}
{\em Let $I$ be a partial interpretation of a vocabulary $V$ and let $v$ be a partial function
$v:\Var \rightarrow Res_I$ (called $valuation$). We define: (i) $[I+v](x)=v(x)$, if $x \in  \Var$, and (ii)
 $[I+v](x)=I(x)$, if $x \in V$. $\Box$
}
\end{definition}

\begin{definition} {\bf (Satisfaction of an ERDF formula w.r.t. a partial interpretation and a valuation)}
\label{def:satisfiesValuation}
{\em Let $F,G$ be ERDF formulas and let $I$ be a partial
interpretation of a vocabulary $V$. Additionally,
let $v$ be a mapping $v:\Var(F) \rightarrow Res_I$.

{\small
\begin{itemize}

\item If $F= p(s,o)$ then $I,v \models F$ iff
 $p \in V \cap \URI, \; s,o \in V \cup \Var, \; I(p) \in Prop_I$, and\\
$\<[I+v](s),[I+v](o)\> \in \PT_I(I(p))$.

\item If $F= \sneg p(s,o)$ then $\; I,v \models F$ iff
 $p \in V \cap \URI, \; s,o \in V \cup \Var, \; I(p) \in Prop_I$, and\\
$\<[I+v](s),[I+v](o)\> \in \PF_I(I(p))$.

\item If $F= \wneg G$ then $\; I,v \models F$ iff $V_G \subseteq V$ and $I,v \not \models G$.

\item If $F= F_1 \AND F_2$ then $\; I,v \models F$ iff $I,v
\models F_1$ and $I,v \models F_2$.

\item If $F= F_1 \OR F_2$ then $\; I,v \models F$ iff $I,v \models
F_1$ or $I,v \models F_2$.

\item If $F= F_1 \mImpl F_2$ then\footnote{{\em Material implication} is the logical relationship between 
any two ERDF formulas such that either the first is {\em non-true} or the second is {\em true}.}
 $\; I,v \models F$ iff $I,v \models \wneg F_1 \OR F_2$.

 \item If $F= \exists x \; G$
then $\; I,v \models F$ iff there exists mapping $u:\Var(G)
\rightarrow Res_I$ such that $u(y)=v(y)$, $\forall y \in
\Var(G)-\{x\}$, and $I,u \models G$.

\item If $F= \forall x \; G$ then $\; I,v \models F$ iff for all
mappings $u:\Var(G) \rightarrow Res_I$ such that $u(y)=v(y)$,
$\forall y \in \Var(G)-\{x\}$, it holds $I,u \models G$.

\item All other cases of ERDF formulas are treated by the
following DeMorgan-style rewrite rules expressing the
falsification of compound ERDF formulas:\\
   $ \sneg (F \wedge G) \rightarrow \sneg F \vee \sneg G, \;\; \sneg (F \vee G) \rightarrow \sneg F \wedge \sneg G,
    \;\; \sneg (\neg F) \rightarrow F, \;\; \sneg (\sim F) \rightarrow F$\footnote{This transformation
    expresses that if it is {\em false}  that $F$ {\em does not hold} then
    $F$ {\em holds}.},\\
    $\sneg (\exists x \; F) \rightarrow \forall x \; \sneg F, \;\;
    \sneg (\forall x \; F) \rightarrow \exists x \; \sneg F,\;\; \sneg (F \mImpl G) \rightarrow F \AND \sneg G.$ $\Box$
\end{itemize}}
}
\end{definition}

Continuing our previous example, let $v: \{?x,?y,?z\} \rightarrow Res_I$
such that $v(?x)=C$, $v(?y)=R$, and $v(?z)=G$. It holds: 
\begin{center}
$I,v \models \sneg ex\?likes(?x,?y) \; \AND \;
ex\?\mathit{denotationOf}(``\mathit{Grigoris}$"$\h\h xsd\?string, \;?z)$.
\end{center}

\begin{definition} [Satisfaction of an ERDF formula w.r.t. a partial interpretation]
\label{def:modelRelation} {\em Let $F$ be an ERDF formula and
let $I$ be a partial interpretation of a vocabulary $V$. We say
that $I$ {\em satisfies} $F$, denoted by
 $I \models F$, iff for every mapping $v:\Var(F)
\rightarrow Res_I$, it holds $I,v \models F.$ $\Box$ }
\end{definition}

Continuing our previous example, $I \models \exists ?x \; \sneg ex\?likes(ex\?\mathit{Carlos},?x)$.

Below we define ERDF graph satisfaction, extending satisfaction of an RDF graph  \cite{rdfsemantics} (see also Appendix A).
\begin{definition} [Satisfaction of an ERDF graph w.r.t. a partial interpretation]
\label{def:ERDF_Graph_modelRelation} {\em Let $G$ be an ERDF graph and
let $I$ be a partial interpretation of a vocabulary $V$. Let $v$ be a mapping $v: \Var(G) \rightarrow Res_I$.
We define:
\begin{itemize}
\item $I, v \models_{\tt GRAPH} G$ iff $\forall t \in G$, $\;\; I,v \models t$.
\item $I$ {\em satisfies} the ERDF graph $G$, denoted by $I \models_{\tt GRAPH} G$, iff there exists a mapping
$v: \Var(G) \rightarrow Res_I$ such that $I, v \models_{\tt GRAPH} G$. $\Box$
\end{itemize}}
\end{definition}

 Intuitively, an ERDF graph $G$ represents an existentially
quantified conjunction of ERDF triples. Specifically, let
$G=\{t_1, ..., t_n\}$ be an ERDF graph, and let
$\Var(G)=\{x_1, ..., x_k\}$. Then, $G$ represents the ERDF formula $\formula(G)=\exists
?x_1, ..., \exists ?x_k \; t_1 \; \AND \; ... \; \AND \; t_n$. This is shown in the following lemma.

\begin{lemma} \label{lem:ERDFInterpretation_ERDF_graph}
{\em Let $G$ be an ERDF graph and let $I$ be a partial interpretation of a vocabulary $V$.
It holds:
$I \models_{\tt GRAPH} G$ iff $I \models \formula(G)$.}
\end{lemma}

Following the RDF terminology \cite{rdfconcepts},
the variables of an ERDF graph are also called {\em blank nodes} and
intuitively denote anonymous web resources.
For example, consider the ERDF graph: 
\begin{center}
$G=\{\rdf\?type(?x, ex\?EuropeanCountry),$
$\sneg \rdf\?type(?x, ex\?EUmember)\}$. 
\end{center}
Then, $G$ represents the ERDF formula $\formula(G)=$ 
\begin{center}
$\exists ?x \; (\rdf\?type(?x, ex\?EuropeanCountry)\; \AND \; \sneg \rdf\?type(?x, ex\?EUmember))$, 
\end{center}
expressing that there is a European country which is not a European Union member.

\godown 
\noindent
{\bf Notational Convention:} {\em Let $G$ be an ERDF graph,
let $I$ be a partial interpretation of a vocabulary $V$, and let $v$ be a mapping $v:\Var(G) \rightarrow Res_I$.
Due to Lemma \ref{lem:ERDFInterpretation_ERDF_graph}, we will write (by an abuse of notation) ``$I,v \models G$" and ``$I \models G$"
instead of  ``$I,v \models_{\tt GRAPH} G$" and ``$I \models_{\tt GRAPH} G$", respectively.}

\subsection{ERDF Interpretations and Entailment}

In this subsection, we define ERDF interpretations and entailment as
an extension of  RDFS interpretations and entailment
\cite{rdfsemantics}. First, we define the vocabularies of RDF, RDFS, and ERDF.

The vocabulary of RDF, $\V_{RDF}$, is a set of $\URI$ references in
the $\rdf\?$ namespace \cite{rdfsemantics}, as shown in Table
\ref{table:Vocabulary_RDF_RDFS}. The vocabulary of RDFS,
$\V_{RDFS}$, is a set of $\URI$ references in the $\mathit{rdfs}\?$
namespace \cite{rdfsemantics}, as shown in Table
\ref{table:Vocabulary_RDF_RDFS}. The {\em vocabulary of} $\mathit{ERDF}$,  $\V_{ERDF}$, is a set of $\URI$
references in the $\erdf\?$ namespace. Specifically, the set of
ERDF  predefined classes is
 $\C_{ERDF}=$
 $\{\erdf\?\mathit{TotalClass},$
 $\erdf\?\mathit{TotalProperty}\}$.
We define $\V_{ERDF}=\C_{ERDF}$.
Intuitively, instances of the metaclass $\erdf\?\mathit{TotalClass}$ are
classes $c$ that satisfy totalness, meaning that each resource belongs to the
truth or falsity extension of $c$. Similarly,
instances of the metaclass $\erdf\?\mathit{TotalProperty}$ are
properties $p$ that satisfy totalness, meaning that each pair of resources belongs to
the truth or falsity extension of $p$.

{\small
\puttableSV{table:Vocabulary_RDF_RDFS}{The vocabulary of RDF and
RDFS}{}{\small { \btbl{|l|l|} \hline $V_{RDF}$ &   $V_{RDFS}$
\\\hline\hline $\rdf\?type$ & $\mathit{rdfs}\?\mathit{domain}$ \\\hline
$\rdf\?\mathit{Property}$ & $\mathit{rdfs}\?\mathit{range}$ \\\hline
$\rdfXMLLiteral$ & $\mathit{rdfs}\?\mathit{Resource}$ \\\hline $\rdf\?nil$
& $\mathit{rdfs}\?\mathit{Literal}$
\\\hline $\rdf\?\mathit{List}$ & $\mathit{rdfs}\?\mathit{Datatype}$ \\\hline $\rdf\?\mathit{Statement}$
& $\mathit{rdfs}\?\mathit{Class}$ \\\hline $\rdf\?\mathit{subject}$ &
$\mathit{rdfs}\?\mathit{subClassOf}$
\\\hline $\rdf\?predicate$ & $\mathit{rdfs}\?\mathit{subPropertyOf}$ \\\hline
$\rdf\?\mathit{object}$ & $\mathit{rdfs}\?\mathit{member}$ \\\hline
 $\rdf\?\mathit{first}$ & $\mathit{rdfs}\?\mathit{Container}$ \\\hline
 $\rdf\?\mathit{rest}$ & $\mathit{rdfs}\?\mathit{ContainerMembershipProperty}$ \\\hline
 $\rdf\?Seq$ & $\mathit{rdfs}\?\mathit{comment}$ \\\hline
 $\rdf\?Bag$ & $\mathit{rdfs}\?\mathit{seeAlso}$ \\\hline
 $\rdf\?Alt$ & $\mathit{rdfs}\?\mathit{isDefinedBy}$ \\\hline
 $\rdf\?\_i$, $\;\;\forall i \in \{1,2,...\}$ & $\mathit{rdfs}\?\mathit{label}$ \\\hline
  $\rdf\?\mathit{value}$ &  \\\hline
 \etbl }}

}

\godown
We are now ready to define an ERDF interpretation over a
vocabulary $V$ as an extension of an RDFS interpretation
\cite{rdfsemantics} (see also Appendix A), where each property and class is associated
not only with a truth extension but also with a falsity extension,
allowing for both partial properties and partial classes. Additionally, an
ERDF interpretation gives special semantics to terms from the ERDF vocabulary.

\begin{definition} [ERDF interpretation] \label{def:ERDFInterpretation}
 {\em An {\em ERDF interpretation} $I$ of a
vocabulary $V$ is a partial interpretation of $V \cup \V_{RDF}
\cup \V_{RDFS} \cup \V_{ERDF}$, extended by the new ontological
categories $Cls_I \subseteq Res_I$ for classes, $\TCls_I \subseteq Cls_I$
for total classes, and $\TProp_I \subseteq Prop_I$ for total
properties, as well as the class-truth extension mapping
$\CT_I:Cls_I \rightarrow \POW(Res_I)$, and the class-falsity extension
mapping $\CF_I:Cls_I \rightarrow \POW(Res_I)$, such that:
\begin{small}
\vspace{-0.1cm}
\begin{enumerate}

\item $x \in \CT_I(y) $ iff $\<x,y\> \in \PT_I(I(\rdf\?type))$, and\\
 $x \in \CF_I(y)$ iff $\<x,y\> \in \PF_I(I(\rdf\?type))$.

\item The ontological categories are defined as follows:

{ \begin{tabular}{ll} $Prop_I=\CT_I(I(\rdf\?\mathit{Property}))$ &
$\;\;Cls_I=\CT_I(I(\mathit{rdfs}\?\mathit{Class}))$\\
    $Res_I=\CT_I(I(\mathit{rdfs}\?\mathit{Resource}))$ &
    $\;\; \LV_I=\CT_I(I(\mathit{rdfs}\?Literal))$\\
    $\TCls_I=\CT_I(I(\erdf\?\mathit{TotalClass}))$ &
    $\;\; \TProp_I=\CT_I(I(\erdf\?\mathit{TotalProperty}))$.
    \end{tabular}
    }\\

\item If $\<x,y\> \; \in \; \PT_I(I(\mathit{rdfs}\?domain))$ and
$\<z,w\> \; \in\; \PT_I(x)$ then $z \in \CT_I(y)$.

\item If $\<x,y\> \; \in \; \PT_I(I(\mathit{rdfs}\?range))$ and
$\<z,w\> \; \in\; \PT_I(x)$ then $w \in \CT_I(y)$.\\

\item If $x \in Cls_I$ then $\<x, I(\mathit{rdfs}\?\mathit{Resource})\> \;
\in \; \PT_I(I(\mathit{rdfs}\?\mathit{subClassOf}))$.\\

\item If $\<x,y\> \in \PT_I(I(\mathit{rdfs}\?\mathit{subClassOf}))$ then
$x,y \in Cls_I$, $\CT_I(x) \subseteq \CT_I(y)$, and \\
$\CF_I(y) \subseteq \CF_I(x)$.

\item $\PT_I(I(\mathit{rdfs}\?\mathit{subClassOf}))$ is a reflexive and
transitive relation on $Cls_I$.\\

\item If $\<x,y\> \in \PT_I(I(\mathit{rdfs}\?\mathit{subPropertyOf}))$ then
$x,y \in Prop_I$, $\PT_I(x) \subseteq \PT_I(y)$, and \\
$\PF_I(y) \subseteq \PF_I(x)$.

\item $\PT_I(I(\mathit{rdfs}\?\mathit{subPropertyOf}))$ is a reflexive and
transitive relation on $Prop_I$.\\

\item If $x \in \CT_I(I(\mathit{rdfs}\?Datatype))$ then $\<x,
I(\mathit{rdfs}\?Literal)\> \; \in \;
\PT_I(I(\mathit{rdfs}\?\mathit{subClassOf}))$.

\item If $x \in
\CT_I(I(\mathit{rdfs}\?\mathit{ContainerMembershipProperty}))$ then \\
$\<x, I(\mathit{rdfs}\?member)\> \in
\PT_I(I(\mathit{rdfs}\?\mathit{subPropertyOf}))$.\\

\item If $x \in \TCls_I$ then $\CT_I(x) \cup \CF_I(x)=Res_I$.

\item If $x \in \TProp_I$ then $\PT_I(x) \cup \PF_I(x)=Res_I \times Res_I$.\\

 \item If $``s"\h\h \rdfXMLLiteral \in V$  and $s$ is a
well-typed XML literal string, then\\
 $\IL_I$($``s"\h\h \rdfXMLLiteral$) is the XML value of $s$, and\\
$\IL_I(``s"\h\h \rdfXMLLiteral) \in \CT_I(I(\rdfXMLLiteral))$.

 \item If $``s"\h\h \rdfXMLLiteral \in V$  and $s$ is an
ill-typed XML literal string then\\
 $\IL_I(``s"\h\h \rdfXMLLiteral)\in Res_I-\LV_I$, and\\
 $\IL_I(``s"\h\h \rdfXMLLiteral) \in \CF_I(I(\mathit{rdfs}\?Literal))$.\\

 \item $I$
satisfies the RDF and RDFS axiomatic triples
\cite{rdfsemantics}, shown in Table \ref{table:RDFaxiomaticTriples} and Table \ref{table:RDFSaxiomaticTriples} of  Appendix A,
respectively.

\item $I$
satisfies the following triples, called {\em ERDF axiomatic triples}:\\
$\mathit{rdfs}\?\mathit{subClassOf}(\erdf\?\mathit{TotalClass},
\mathit{rdfs}\?\mathit{Class})$. \\
$\mathit{rdfs}\?\mathit{subClassOf}(\erdf\?\mathit{TotalProperty}, \mathit{rdfs}\?\mathit{Class})$.

\end{enumerate}
\end{small}}
\end{definition}

Note that while RDFS intepretations \cite{rdfsemantics} imply a two-valued interpretation 
 of the instances of  $\mathit{rdf\?Property}$, this is no longer the case with 
ERDF interpretations. Specifically, let $I$ be an ERDF interpretation,  let $\mathit{p \in CT_I(I(rdf\?Property))}$, and 
let $\<x,y\> \in Res_I \times Res_I$. It  may be the case that neither  $\<x,y\> \in PT_I(p)$ nor $\<x,y\> \in 
 PF_I(p)$. That is $p(x,y)$ is neither true nor false. 

Semantic conditions of ERDF interpretations may
impose constraints to both the truth and falsity extensions of
properties and classes. Specifically, consider semantic condition 6
of Definition \ref{def:ERDFInterpretation} and assume that
$\<x,y\> \in \PT_I(I(\mathit{rdfs}\?\mathit{subClassOf}))$. Then, $I$ should not only
satisfy $\CT_I(x) \subseteq \CT_I(y)$ (as an RDFS interpretation $I$ does), but also $\CF_I(y)
\subseteq \CF_I(x)$. 
The latter is true because if it is certain that a resource $z$ does not belong to the
truth extension of class $y$ then it is certain that $z$ does not belong to the
truth extension of class $x$. Thus, the falsity extension of $y$ is contained in the
falsity extension of $x$.
Similar is the case for semantic condition 8.
Semantic conditions 12 and 13 represent our definition of total classes and total properties, 
respectively.
Semantic condition 15 expresses that the denotation of 
an ill-typed XML literal is not a literal value. Therefore (see semantic condition 2), it is 
certain that it is not contained in the truth extension of the class $\mathit{rdfs}\?Literal$. Thus, it is
contained in the falsity extension of the class $\mathit{rdfs}\?Literal$.

Let $I$ be a coherent ERDF interpretation of a vocabulary $V$. Since $I(\rdf\?type) $ $\in Prop_I$,
it holds:
$\forall x \in Cls_I$, $\; \CT_I(x)\cap \CF_I(x)=\emptyset$. 
Thus, all properties and classes of coherent ERDF interpretations are coherent.

\godown

\noindent{\bf Convention:} {\em In the rest of the document, we consider only coherent ERDF interpretations.
This means that referring to an ``ERDF interpretation", we implicitly mean a ``coherent" one.
Moreover, to improve the readability of our examples, we will ignore the example namespace $ex\?$.}

\godown
According to RDFS semantics \cite{rdfsemantics}, the only source
of RDFS-inconsistency is the appearance of an ill-typed XML
literal $l$ in the RDF graph, in combination with the derivation
of the RDF triple ``$x$ $\rdf\?type$ $\mathit{rdfs}\?Literal.$" by
the RDF and RDFS entailment rules, where $x$ is a blank node
allocated to $l$\footnote{In RDF(S), literals are not allowed in the subject position of 
RDF triples, whereas blank nodes are. For this reason, before the RDF and RDFS entailment rules
are applied on an RDF graph, each literal is replaced by a unique blank node.
This way inferences can be drawn on the literal value denoted by this literal,
without concern for the above restriction \cite{rdfsemantics}.}. Such a
triple is called {\em XML clash}. 
To understand this, note that from semantic condition 3 of Definition \ref{def:RDFInterpretation}
(RDF interpretation, Appendix A), it follows that  the denotation of an ill-typed XML literal
cannot be a literal value. Now, from semantic conditions 1 and 2 of Definition \ref{def:RDFSInterpretation} 
(RDFS interpretation, Appendix A), it follows that the denotation of an ill-typed XML literal
cannot be of type $\mathit{rdfs}\?Literal$. Therefore, the derivation of an XML clash from
an RDF graph $G$ through the application of the RDF and RDFS entailment rules,
indicates that there is no RDFS interpretation that satisfies $G$.

An ERDF graph can be
ERDF-inconsistent\footnote{Meaning that there is no
 (coherent) ERDF interpretation that satisfies the ERDF graph.}, not only
due to the appearance of an ill-typed XML literal in the ERDF
graph (in combination with semantic condition 15 of Definition \ref{def:ERDFInterpretation}), 
but also due to the additional semantic conditions for coherent
ERDF interpretations.

For example, let $p,q,s,o \in \URI$ and let $G=\{p(s,o),  \; \mathit{rdfs}\?\mathit{subPropertyOf}(p, q), \;
\sneg q(s,o)\}$. Then, $G$ is
ERDF-inconsistent, since there is no (coherent) ERDF interpretation that satisfies $G$.

\godown
The following proposition shows that for total properties and
total classes of (coherent) ERDF interpretations, weak negation and  strong negation coincide (boolean truth values).

\begin{proposition} \label{prop:classicalNegation}
{ \em Let $I$ be an ERDF interpretation of a vocabulary $V$ and let $V'=V \cup \V_{RDF}
\cup \V_{RDFS} \cup \V_{ERDF}$. Then,
\begin{enumerate}

\item For all $p,s,o \in V'$ such that $I(p) \in \TProp_I$, it holds:\\
   $I \models \wneg p(s,o)$ iff $I \models \sneg p(s,o)$ (equivalently, $I \models p(s,o) \vee \sneg p(s,o)$).

\item For all $x,c \in V'$ such that $I(c) \in \TCls_I$, it holds:\\
 $I \models \wneg \rdf\?type(x, c)$ iff $I \models \sneg \rdf\?type(x, c)$ \\
 (equivalently, $I \models \rdf\?type(x, c) \vee \sneg \rdf\?type(x, c)$).

\end{enumerate}}
\end{proposition}

Below we define ERDF entailment between two ERDF
formulas or ERDF graphs.

\begin{definition} [ERDF entailment]
{\em Let $F,F'$ be ERDF formulas or ERDF graphs. We say that $F$ {\em
ERDF-entails} $F'$ ($F \models^{\ERDF} F'$) iff for every ERDF
interpretation $I$, if $I \models F$ then $I \models F'$. $\Box$
}
\end{definition}

For example, let: 
\begin{center}
$F=\forall ?x \; \exists ?y \; (\rdf\?type(?x,
\mathit{Person}) \mImpl  \mathit{hasFather}(?x, ?y))\;$
 $\AND \; \rdf\?type(John, \mathit{Person})$.
 \end{center}
Additionally, let
 $F'=\exists ?y \; \mathit{hasFather}(John,?y) \;$
 $\AND \; $ 
 $\rdf\?type(\mathit{hasFather}, \rdf\?\mathit{Property})$.\\
Then $F \models^{\ERDF} F'$.

The following proposition shows that ERDF entailment extends RDFS
entailment \cite{rdfsemantics} (see also Appendix A) from RDF graphs to ERDF formulas.
In other words, ERDF entailment is upward compatible with RDFS entailment.

\begin{proposition} \label{ERDFextendsRDFS}
{ \em Let $G,G'$ be RDF graphs such that $V_G \cap
\V_{\ERDF}=\emptyset$ and $V_{G'} \cap \V_{\ERDF}=\emptyset$. Then,
$G \models^{RDFS} G'$ iff $G \models^{\ERDF} G'$.}
\end{proposition}

It easily follows from Proposition \ref{ERDFextendsRDFS} that an RDF graph is RDFS
satisfiable iff it is ERDF satisfiable. Thus, an RDF graph can be ERDF-inconsistent only due to an XML clash.

\section{ERDF Ontologies \& Herbrand Interpretations} \label{sec:ERDF_ontologies}

In this section, we define an ERDF ontology as a pair of an ERDF
graph $G$ and a set $P$ of ERDF rules. ERDF rules should be
considered as derivation rules that allow us to infer more
ontological information based on the declarations in $G$.
Moreover, we define the Herbrand interpretations and the minimal Herbrand models
of an ERDF ontology.

\begin{definition}[ERDF rule, ERDF program]
{\em An {\em ERDF rule} $r$ over a vocabulary $V$ is an expression of the form:
$G \leftarrow F$, where $F \in L(V) \cup \{true\}$ is called \emph{condition}
and $G \in L(V|\{\sneg\}) \cup \{\mathit{false}\}$ is called
\emph{conclusion}. We assume that no bound variable in $F$ appears free in $G$. We denote the set of variables and the set of
free variables of
$r$ by $\Var(r)$ and $\FVar(r)$\footnote{$\FVar(r)=\FVar(F) \cup \FVar(G)$.}, respectively. Additionally, we write
$Cond(r)=F$ and $Concl(r)=G$.\\
An {\em ERDF program} $P$ is a set of ERDF rules over some
vocabulary $V$. We denote the set of URI references and literals
appearing in $P$ by $V_P$. $\Box$}
\end{definition}

Recall that $L(V|\{\sneg\})$ denotes the set of ERDF triples over $V$. Therefore,
the conclusion of an ERDF rule, unless it is $\mathit{false}$, it is either a positive ERDF triple $p(s,o)$  or a negative
ERDF triple $\neg p(s,o)$.

For example,  consider the derivation rule $r$:
{\small
\begin{center}
$\mathit{allRelated}(?P, ?Q) \leftarrow \forall ?p \; \rdf\?type(?p,?P) \mImpl \exists ?q \; (\rdf\?type(?q,?Q) \; \AND \; \mathit{related}(?p,?q))$,
\end{center}
}
\noindent
Then, $r$ is an ERDF rule, indicating that between two classes $P$ and $Q$, it holds $\mathit{allRelated}(P, $ 
$Q)$ if for all instances $p$ of the class $P$,
there is an instance $q$ of the class $Q$ such that it holds $\mathit{related}(p,q)$.
Note that $\Var(r)=\{?P,?Q,?p,?q\}$ and $\FVar(r)=\{?P,?Q\}$.

When $Cond(r)=true$ and $\Var(r)=\{\}$, rule $r$ is called {\em ERDF fact}.
When $Concl(r)=\mathit{false}$, rule $r$ is called {\em ERDF constraint}.
We assume that for every partial interpretation $I$ and every function $v: \Var \rightarrow Res_I$, 
it holds $I,v \models true$, $I \models true$, $I,v \not \models \mathit{false}$, and $I \not \models \mathit{false}$.

Intuitively, an ERDF ontology is the combination of (i) an ERDF graph $G$ containing
(implicitly existentially quantified) positive and negative information, and (ii)
an ERDF program $P$ containing
derivation rules (whose free variables are implicitly universally quantified).

\begin{definition}[ERDF ontology]
{\em An {\em ERDF ontology} (or {\em ERDF knowledge base}) is a
pair $O=\< G,P \>$, where $G$ is an ERDF graph and $P$
is an ERDF program. $\Box$}
\end{definition}

The following definition defines the models of an ERDF ontology.

\begin{definition} [Satisfaction of an ERDF rule and an ERDF ontology] \label{def:modelsRuleOntology}
{\em Let $I$ be an ERDF interpretation of a vocabulary $V$.
\vspace{-0.1cm}
\begin{itemize}
\item We say that $I$ {\em satisfies} an ERDF rule $r$, denoted by $I \models r$,
iff for all mappings $v:\Var(r) \rightarrow Res_I$ such that
$I,v \models Cond(r)$, it holds $I, v \models Concl(r)$.

\item We say that $I$ {\em satisfies} an ERDF ontology $O=\<G,P\>$ (also, $I$ is a $model$ of $O$), denoted by $I
\models O$, iff $I \models G$ and $I \models r$, $\forall \; r \in
P$. $\Box$
\end{itemize}
}
\end{definition}

In this paper, existentially quantified variables in ERDF graphs are handled by 
{\em skolemization}, a syntactic transformation commonly used in automatic inference systems for removing existentially quantified variables.
\begin{definition} [Skolemization of an ERDF graph]
 {\em Let $G$ be an ERDF graph. The {\em skolemization function} of $G$ is an
$1\?1$ mapping $sk_G:\Var(G) \rightarrow \URI$, where for each $x \in
\Var(G)$, $sk_G(x)$ is an artificial URI, denoted by $G\?x$. The
set $sk_G(\Var(G))$ is called the {\em Skolem vocabulary} of $G$.\\
The {\em skolemization} of $G$, denoted by $sk(G)$, is the ground
ERDF graph derived from $G$ after replacing each variable $x \in
\Var(G)$ by $sk_G(x)$. $\Box$}
\end{definition}

Intuitively, the Skolem vocabulary of $G$ (that is, $sk_G(\Var(G))$) contains artificial URIs giving ``arbitrary" names to the anonymous entities whose existence was asserted by the use of blank nodes in $G$.

For example, let:
 $G=\{\rdf\?type(?x, EuropeanCountry),$ $\sneg \rdf\?type(?x, EUmember)\}$. 
Then, 
{\small
\begin{center}
$sk(G)=\{\rdf\?type(sk_G(?x), EuropeanCountry),$
$\sneg \rdf\?type(sk_G(?x), EUmember)\}$.
\end{center}
}

The following proposition expresses that the skolemization of an ERDF graph
has the same
entailments as the original graph, provided
that these do not contain URIs from the skolemization vocabulary.

\begin{proposition} \label{ERDFentailmentEquivalence1}
{\em Let $G$ be an ERDF graph and let $F$ be an ERDF formula such that $V_{F} \cap sk_G(\Var(G)) =\emptyset$. It holds:
$G \models^{ERDF} F$  iff $sk(G) \models^{ERDF} F$.}
\end{proposition}

Below we define the vocabulary of an ERDF ontology $O$.

\begin{definition} [Vocabulary of an ERDF ontology]
{\em Let $O=\<G,P\>$ be an ERDF ontology. The {\em vocabulary} of
$O$ is defined as $V_O=V_{sk(G)} \cup V_P \cup \V_{RDF} \cup
\V_{RDFS} \cup \V_{ERDF}$. $\Box$}
\end{definition}

\vspace{-0.1cm}

Note that the vocabulary of an ontology $O=\<G,P\>$ contains the
skolemization vocabulary of $G$.

Let $O=\<G,P\>$ be an ERDF ontology. We denote by $Res^H_{O}$ the
union of $V_{O}$ and the set of XML values of the well-typed XML
literals in $V_{O}$ minus the well-typed XML literals.

The following definition defines the Herbrand interpretations and the Herbrand models
of an ERDF ontology.

\begin{definition} [Herbrand interpretation, Herbrand model of an ERDF ontology]
\label{def:HerbrandInterpretation} {\em Let $O=\<G,P\>$ be an ERDF ontology and let $I$ be an
ERDF interpretation of $V_{O}$.
We say that $I$ is a {\em Herbrand interpretation} of $O$ iff:
\vspace{-0.1cm}
 \begin{itemize}

\item $Res_I=Res^H_{O}$.

\item $I_V(x)=x$, for all $x \in V_{O} \cap \URI$.

\item $\IL_I(x)=x$, if $x$ is a typed literal in $V_{O}$ other than
a well-typed XML literal, and $\IL_I(x)$ is the XML value of $x$,
if $x$ is a well-typed XML literal in $V_{O}$.

\end{itemize}

\vspace{-0.1cm}
\noindent
We denote the set of Herbrand interpretations of $O$ by
$\I^H(O)$.\\
A Herbrand interpretation $I$ of $O$ is a {\em Herbrand model} of
$O$ iff $I \models \<sk(G),P\>$. We denote the set of Herbrand models of $O$
by $\M^H(O)$. $\Box$}
\end{definition}

Note that if $I$ is a Herbrand interpretation of an ERDF ontology $O$ then $I(x)=x$,
for each $x \in V_O$ other than a well-typed XML literal.

It is easy to see that 
every Herbrand model of an ERDF ontology $O$ is a model of $O$.
Moreover, note that 
every Herbrand interpretation of an ERDF ontology $O$ is uniquely identified by (i) 
its set of properties and (ii) its property-truth and property-falsity extension mappings.

However, not all Herbrand models of an ERDF ontology $O$ are
desirable. For example, let $p,s,o \in \URI$, let $G=\{p(s,o)\}$, and let
$O=\<G,\emptyset\>$. Then, there is a Herbrand model $I$ of $O$
such that $I \models p(o,s)$, whereas we want $\wneg p(o,s)$ to be
satisfied by all intended models of $O$. This is because $p$ is not a total
property
and $p(o,s)$ cannot be derived from $O$
(negation-as-failure)\footnote{On the other hand, if $p$
is a total property then $p(o,s) \OR \sneg p(o,s)$ should be
satisfied by all intended models. Therefore, in this case, there should be an intended model of $O$ that satisfies  $p(o,s)$.} .

Before we define the minimal Herbrand interpretations of an ERDF ontology
$O$, we need to define a partial ordering on the Herbrand
interpretations of $O$.

\begin{definition} [Herbrand interpretation ordering] \label{def:ordering}
{\em Let $O=\<G,P\>$ be an ERDF ontology.  Let $I, J \in \I^H(O)$.
We say that $J$ $extends$ $I$, denoted by $I \leq J$ (or $J \geq I$), iff
 $Prop_{I} \subseteq Prop_{J}$, and for all $p \in Prop_I$, it holds $\PT_{I}(p) \subseteq \PT_{J}(p)$  and
$\PF_{I}(p) \subseteq \PF_{J}(p)$. $\Box$
}
\end{definition}

It is easy to verify that the relation $\leq$  is reflexive, transitive, and antisymmetric. Thus, it is  a partial ordering on $\I^H(O)$.

The intuition behind Definition \ref{def:ordering} is that by extending a Herbrand interpretation,
we extend both the truth and falsity extension
for all properties, and thus (since $\rdf\?type$ is a property), for all classes.

The following proposition expresses that two Herbrand interpretations $I,J$ of an ERDF ontology $O$ are incomparable,
if the property-truth or property-falsity extension of a total property $p$ w.r.t. $I$ and $J$ are different.

\begin{proposition} \label{prop:incomparable}
{\em Let $O=\<G,P\>$ be an ERDF ontology and let $I, J \in \I^H(O)$.
Let $p \in \TProp_I \cap \TProp_J$. If $\PT_I(p) \not = \PT_J(p)$ or 
$\PF_I(p) \not = \PF_J(p)$ then $I \not \leq J$ and $J \not \leq I$.}
\end{proposition}

\begin{definition} [Minimal Herbrand interpretations]
{\em Let $O$ be an ERDF ontology and let $\I \subseteq \I^H(O)$.
We define $minimal(\I)=\{ I \in \I \;|\; \not \exists J \in \I: \; J \not= I$ and $J \leq I\}$.  $\Box$}
\end{definition}

We define
the {\em minimal Herbrand models} of $O$, as: 
\begin{center}
 $\M^{min}(O)= minimal(\M^H(O))$.
\end{center}

However minimal Herbrand models do not give the intended semantics
to all ERDF rules. This is because ERDF rules are derivation and
not implication rules. Derivation rules are often identified with
implications. But, in general, these are two different concepts.
While an implication is an expression of a logical formula
language, a derivation rule is rather a meta-logical expression.
There are logics, which do not have an implication connective, but
which have a derivation rule concept. In standard logics (such as
classical and intuitionistic logic), there is a close relationship
between a derivation rule (also called ``sequent") and the
corresponding implicational formula: they both have the same
models. For non-monotonic rules (e.g. with negation-as-failure),
this is no longer the case: the intended models of such a rule
are, in general, not the same as the intended models of the
corresponding implication. This is easy to see with the help of an
example. Consider the rule $p \leftarrow \wneg q$
whose model set, according to the stable model semantics
\cite{gelfond88:sms,gelfond90:asets,HW97,HJW96},
is $\{\{p\}\}$, that is,
it entails $p$. On the other hand, the model set of the
corresponding implication $\wneg q \mImpl p$,
which is equivalent to the disjunction $p \lor q$, is $\{\{p\},
\{q\}, \{p,q\}\}$; consequently, it does not entail $p$.

Similarly, let $O=\<\emptyset, P\>$, where $P=\{p(s,o) \leftarrow
\wneg q(s,o)\}$ and $p,q,s,o \in \URI$.  Not all minimal Herbrand
models of $O$ are intended. In particular,
there is $I \in \M^{min}(O)$ such that $I \models q(s,o) \; \AND \;
\wneg p(s,o)$, whereas we want $\wneg q(s,o) \; \AND \; p(s,o)$ to
be satisfied by all intended models of $O$, as $q$ is not a total property
and $q(s,o)$ cannot be derived by any rule
(negation-as-failure).

To define the intended ($stable$) models of an ERDF ontology, we
need first to define grounding of ERDF rules.

\begin{definition} [Grounding of an ERDF program]
{\em Let $V$ be a vocabulary and let $r$ be an ERDF rule. We denote by
$[r]_{V}$ the set of rules that result from $r$ if we replace
each variable $x \in \FVar(r)$ by
$v(x)$, for all mappings $v:\FVar(r) \rightarrow V$.\\

Let $P$ be an ERDF program. We define $[P]_{V}= \bigcup_{r \in P}
[r]_{V} $. $\Box$}
\end{definition}

Note that a rule variable  can naturally appear in the subject position of an ERDF triple.
Since variables can be instantiated by a literal, a literal can naturally appear in the subject position of 
an ERDF triple in the grounded version of an ERDF program. 
This case further supports our choice of allowing literals in the subject position of an ERDF triple.

\section{ERDF Stable Models} \label{sec:ERDF_stableModel}

In this section, we define the intended models of an ERDF ontology $O$,
called {\em stable models} of $O$, based on minimal Herbrand
interpretations. In particular, defining the stable models of $O$,
only the minimal interpretations from a set of Herbrand
interpretations that satisfy certain criteria are considered.

Below, we define the stable models of an ERDF ontology, 
based on the {coherent} 
stable models\footnote{Note that these models on  extended logic programs are equivalent \cite{HJW96} to Answer Sets  of answer set semantics \cite{gelfond90:asets}.} of Partial Logic \cite{HJW96}.

\begin{definition} [ERDF stable model] \label{def:stableModel}
{\em Let $O=\<G,P\>$ be an ERDF ontology and let $M \in
\I^H(O)$. We say that $M$ is an {\em (ERDF) stable model} of $O$ iff there
is a chain of Herbrand interpretations of $O$, $I_0 \leq... \leq
I_{k+1}$ such that $I_{k}=I_{k+1}=M$ and:
\vspace{-0.1cm}
\begin{enumerate}

\item $I_0 \in minimal(\{I \in \I^H(O) \; | \; I \models sk(G)\})$.

\item For successor ordinals $\alpha$ with $0 < \alpha \leq k+1$:
\\
$I_\alpha \in minimal(\{I \in \I^H(O) \;|\; I \geq I_{\alpha-1}$ and $ I
\models Concl(r),$ $\forall r \in P_{[I_{\alpha-1},
M]}\})$, where\\
$P_{[I_{\alpha-1}, M]}=\{r \in [P]_{V_{O}} \; | \; I \models
Cond(r), \; \forall I \in \I^H(O)$ s.t. $I_{\alpha-1} \leq I \leq M\}$.
\vspace{0.1cm}

\end{enumerate}
\vspace{-0.1cm}
The set of stable models of $O$ is denoted by $\M^{st}(O)$.
$\Box$}
\end{definition}

Note that $I_0$ is a minimal Herbrand interpretation of $O=\<G,P\>$ that satisfies $sk(G)$,
while Herbrand interpretations $I_1,...,I_{k+1}$ correspond to a stratified sequence of rule applications,
where all applied rules remain applicable throughout the generation of a stable model $M$.
In our words, a stable model is generated bottom-up by the iterative application 
of the rules in the ERDF program $P$, starting
from the information in the ERDF graph $G$.
Thus, ERDF stable model semantics, as a refinement of minimal model semantics, 
captures the intuition that:
\begin{itemize}
\item Assertions $\rdf\?type(p,\erdf\?\mathit{TotalProperty})$ and $\rdf\?type(c,\erdf\?\mathit{TotalClass})$ 
should only be accepted if the ontology contains some direct support for them in the form of an acceptable rule sequence (that corresponds to a proof).
\item 	Assertions $p(s,o)$ and $\sneg p(s,o)$ should only be accepted if the ontology 
contains some direct support for them in the form of an acceptable rule sequence, or \\
$\rdf\?type(p,\erdf\?\mathit{TotalProperty})$ is accepted.

\item	Assertions $\rdf\?type(o,c)$ and $\sneg rdf\?type(o,c)$ should only be accepted if the ontology 
contains some direct support for them in the form of an acceptable rule sequence, or 
$\rdf\?type(c,\erdf\?\mathit{TotalClass})$ is accepted.

\end{itemize}

\godown
\noindent
{\bf Wine Selection Example}: Consider a class {\em Wine} whose instances are wines, and 
a property $likes(X,Y)$ indicating that person $X$ likes object $Y$.
Assume now that we want to select wines for a dinner
such that, for each guest, there is on the table exactly one wine that he/she likes.
Let the class $\mathit{Guest}$ indicate the
persons that will be invited to the dinner and let the class 
$\mathit{SelectedWine}$ indicate the wines chosen to be served.
An ERDF program $P$ that describes this wine selection problem 
is the following (commas ``," in the body of the rules
indicate conjunction $\AND$):
{\small
\begin{tabbing}
 aaaaaaaaaa\= aaa \= aaaaaaaaaa \= aaaaaa \=  \kill

 $id(?x,?x) $  $\;\; \leftarrow \; \rdf\?type(?x, \rdfs\?\mathit{Resource}).$\\
 $\rdf\?type(?y, \mathit{SelectedWine})$ \> \> \> $\leftarrow \rdf\?type(?x,\mathit{Guest}), \rdf\?type(?y,\mathit{Wine}), likes(?x,?y),$\\
 \> \>  $\forall ?z \; (\rdf\?type(?z, \mathit{SelectedWine}), \wneg id(?z,?y) \; \mImpl \; \wneg likes(?x,?z))$.
 \end{tabbing}
}

Consider now the ERDF graph $G$, containing the factual information:
{\small
\begin{tabbing}
 aaaaaa\= aaaaaaa \= aaaaa \= aaaaaa \=  \kill

$G= \{$ \> $\rdf\?type(\mathit{Carlos}, \mathit{Guest})$, $\rdf\?type(\mathit{Gerd}, \mathit{Guest})$, $\rdf\?type(\mathit{Riesling}, \mathit{Wine})$,\\
\> $\rdf\?type(\mathit{Retsina}, \mathit{Wine})$, $\rdf\?type(\mathit{Chardonnay}, \mathit{Wine})$, $likes(\mathit{Gerd}, \mathit{Riesling})$,\\ 
\> $likes(\mathit{Gerd}, \mathit{Retsina})$,  $likes(\mathit{Carlos}, \mathit{Chardonnay})$, $likes(\mathit{Carlos}, \mathit{Retsina})\; \}$.
 \end{tabbing}
}

Then, according to Definition \ref{def:stableModel},
the ERDF ontology $O=\<G, P\>$ has two stable models, $M_1$ and $M_2$, such that:
{\small
\begin{tabbing}
 aaaaaa\= aaaaaaa \= aaaaa \= aaaaaa \=  \kill
$M_1 \models$ \> $\rdf\?type(\mathit{Riesling},\mathit{SelectedWine}) \;\; \AND \;\; \rdf\?type(\mathit{Chardonnay},\mathit{SelectedWine}) \;\; \AND \;$\\
\> $\wneg \; \rdf\?type(\mathit{Retsina},\mathit{SelectedWine})$.\\\\
$M_2 \models$ \> $\rdf\?type(\mathit{Retsina},\mathit{SelectedWine}) \;\; \AND \;\; \wneg \; \rdf\?type(\mathit{Riesling},\mathit{SelectedWine}) \;\; \AND \;$\\ 
\> $\wneg \; \rdf\?type(\mathit{Chardonnay},\mathit{SelectedWine})$.
 \end{tabbing}
}

Note that, according to stable model $M_1$, the wines selected for the dinner are $\mathit{Riesling}$ and $\mathit{Chardonnay}$. 
This is because, (i)
$\mathit{Gerd}$ likes $\mathit{Riesling}$ but does not like $\mathit{Chardonnay}$, and (ii)
$\mathit{Carlos}$ likes $\mathit{Chardonnay}$ but does not like $\mathit{Riesling}$.

According to stable model $M_2$, only $\mathit{Retsina}$ is selected for the dinner. This is because,
both $\mathit{Gerd}$ and $\mathit{Carlos}$ like $\mathit{Retsina}$. 

Stable model $M_1$ is reached through the chain $I_0 \leq M_1  \leq M_1$, 
where $I_0$ is the single Herbrand interpretation
in $minimal(\{I \in \I^H(O) \; | \; I \models sk(G)\})$.
To verify this, note that: 

{\small
\begin{tabbing}
 aaaaa\= aaaa \= aaaaaaaaaaaaaaaaaaaaa \= aa \=  \kill
 $P_{[I_0, M_1]} = P_{[M_1, M_1]}=$\\\\

 $[id(?x,?x) $  $\; \leftarrow \; \rdf\?type(?x, \rdfs\?\mathit{Resource})]_{V_O}\;  \cup $\\\\
$\{ \rdf\?type(\mathit{Riesling}, \mathit{SelectedWine})$ \> \> \> $\leftarrow \; \rdf\?type(\mathit{Gerd},\mathit{Guest}),$\\ 
  \> \>    $\rdf\?type(\mathit{Riesling},\mathit{Wine}), likes(\mathit{Gerd},\mathit{Riesling}),$\\
 \> \>  $\forall ?z \; (\rdf\?type(?z, \mathit{SelectedWine}), \wneg id(\mathit{?z, Riesling}) \; \mImpl \; \wneg likes(\mathit{Gerd},?z))\} \; \cup$ \\\\
 $\{ \rdf\?type(\mathit{Chardonnay}, \mathit{SelectedWine})$ \> \> \> \> $\leftarrow \rdf\?type(\mathit{Carlos},\mathit{Guest}),$\\ 
\> \> $\rdf\?type(\mathit{Chardonnay},\mathit{Wine}), likes(\mathit{Carlos},\mathit{Chardonnay}),$\\
 \> \>  $\forall ?z \; (\rdf\?type(?z, \mathit{SelectedWine}), \wneg id(\mathit{?z, Chardonnay}) \; \mImpl \; \wneg likes(\mathit{Carlos},?z))\}.$
 \end{tabbing}
}

\noindent
Similarly, stable model $M_2$ is reached through the chain $I_0 \leq M_2  \leq M_2$.
To verify this, note that: 

{\small
\begin{tabbing}
 aaaaa\= aaa \= aaaaaaaaaaaaaaaaaaaa \= aaaaa \=  \kill
$P_{[I_0, M_2]} = P_{[M_2, M_2]}= $\\\\

$[id(?x,?x) $  $\; \leftarrow \; \rdf\?type(?x, \rdfs\?\mathit{Resource})]_{V_O}\;  \cup $\\\\
$\{ \rdf\?type(\mathit{Retsina}, \mathit{SelectedWine})$ \> \> \> $\leftarrow \; \rdf\?type(\mathit{Gerd},\mathit{Guest}),$\\ 
  \> \>    $\rdf\?type(\mathit{Retsina},\mathit{Wine}), likes(\mathit{Gerd},\mathit{Retsina}),$\\
 \> \>  $\forall ?z \; (\rdf\?type(?z, \mathit{SelectedWine}), \wneg id(?z, \mathit{Retsina}) \; \mImpl \; \wneg likes(\mathit{Gerd},?z))\} \; \cup$ \\\\
 $\{ \rdf\?type(\mathit{Retsina}, \mathit{SelectedWine})$ \> \> \> $\leftarrow \rdf\?type(\mathit{Carlos},\mathit{Guest}),$\\ 
\> \> $\rdf\?type(\mathit{Retsina},\mathit{Wine}), likes(\mathit{Carlos},\mathit{Retsina}),$\\
 \> \>  $\forall ?z \; (\rdf\?type(?z, \mathit{SelectedWine}), \wneg id(?z, \mathit{Retsina}) \; \mImpl \; \wneg likes(\mathit{Carlos},?z))\}.$
 \end{tabbing}
}

Assume now that $Retsina$ should not be one of the selected wines,
because it does not match with the food.
To indicate this, we add to $P$ the ERDF constraint: 
\begin{center}
$\; \mathit{false \leftarrow \rdf\?type(Retsina,SelectedWine)}$.
\end{center}
Then, $M_1$ is the single model of the modified ontology.

It is easy to verify that if $O$ is an ERDF ontology and $O'$ is exactly as $O$, but without the ERDF constraints
appearing in $O$, then $\M^{st}(O) \subseteq \M^{st}(O')$. In other words, the ERDF constraints appearing in
an ERDF ontology eliminate undesirable stable models.

\godown
\noindent 
{\bf Paper Assignment Example}: Consider a class $\mathit{Paper}$ whose instances are
papers submitted to a conference, a class $\mathit{Reviewer}$ whose instances are potential reviewers
for the submitted papers, and a property 
$\mathit{conflict(R,P)}$ indicating that there is a conflict
of interest between reviewer $R$ and paper $P$. Assume now that we want to assign papers to
reviewers based on the following criteria: (i) a paper is assigned to at most one reviewer, (ii) a reviewer is
assigned at most one paper, and (iii) a paper is not assigned to a reviewer, if there is a
conflict of interest. The assignment of a paper $P$ to a reviewer $R$ is indicated through the property
$\mathit{assign}(P,R)$. The ERDF triple $\mathit{allAssigned}(\mathit{Paper}, \mathit{Reviewer})$ indicates
that each paper has been assigned to one reviewer. An ERDF program $P$ describing the assignment of papers 
is the following:

{\small
\begin{tabbing}
 aaaaaaaaaaaaaaa\= aa \= aaaaa \= aaaaaa \= aaa \= \kill

 $id(?x,?x) $  $\;\; \leftarrow \; true.$\\
 $\sneg \mathit{assign}(?p, ?r)$ \> $\leftarrow \rdf\?type(?p,\mathit{Paper}), \rdf\?type(?p',\mathit{Paper}), \mathit{assign}(?p',?r), \wneg id(?p,?p')$.\\
 $\sneg \mathit{assign}(?p, ?r)$ \> $\leftarrow \rdf\?type(?r,\mathit{Reviewer}), \rdf\?type(?r',\mathit{Reviewer}), \mathit{assign}(?p,?r'), 
  \wneg id(?r,?r')$.\\
 $\sneg \mathit{assign}(?p, ?r)$ \> $\leftarrow \mathit{conflict}(?r,?p)$.\\
 $\mathit{assign}(?p, ?r)$ \> $\leftarrow \rdf\?type(?r,\mathit{Reviewer}), \rdf\?type(?p,\mathit{Paper}), \wneg \; \sneg \mathit{assign}(?p,?r).$\\
 $\mathit{allAssigned}(\mathit{Paper}, \mathit{Reviewer})$  $\leftarrow \forall ?p \; (\rdf\?type(?p,\mathit{Paper}) \mImpl $\\
 \>\> \>\> \> $\exists ?r \; (\rdf\?type(?r,\mathit{Reviewer}) \; \AND \; \mathit{assign}(?p,?r))). $
 \end{tabbing}
}

Consider now the ERDF graph $G$, containing the factual information:
{\small
\begin{tabbing}
 aaaaaa\= aaaaaaa \= aaaaa \= aaaaaa \=  \kill

$G= \{$ \> $\rdf\?type(P1, \mathit{Paper})$, $\rdf\?type(P2, \mathit{Paper})$, $\rdf\?type(P3, \mathit{Paper})$, $\rdf\?type(R1, \mathit{Reviewer})$,\\ \> $\rdf\?type(R2, \mathit{Reviewer})$, $\rdf\?type(R3, \mathit{Reviewer})$, $\mathit{conflict}(P1,R3)$, $\mathit{conflict}(P2,R2)$,\\ 
\> $\mathit{conflict}(P3,R2)\; \}$.
 \end{tabbing}
}

Then, according to Definition \ref{def:stableModel},
the ERDF ontology $O=\<G, P\>$ has four stable models, denoted by $M_1$, ..., $M_4$, such that:
{\small
\begin{tabbing}
 aaaaaa\= aaaaaaa \= aaaaa \= aaaaaa \=  \kill
$M_1 \models$ \> $\mathit{assign}(P1,R1) \; \AND \; \mathit{assign}(P2,R3) \; \AND \; \wneg \mathit{allAssigned}(\mathit{Paper}, \mathit{Reviewer})$,\\
$M_2 \models$ \> $ \mathit{assign}(P1,R1) \; \AND \; \mathit{assign}(P3,R3) \; \AND \; \wneg \mathit{allAssigned}(\mathit{Paper}, \mathit{Reviewer})$,\\
$M_3 \models$ \> $ \mathit{assign}(P1,R2) \; \AND \; \mathit{assign}(P2,R1) \; \AND \; \mathit{assign}(P3,R3) \; \AND $ $\mathit{allAssigned}(\mathit{Paper}, \mathit{Reviewer})$, \\
$M_4 \models$ \> $ \mathit{assign}(P1,R2) \; \AND \; \mathit{assign}(P2,R3) \; \AND \; \mathit{assign}(P3,R1) \; \AND$  $\mathit{allAssigned}(\mathit{Paper}, \mathit{Reviewer})$.
 \end{tabbing}
}

\godown
We would like to note that, in contrast to the previous examples, given an ERDF ontology $O=\<G,P\>$,
it is possible that $|\mathit{minimal}(\{I \in \I^H(O) \; | \; I \models sk(G)\})| >1$,
due to the declaration of total properties and total classes. Specifically, the number of 
the interpretations $I_0$ in item 1 of Definition \ref{def:stableModel} is more than one 
iff $G$ contains ERDF triples of the form $\rdf\?type(p, \erdf\?\mathit{TotalProperty})$ or 
$\rdf\?type(c, \erdf\?\mathit{TotalClass})$.
For example, let $O=\<G, \emptyset\>$, where:

{\small
\begin{center}
 $G= \{\mathit{authorOf}(\mathit{John},\mathit{book1}), \mathit{authorOf}(\mathit{Peter},\mathit{book2}), $
$\rdf\?type(\mathit{authorOf}, \erdf\?\mathit{TotalProperty})\}$.
\end{center}
}

Then, there are $I_0,I'_0 \in \mathit{minimal}(\{I \in \I^H(O) \; | \; I \models sk(G)\})$ such that:
\begin{center}
{ $I_0 \models \mathit{authorOf}(\mathit{John}, \mathit{book2})\;$} and 
{ $\;I'_0 \models \sneg \mathit{authorOf}(\mathit{John}, \mathit{book2})$}.
\end{center}
Note that both $I_0$ and $I'_0$ are stable models of $O$. However, $I_0$ satisfies $\mathit{authorOf}(\mathit{John}, \mathit{book2})$,
even though there is no evidence that $\mathit{John}$ is an author of $\mathit{book2}$.

\godown

The following proposition shows that a stable model of an ERDF ontology $O$ is a
Herbrand model of $O$.

\begin{proposition} \label{prop:stableImpliesHerbrand}
{\em Let $O=\<G,P\>$ be an ERDF ontology and let $M \in \M^{st}(O)$. It
holds $M \in \M^H(O)$.}
\end{proposition}

On the other hand, if all properties are total, a Herbrand model $M$ of an
ERDF ontology $O=\<G,P\>$ is a stable model of $O$\footnote{
Note that, in this case, $M \in minimal(\{I \in \I^H(O) \; | \; I \models sk(G)\})$
and $M \in minimal(\{I \in \I^H(O) \;|\; I \geq M$ and $ I
\models Concl(r),$ $\forall \; r \in P_{[M,M]}\})$.}.
Obviously, this is a desirable result since, in this case, an open-world assumption is
made for all properties. Thus, there is no preferential entailment of  weak negation, for any of the properties.
Of course, the term ``stable model" is not very descriptive, for this degenerative case.

\begin{proposition} \label{TotalStableHerbrand}
{\em Let $O=\<G,P\>$ be an ERDF ontology such that\\
{\small $\mathit{rdfs\?subClassOf}(\rdf\?\mathit{Property},
 \erdf\?\mathit{TotalProperty})$} $ \in G$. Then, 
$\M^{st}(O)=\M^H(O)$.}
\end{proposition}

A final note is that, similarly to stable models defined by Gelfond \& Lifschitz \citeyear{gelfond88:sms,gelfond90:asets}
and Herre et al. \citeyear{HJW96}, 
ERDF stable models do not preserve Herbrand model satisfiability. For example,
let $O=\<\emptyset, P\>$, where $P=\{p(s,o) \leftarrow \wneg
p(s,o)\}$ and $p,s,o \in \URI$. Then, $\M^{st}(O)=\emptyset$,
whereas there is a Herbrand model of $O$ that satisfies $p(s,o)$.

\section{ERDF Stable Model Entailment \& Stable Answers} \label{sec:StableEntailment}

In this section, we define stable model entailment on ERDF ontologies, showing that it extends ERDF entailment on ERDF graphs.
Moreover, we define the skeptical and credulous answers of an ERDF formula (query) $F$
w.r.t. an ERDF  ontology $O$.

\begin{definition} [Stable model entailment]
{\em Let $O=\<G,P\>$ be an ERDF ontology and let $F$ be an ERDF
formula or ERDF graph. We say that $O$ {\em entails} $F$ under the {\em (ERDF) stable model semantics}, denoted by $O
\models^{st} F$ iff for all $M \in \M^{st}(O)$, $\;\; M \models
F$. $\Box$}
\end{definition}

For example, let $O=\<\emptyset, P\>$, where $P=\{p(s,o) \leftarrow
\wneg q(s,o)\}$ and $p,q,s,o \in \URI$.  Then, $O \models^{st} \wneg q(s,o) \; \AND \;
p(s,o)$. 

Now, let $G=\{\mathit{rdfs\?subClassOf}(\rdf\?\mathit{Property},
 \erdf\?\mathit{TotalProperty})\}$ and let $P$ be as in the previous example. Then, $\<G,P\> \models^{st} q(s,o) \; \OR \;
p(s,o)$, but $\<G,P\> \not \models^{st} \wneg q(s,o)$ and $\<G,P\> \not \models^{st} p(s,o)$.
Note that this is the desirable result, since now $q$ is a total property (and thus,
an open-world assumption is made for $q$).

As another example, let $p,s,o \in \URI$, let $G=\{p(s,o)\}$, and let $P=\{\sneg p(?x,?y) \leftarrow \wneg p(?x,?y)\}$. Then,
$\<G,P\> \models^{st} \wneg p(o,s) \;\AND\;
\sneg p(o,s)$ (note that $P$ contains a CWA on $p$).

Now, let $G=\{\rdf\?type(p,
\erdf\?\mathit{TotalProperty}),$
$ p(s,o)\}$ and let $P$ be as in the previous example. Then,
$\<G,P\> \models^{st} \forall ?x \;  \forall ?y \; $
$(p(?x,?y) \;
\OR \sneg p(?x,?y))$ (see Proposition \ref{prop:classicalNegation}), but $\<G,P\> \not \models^{st} \wneg p(o,s)$ and $\<G,P\> \not \models^{st} \sneg p(o,s)$.
Indeed, the CWA in $P$ does not affect the semantics of $p$, since $p$ is a total property.

\godown
\noindent
{\bf EU Membership Example}: Consider the following ERDF program $P$, specifying some rules for
concluding that a country is not a member state of the European
Union (EU).

{\small 
\[
\begin{array}{l@{\qquad}ll}
(r_1) & \sneg \rdf\?type(?x,\mbox{\em EUMember}) \arrow & \rdf\?type(?x,AmericanCountry).\\
(r_2) & \sneg \rdf\?type(?x,\mbox{\em EUMember}) \arrow & \rdf\?type(?x,EuropeanCountry),\\
& & \wneg \rdf\?type(?x,\mbox{\em EUMember}).
\end{array}
\]
}

A rather incomplete ERDF ontology $O=\<G,P\>$ is obtained by including the following information in the  ERDF
graph $G$:
{\small 
\[
\begin{array}{l@{\qquad}l}
\sneg \rdf\?type(Russia,\mbox{\em EUMember}). & \rdf\?type(Canada,AmericanCountry).\\
\rdf\?type(Austria,\mbox{\em EUMember}). & \rdf\?type(Italy,EuropeanCountry).\\
\rdf\?type(?x, EuropeanCountry). & \sneg \rdf\?type(?x, \mbox{\em EUMember}).
\end{array}
\]
}
Using stable model entailment on $O$, it can be concluded that Austria is a
member of EU, that Russia and Canada are not
members of EU, and that it exists a European Country which is not a member of EU.
However, it is also concluded that Italy is not a member of EU, which is a wrong
statement. This is because $G$ does not contain complete information of the European countries that are EU members
(e.g., it does not contain $\rdf\?type(Italy, \mbox{\em EUMember})$). Thus, incorrect information is obtained by the
closed-world assumption expressed in rule $r_2$.
In the case that $\rdf\?type(\mbox{\em EUMember}, \erdf\?\mathit{TotalClass})$ is added to $G$ (that is, an open-world assumption is made for
the class $\mbox{\em EUMember})$ then $\wneg \rdf\?type(Italy, \mbox{\em EUMember})$ and thus,
$\sneg \rdf\?type(Italy, \mbox{\em EUMember})$ are no
longer entailed. This is because, there is a stable model of the extended ERDF ontology $O$ that satisfies
$\rdf\?type(Italy, \mbox{\em EUMember})$. Moreover, if complete information for all European countries that are members of EU is included in $G$ then the stable model conclusions of $O$ will also be correct (the closed-world assumption will be correctly applied). Note that, in this case, $G$ will include the ERDF triple $\rdf\?type(Italy, \mbox{\em EUMember})$.

\godown

The following proposition follows directly from the fact that any stable model  of an ERDF ontology $O$ is an ERDF interpretation.

\begin{proposition} \label{prop:stable_stable}
{\em Let $O=\<G,P\>$ be an ERDF ontology and let $F, F'$ be ERDF
formulas. If $O \models^{st} F$ and $F \models^{ERDF} F'$ then $O
\models^{st} F'$.}
\end{proposition}

For ERDF graphs $G$, $G'$, it can be proved that
$\<G, \emptyset\> \models^{st} G'$ iff  $G \models^{ERDF} G'$ (see below).
Now the question arises whether this result can be generalized by replacing the ERDF graph $G'$ by any ERDF formula $F$.
The following example shows that this is not the case. Let $G=\{p(s,o)\}$ and let $F=\wneg p(o,s)$, where $p,s,o \in \URI$. 
Then $\<G, \emptyset\> \models^{st} F$, whereas
$G \not \models^{ERDF} F$. However, $G'$ can be replaced by any ERDF {\em d}-formula $F$,
defined as follows:

\begin{definition} [ERDF $d$-formula]
{\em 
Let $F$ be an ERDF formula. We say that $F$ is an {\em ERDF $d$-formula} iff (i) $F$ is the disjunction
of existentially quantified conjunctions of ERDF triples, and (ii) $\FVar(F)=\emptyset$. $\Box$
}
\end{definition}

\noindent
For example, let:
{
\begin{tabbing}
 aaaaaaaaaaaaa\= aaaa\= aaaaaaa \= aaaaa \= aaaaaa \=  \kill
\> $F= $ \> $\mathit{(\exists ?x \; \rdf\?type(?x, Vertex) \; \AND \; \rdf\?type(?x, Red))}\; \OR$\\
\> \> $\mathit{(\exists ?x \; \rdf\?type(?x, Vertex) \; \AND \; \rdf\?type(?x, Blue))}$.
 \end{tabbing}
}
\noindent
Then, $F$ is an ERDF $d$-formula. It is easy to see that if $G$ is an ERDF graph then $\formula(G)$
is an ERDF {\em d}-formula.

\begin{proposition} \label{ERDFentailmentEquivalence2}
{\em Let $G$ be an ERDF graph and let $F$ be an ERDF formula such that $V_{F} \cap sk_G(\Var(G)) =\emptyset$.
It holds:
\begin{enumerate}
\item If $F$ is an ERDF {\em d}-formula and $\<G,\emptyset\> \models^{st} F$ then $G \models^{ERDF} F$.
\item If $G \models^{ERDF} F$  then $\<G,\emptyset\> \models^{st} F$.
\end{enumerate}}
\end{proposition}

Let $G$ be an ERDF graph and let $F$ be an ERDF {\em d}-formula or an ERDF graph 
such that $V_{F} \cap sk_G(\Var(G)) =\emptyset$. A direct consequence of Proposition
\ref{ERDFentailmentEquivalence2} is that: 
\begin{center}
$\<G, \emptyset\> \models^{st} F$ iff $G \models^{ERDF} F$.
\end{center}


 The following proposition is a direct consequence of Proposition \ref{ERDFextendsRDFS} and 
 Proposition \ref{ERDFentailmentEquivalence2}, and shows that stable model
entailment extends RDFS entailment from RDF graphs to ERDF
ontologies.

\begin{proposition} \label{Prop:RDFSentailmentExtension}
{\em Let $G, G'$ be RDF graphs such that $V_G \cap \V_{ERDF} = \emptyset$, $V_{G'} \cap \V_{ERDF} =\emptyset$, and $V_{G'} \cap sk_G(\Var(G)) =\emptyset$. It holds:
$G \models^{RDFS} G'$  iff $\<G,\emptyset\> \models^{st} G'$.}
\end{proposition}

Recall that the Skolem vocabulary of $G$ (that is, $sk_G(\Var(G))$) contains artificial URIs giving ``arbitrary" names to the anonymous entities whose existence was asserted by the use of blank nodes in $G$. Thus, the condition
$V_{G'} \cap sk_G(\Var(G)) =\emptyset$ in Proposition \ref{Prop:RDFSentailmentExtension} is actually trivial.

\begin{definition} [ERDF query, ERDF stable answers] \label{stableAnswers}
{\em Let $O=\<G,P\>$ be an ERDF ontology. An  {\em (ERDF) query} $F$ is an ERDF
formula. The {\em (ERDF) stable answers} of $F$ w.r.t.  $O$ are
defined as follows:
 {\small
\[ \Ans^{st}_{O}(F)= \left\{ \begin{array}{ll}
\mbox{``yes"}  \;\; \;\;\;\; \mbox{if $\FVar(F)=\emptyset$ and $\forall M \in \M^{st}(O): M \models F$ }\\
\mbox{``no" }  \;\; \;\;\;\; \mbox{if $\FVar(F)=\emptyset$ and $\exists M \in \M^{st}(O): M \not \models F$ }\\
\{v:\FVar(F) \rightarrow V_O  \; |\;\;
\forall M \in \M^{st}(O), \; M \models v(F) \} & \mbox{ if $\FVar(F)\not=\emptyset$, }
\end{array} \right. \]
}
\noindent
 where $v(F)$ is the
formula $F$ after replacing all the free variables $x$ in $F$ by
$v(x)$.
 $\Box$}
\end{definition}

For example, let $p,q,c,s,o \in \URI$, let $G=\{p(s,o),
\rdf\?type(s,c), \rdf\?type(o,$ $c)\}$, and let
$P=$ $\{q(?x,?y) \leftarrow \rdf\?type(?x,c) \; \AND \;
\rdf\?type(?y,c) \; \AND \; \wneg p(?x,?y)\}$. Then, the stable
answers of $F=q(?x,?y)$ w.r.t. $O=\<G,P\>$ are $\Ans^{st}_O(F)= \{\{?x=o,
?y=o\}$, \{$?x=s, ?y=s\}$, $\{?x=o, ?y=s$\}\}.

Let $O=\<G,P\>$, where $q,s,o \in \URI$,  $G=\{\rdf\?type(p, \erdf\?\mathit{TotalProperty}), q(s,o)\}$, and
$P=\{ \sneg p(?x,?y) \leftarrow $ $\wneg p(?x,?y)\}$.
Then, it holds $\Ans^{st}_O(p(?x,?y))$= $\Ans^{st}_O(\wneg p(?x,?y))=$
$\Ans^{st}_O(\sneg p(?x,?y))=\emptyset$. This is because, in
contrast to the above example, $p$ is a total property. Thus,
for all mappings $v:\{?x,?y\} \rightarrow V_O$, there is a stable model $M$ of $O$ such that $M \models v(p(?x,?y)
\; \AND \wneg \ \sneg p(?x,?y))$, and another stable model $M'$ of
$O$ such that $M' \models v( \wneg p(?x,?y) \; \AND \; \sneg
p(?x,?y))$.

Consider the ERDF ontology $O$ of the paper assignment example, below Definition \ref{def:stableModel}.
Then, $\Ans^{st}_O(\mathit{assign}(P1,R2))=$``yes" and $\Ans^{st}_O(\mathit{assign}(P2,R1))$
=``no".
Though $\Ans^{st}_O(\mathit{assign}(P2,R1))=$``no", that is $\mathit{assign}(P2,R1)$ is not satisfied by
all stable models of $O$, there is a stable model ($M_3$) that satisfies $\mathit{assign}(P2,R1)$.
Indeed the answers of the query $\mathit{assign}(?x,?y)$ w.r.t. the stable models $M_3$ and $M_4$ are
of particular interest since both $M_3$ and $M_4$ satisfy $\mathit{allAssigned(Paper,Reviewer)}$,
indicating that the desirable paper assignment has been achieved.

The following definition defines the credulous stable answers of a query $F$ w.r.t. an ERDF ontology $O$,
that is the answers of $F$ w.r.t. the particular stable models of $O$.

\begin{definition} [Credulous ERDF stable answers] \label{credulousStableAnswers}
{\em Let $O=\<G,P\>$ be an ERDF ontology. The {\em credulous (ERDF) stable answers} of a query $F$ w.r.t.  $O$ are
defined as follows:

{\small
\[ \c-\Ans^{st}_{O}(F)= \left\{ \begin{array}{ll}
\mbox{``yes"}  \;\;\;\; \mbox{if $\FVar(F)=\emptyset$ and $\exists M \in \M^{st}(O): M \models F$ }\\
\mbox{``no" }  \;\;\;\; \mbox{if $\FVar(F)=\emptyset$ and $\forall M \in \M^{st}(O): M \not \models F$ }\\
\{\ans_M(F) \not = \emptyset \;\;|\;\;  M \in \M^{st}(O) \} \;\;\;\;\;\;\;\; \mbox{if $\FVar(F)\not=\emptyset$, }
\end{array} \right. \]
}
\noindent
 where $\ans_M(F)=\{v:\FVar(F) \rightarrow V_O  \; |\;
M \models v(F)\}$.
 $\Box$}
 \end{definition}

Continuing with the paper assignment example, consider the query:  
\begin{center}
$\mathit{F= allAssigned(Paper,Reviewer)}$. 
\end{center}
Then, although $\Ans^{st}_{O}(F)=$``no",
it holds $\c-\Ans^{st}_{O}(F)=$``yes", indicating that there is at least one
desirable assignment of the papers $P1,P2,P3$ to reviewers $R1,R2,R3$.

Consider now  the query
$F= \mathit{allAssigned(Paper,Reviewer) \; \AND \; assign(?x,?y)}$.
Then,
{\small
\begin{tabbing}
 aaaaaaaaaaaaa\= aaaaaaa \= aaaaa \= aaaaaa \=  \kill
$\c-\Ans^{st}_{O}(F)= \{$ \> $\{\{?x=P1, ?y=R2\}, \{?x=P2, ?y=R1\}, \{?x=P3, ?y=R3\}\},$\\
\> $\{\{?x=P1, ?y=R2\}, \{?x=P2, ?y=R3\}, \{?x=P3, ?y=R1\}\}\}$,
\end{tabbing}
}
\noindent
indicating all possible desirable assignments of papers. Obviously, the credulous stable answers of a query $F$ can provide alternative solutions,
which can be useful in a range of applications, where alternative scenarios naturally appear.

Closing this section, we would like to indicate several differences of the ERDF stable model semantics
w.r.t. first-order logic (FOL). First, in our semantics a {\em domain closure assumption } is made.
This is due to the fact that the domain of every Herbrand interpretation of an ERDF ontology $O$
is $Res^H_O$, that is the union of the vocabulary of $O$ ($V_O$)
and the set of XML values of the well-typed XML literals in  $V_O$ minus the well-typed XML literals.
This implies that quantified variables always range in a closed domain.
To understand the implications of this assumption, consider the ERDF graph:

{\small \[G=\{\rdf\?type(x,c1) \;\;|\;\;  x \in \{c1, c2\} \cup V'\},\]}
\noindent
where  $V'=(\V_{RDF} - \{\rdf\?\_i \;|\;i \in \NN\}) \cup \V_{RDFS} \cup \V_{ERDF}$.
Additionally, consider the ERDF program:

{\small
\begin{tabbing}
 aaaaaa\= aaaaaaa \= aaaaa \= aaaaaa \=  \kill
$P= \{$ \> $\rdf\?type(?x,c1) \leftarrow \rdf\?type(?x, \mathit{rdfs}\?\mathit{ContainerMembershipProperty})$.\\
\> $\rdf\?type(?x, c2) \leftarrow true.\}$.
 \end{tabbing}
}

\noindent
Let $F=\forall ?x \; \rdf\?type(?x,c2) \mImpl \rdf\?type(?x,c1)$.
It holds that $\<G,P\> \models^{st} F$.
However,  $G \cup P \not \models^{FOL} F$.
This is because, there is a FOL model $M$ of $G \cup P$ with a domain $D$ and
a variable assignment $v$:$\{?x\} \rightarrow D$ such that  $M,v \models \rdf\?type(?x,c2)$ and
$M,v \not \models \rdf\?type(?x,c1)$.

Another difference is due to the fact that in the definition of the ERDF stable model semantics,
only minimal Herbrand interpretations are considered. Let

{\small
\begin{center}
$G= \{teaches(Anne, \mathit{CS301}), teaches(\mathit{Peter}, \mathit{CS505}), \rdf\?type(\mathit{CS505}, \mathit{GradCourse})\}$.
\end{center}
}

Let $F=\forall ?x \; teaches(\mathit{Peter}, ?x) \mImpl \rdf\?type(?x, \mathit{GradCourse})$.
Then, $\<G,\emptyset\> $ $\models^{st} F$.  However, $G \not \models^{FOL} F$.
This is because, there is a FOL model $M$ of $G$ with a domain $D$ and
a variable assignment $v$:$\{?x\} \rightarrow D$ such that  $M,v \models teaches(\mathit{Peter}, ?x)$
and  $M,v \not \models \rdf\?type(?x,\mathit{GradCourse})$.
In other words, FOL makes an open-world assumption for {\em teaches}.

Consider now $G'=G \cup \{\rdf\?type(teaches, \erdf\?\mathit{TotalProperty})\}$.
Then, similarly to FOL, it holds $O=\<G',\emptyset\> \not \models^{st} F$.
This is because now $teaches$ is a total property. Thus, there is a stable model $M$ of $O$  and
a variable assignment $v$: $\{?x\} \rightarrow Res^H_O$ such that  $M,v \models teaches(\mathit{Peter}, ?x)$
and  $M,v \not \models \rdf\?type(?x,\mathit{GradCourse})$. In other worlds, now an open-world
assumption is made for $teaches$, as in FOL. Thus, there might exist a course taught by $\mathit{Peter}$,
even if it is not explicitly indicated so in $G'$.

This example also shows that, in contrast to FOL, stable model entailment is non-monotonic.

Note that the previous ERDF graph $G$ can also be seen as a Description Logic   A-Box $A$ \cite{DLhandbook},
where
\[A=\{teaches(Anne, CS301),  teaches(\mathit{Peter}, CS505), \mathit{GradCourse}(CS505)\}\]

Consider a T-Box $T=\emptyset$.
Since Description Logics (DLs) are fragments of first-order logic, it holds
that $L=\<A,T\> \not \models^{DL} \forall teaches.\mathit{GradCourse}(\mathit{Peter})$,
meaning that $L$ does not satisfy that all courses taught by Peter are graduate courses.
An interesting approach for supporting non-monotonic conclusions in DLs
is taken by Donini et al. \citeyear{DNR02}, where {\em DLs of minimal knowledge
and negation as failure} (MKNF-DLs) are defined, by extending DLs with two modal operators {\bf K, A}.
Intuitively, {\bf K} expresses minimal knowledge and
$\sneg${\bf A} expresses weak negation. It holds that
{\small $L \models^{\mbox{\tiny MKNF-DL}} \forall \mbox{{\bf K}}teaches.\mbox{{\bf K}}\mathit{GradCourse}(\mathit{Peter})$},
expressing that all courses known to be taught by Peter are known to be graduate
courses. Note that this conclusion is non-monotonic, and thus it cannot be derived by ``classical"
DLs. However, compared to our theory, MKNF-DLs do not support rules and closed-world
assumptions on properties (i.e., $\sneg p(?x,?y) \leftarrow \wneg p(?x,?y)$).

\section{An XML-based Syntax for ERDF} 
\label{sec:ERDF/XMLsyntax}

A natural approach to define an XML syntax for ERDF is: (i) to
follow the RDF/XML syntax \cite{rdfXml}, as much as possible, and (ii) to extend it in a
suitable way, where necessary. Following this approach, we briefly present here an XML syntax for ERDF.
Details are going to be given in a subsequent paper.

Classes and properties are defined with the help of the
\texttt{rdfs:Class} and \texttt{rdf:Property} elements of the RDF/XML syntax. Similarly, total classes and total properties are
defined with the help of the \texttt{erdf:TotalClass} and \texttt{erdf:TotalProperty} elements of the ERDF/XML syntax.

\begin{example} {\em 
The following ERDF/XML statements:
{\small
\begin{verbatim}
 <rdf:Property rdf:about="#likes">
   <rdfs:domain rdf:resource="#Person"/>
 </rdf:Property>

 <erdf:TotalProperty rdf:about="#authorOf">
   <rdfs:domain rdf:resource="#Person"/>
   <rdfs:range rdf:resource="#Book"/>
 </erdf:TotalProperty>
\end{verbatim}
}

\noindent
correspond to the ERDF graph:
{\small
\begin{tabbing}
aaaaaa\= aaaaaaa \= aaaaa \= aaaaaa \=  \kill
  $G= \{$ \> $\mathit{rdf\?type(likes, rdf\?Property), rdfs\?domain(likes, Person)},$ \\
  \> $ \mathit{rdf\?type(authorOf, erdf\?TotalProperty), rdfs\?domain(authorOf, Person)},$\\
  \> $\mathit{rdfs\?range(authorOf, Book)}\}$.
\end{tabbing}
}
}
\end{example}

ERDF triples (and sets of ERDF triples sharing the same subject term) are
encoded by means of the \texttt{erdf:Description} element. Each
description contains a non-empty list of (possibly negated)
property-value slots {\em about} the subject term.

\begin{itemize}

\item URI references, blank node identifiers, and variables that 
appear in the {\em subject} position of an ERDF triple are expressed as values of the
\texttt{erdf:about} attribute, using the SPARQL syntax \cite{SPARQL} for blank node
identifiers and variables. On the other hand, literals that appear in the subject position 
of an ERDF triple are expressed as the text content of the \texttt{erdf:about} subelement.

\item URI references, blank node identifiers,  and variables that  appear 
in the {\em object} position of an ERDF triple are expressed as values of the 
attributes \texttt{rdf:resource},
\texttt{rdf:nodeID}, and \texttt{erdf:variable}, respectively.
On the other hand, literals that appear in the object position 
of an ERDF triple are expressed as the text content of the corresponding property subelement.

\end{itemize}

\begin{example}

{\em The following \texttt{erdf:Description} statements:
{\small
\begin{verbatim}
<erdf:Description erdf:about="#Gerd">
  <ex:authorOf rdf:nodeID="x"/>
  <ex:likes rdf:resource="#Chicken"/>
  <ex:likes erdf:negationMode="Sneg" rdf:resource="#Pork"/>
</erdf:Description>

<erdf:Description>
  <erdf:About rdf:datatype="&xsd;string">Grigoris</erdf:About>
  <ex:denotationOf rdf:resource="#Grigoris"/>
</erdf:Description>
\end{verbatim}
}

\noindent
correspond to the ERDF graph:
{\small
\begin{tabbing}
aaaaaa\= aaaaaaa \= aaaaa \= aaaaaa \=  \kill
  $G= \{$ \> $\mathit{authorOf(Gerd, ?x), likes(Gerd, Chicken), \sneg likes(Gerd, Pork)}, $\\
  \> $\mathit{denotationOf(``Grigoris"\h\h xsd\?string, Grigoris)}\; \}$.
\end{tabbing}
}
}
\end{example}

Now, in order to express ERDF rules with XML, we use the rule markup
language R2ML ({\em REWERSE Rule Markup Language}) \cite{RoW2006:Wagner,WagnerGL05}, which is a general XML-based markup language
for representing derivation rules and integrity constraints. This is demonstrated in the following example:

\begin{example}
{\em The following \texttt{erdf:DerivationRule} statement:

{\small
\begin{verbatim}
<r2ml:DerivationRule r2ml:ruleID="R1">
  <r2ml:conditions>
    <erdf:Description erdf:about="?x">
      <rdf:type rdf:resource="#MainDish"/>
    </erdf:Description>
    <erdf:Description erdf:about="?y">
      <rdf:type rdf:resource="#Guest"/>
      <ex:likes erdf:variable="x"/>
    </erdf:Description>
    <r2ml:NegationAsFailure>
      <r2ml:ExistentiallyQuantifiedFormula>
        <r2ml:GenericVariable r2ml:name="z" r2ml:class="#Guest"/>
        <erdf:Description erdf:about="?z">
          <ex:likes erdf:negationMode="Sneg" erdf:variable="x"/>
        </erdf:Description>
      </r2ml:ExistentiallyQuantifiedFormula>
    </r2ml:NegationAsFailure>
  </r2ml:conditions>
  <r2ml:conclusion>
    <erdf:Description erdf:about="?x">
      <rdf:type rdf:resource="#SelectedMainDish"/>
    </erdf:Description>
  </r2ml:conclusion>
</r2ml:DerivationRule>
\end{verbatim}
}

\noindent
expresses that a main dish is selected for dinner, if there is a guest who likes it and no guest who dislikes it.
Specifically, it corresponds to the ERDF rule:
{\small
\begin{tabbing}
aaaaaaaaaaaaa\= aaa \= aaaaaaaaaa \= aa \=  \kill
$\rdf\?type(?x, \mathit{SelectedMainDish})$ \> \> \> $\leftarrow \rdf\?type(?x,\mathit{MainDish}), \rdf\?type(?y,\mathit{Guest}), likes(?y,?x),$\\
\> \> \>  \> $ \wneg \; (\exists ?z \; \rdf\?type(?z,\mathit{Guest}), \sneg likes(?z,?x))$.
 \end{tabbing}
}

}
\end{example}

\section{Undecidability of the ERDF Stable Model Semantics} \label{sec:Complexity}

The main difficulty in the computation of the ERDF stable model semantics is the fact that
$\V_{RDF}$ is infinite, and thus the vocabulary of any ERDF ontology $O$ is
also infinite (note that $\{\rdf\?\_i \;|\; i \in \NN \} \subseteq \V_{RDF} \subseteq V_O$). 
Due to this fact, satisfiability and entailment under the ERDF stable model semantics are in
general undecidable.

The proof of undecidability exploits a reduction from the {\em unbounded tiling problem}.
The unbounded tiling problem consists in placing tiles on an infinite grid, 
satisfying a given set of constraints on adjacent tiles. Specifically, the
unbounded tiling problem is a structure $\D=\<\T, H,V\>$, where $\T=\{T_1, ..., T_n\}$ 
is a finite set of tile types and $H$, $V \subseteq \T \times \T$ specify 
which tiles can be adjacent horizontally and vertically, respectively.
A {\em solution} to $\D$ is a {\em tiling}, that is, a total function $\tau: \NN \times \NN \rightarrow \T$
such that: $(\tau(i, j), \tau(i+1,j)) \in H$ and
$(\tau(i, j), \tau(i,j+1)) \in V$, for all $i,j \in \NN$.
The existence of a solution for a given unbounded tiling problem is known to be undecidable \cite{Berger66}.

Let $\D=\<\T, H, V\>$ be an  instance of the unbounded tiling problem, where $\T=\{T_1, ..., T_n\}$. 
We will construct an ERDF ontology $O_{\D}=\<G,P\>$ and an ERDF formula $F_{\D}$ such that
$\D$ has a solution iff $O_{\D}$ does not entail $F_{\D}$ under the ERDF stable model semantics.

Consider (i) a class $\mathit{Tile}$ whose instances are the tiles placed on the infinite grid, 
(ii) a property $\mathit{right(x,y)}$ indicating that tile $y$ is right next
to tile $x$, (iii) a property $\mathit{above(x,y)}$ indicating that tile $y$ is exactly 
above tile $x$, (iv) a class $\mathit{HasRight}$ whose instances are the tiles for which there exists
a tile right next to them, (v) 
a class $\mathit{HasAbove}$ whose instances are the tiles for which there exists
a tile exactly above them, (vi) a property $\mathit{Type}(x,T)$,
indicating that the type of tile $x$ is $T$, (vii) a property $\mathit{HConstraint}(T,T')$, indicating that 
$(T, T') \in H$, and (viii) a property $\mathit{VConstraint}(T,T')$, indicating that $(T, T') \in V$.

\godown
Let $G$ be the ERDF graph:
{\small
\begin{tabbing}
 aaaaaa\= a\=  \kill

$G=$ \> $\{$ \> $\rdfs\?\mathit{subClassOf}(\rdfs\?\mathit{ContainerMembershipProperty}, \mathit{Tile})$,\\
\>   \> $\rdfs\?\mathit{subClassOf}(\mathit{Tile}, \rdfs\?\mathit{ContainerMembershipProperty}) \} \; \cup$\\
\> $\{$ \> $ \mathit{HConstraint}(T, T') \; |\; (T, T') \in H\} \; \cup \; \{\mathit{VConstraint}(T, T') \; |\; (T, T') \in V\}$ .
 \end{tabbing}
}

Let $P$ be the ERDF program, containing the following rules (and constraints):

{\small
\begin{tabbing}
 aaaa \= aaaaaaaaaaaaaaa\= aaaa \= a \= aaaaa \= aaaaaa \=  \kill

(1) \> $\mathit{Type}(?x, T_1) \; \leftarrow$ \> $\rdf\?\mathit{type}(?x, \mathit{Tile}), \wneg \mathit{Type}(?x, T_2), ..., \wneg \mathit{Type}(?x, T_{n}).$ \\\\

\> $\mathit{Type}(?x, T_i) \; \leftarrow$ \> $\rdf\?\mathit{type}(?x, \mathit{Tile}), \wneg \mathit{Type}(?x, T_1), ..., \wneg \mathit{Type}(?x, T_{i-1}),$ \\
\> \>  $\wneg \mathit{Type}(?x, T_{i+1}), ..., \wneg \mathit{Type}(?x, T_n),$ for all $i=2, ..., n-1$.\\\\

\> $\mathit{Type}(?x, T_n) \; \leftarrow$ \> $\rdf\?\mathit{type}(?x, \mathit{Tile}), \wneg \mathit{Type}(?x, T_1), ..., \wneg \mathit{Type}(?x, T_{n-1}).$ \\\\

 
(2) \> $\mathit{right}(?x, ?y) $ \> $\leftarrow$ \> 
                $\rdf\?\mathit{type}(?x, \mathit{Tile}), \rdf\?\mathit{type}(?y, \mathit{Tile}),  \wneg \sneg \mathit{right}(?x, ?y)$.\\
     \> $\sneg \mathit{right}(?x, ?y)$ \> $ \leftarrow$ \> 
                $\rdf\?\mathit{type}(?x, \mathit{Tile}), \rdf\?\mathit{type}(?y, \mathit{Tile}),  \wneg \mathit{right}(?x, ?y)$.\\\\
                
(3) \> $\mathit{above}(?x, ?y) $ \> $\leftarrow$ \> 
                $\rdf\?\mathit{type}(?x, \mathit{Tile}), \rdf\?\mathit{type}(?y, \mathit{Tile}),  \wneg \sneg \mathit{above}(?x, ?y)$.\\
     \> $\sneg \mathit{above}(?x, ?y)$ \> $ \leftarrow$ \> 
                $\rdf\?\mathit{type}(?x, \mathit{Tile}), \rdf\?\mathit{type}(?y, \mathit{Tile}),  \wneg \mathit{above}(?x, ?y)$.\\\\
                
(4) \> $\rdf\?\mathit{type}(?x, \mathit{HasRight}) $ \> \> \> $\leftarrow \; \mathit{right}(?x, ?y)$.\\
     \> $\rdf\?\mathit{type}(?x, \mathit{HasAbove}) $ \> \> \> $\leftarrow \; \mathit{above}(?x, ?y)$.\\\\

\> $\mathit{false} \; \leftarrow \; \rdf\?\mathit{type}(?x, \mathit{Tile}), \wneg \rdf\?\mathit{type}(?x, \mathit{HasRight}).$\\
     \> $\mathit{false} \; \leftarrow \; \rdf\?\mathit{type}(?x, \mathit{Tile}), \wneg \rdf\?\mathit{type}(?x, \mathit{HasAbove}).$\\\\ 
     
     \> $\mathit{id}(?x, ?x) \; \leftarrow \; \rdf\?\mathit{type}(?x, \rdfs\?\mathit{Resource})$.\\\\            
 
\> $\mathit{false} \; \leftarrow \; \mathit{right}(?x, ?y), \mathit{right}(?x, ?y'), \wneg \mathit{id}(?y, ?y').$\\
     \> $\mathit{false} \; \leftarrow \; \mathit{above}(?x, ?y), \mathit{above}(?x, ?y'), \wneg \mathit{id}(?y, ?y').$\\\\\\

(5) \> $\mathit{false} \; \leftarrow \; \mathit{right}(?x, ?y), \mathit{Type}(?x, ?T), \mathit{Type}(?y, ?T'), \wneg \mathit{HConstraint}(?T, ?T').$\\
     \> $\mathit{false} \; \leftarrow \; \mathit{above}(?x, ?y), \mathit{Type}(?x, ?T), \mathit{Type}(?y, ?T'), \wneg \mathit{VConstraint}(?T, ?T').$
          \end{tabbing}
}

Note that in all stable models of $O_{\D}=\<G,P\>$, the class $\mathit{Tile}$ contains exactly the (infinite in  mumber) $\rdf\?\_i$ terms, for $i \in \NN$. 
This is because, computing the stable models of $O$,   only the minimal models of $sk(G)$ are considered  (see Definition \ref{def:stableModel}, Step 1).
Thus, each tile on the infinite grid is represented by an $\rdf\?\_i$ term, for $i \in \NN$.

Intuitively, rule set (1) expresses that each tile should have exactly one associated type in $\T$.
Rule set (2) expresses that two tiles are either horizontally adjacent on the grid or not horizontally adjacent.
Rule set (3) expresses that two tiles are either vertically adjacent on the grid or not vertically adjacent.
Rule set (4) expresses that each tile should have exactly one tile right next to it and exactly one tile right above it.
Rule set (5) expresses that the types of horizontally and vertically adjacent tiles should respect the $H$ and $V$
relations of $\D$, respectively.

To finalize the reduction, we define: 
{\small
\begin{tabbing}
 aaaaa\= aaaaaaaaaaaaaaaaaaaaaaa\= a \= aa \= aaaaaa \=  \kill

$F_{\D}=$ \> $\exists ?x, \exists ?y, \exists ?x', \exists ?y', \exists ?x''$ \> $\mathit{right}(?x, ?y) \; \AND \; \mathit{above}(?y, ?y') \; \AND \; \mathit{right}(?x', ?y') \; \AND \; \mathit{above}(?x'', ?x') \; \AND$\\
\> \>  $\wneg  \mathit{id}(?x, ?x'') $.
 \end{tabbing}
}

\noindent
Formula $F_{\D}$ expresses that there  is a tile $x$ such that, starting from $x$, if we move:
{\small
\begin{center}
 {\em one step right} $\rightarrow$ {\em one step up} $\rightarrow$ {\em one step left} $\rightarrow$ {\em one step down} 
\end{center}
}
\noindent
then we will meet a tile $x''$ different than $x$.

\begin{proposition} \label{Prop:TilingReduction}
{\em Let $\D$ be an  instance of the unbounded tiling problem. It holds:
\begin{enumerate}

\item $\D$ has a solution iff $O_{\D} \cup \{\mathit{false} \; \leftarrow \; F_{\D}\}$ has a stable model.
\item $\D$ has a solution iff $O_{\D} \not \models^{st} F_{\D}$.
\end{enumerate}
}
\end{proposition}

\noindent 
Since the unbounded tiling problem is undecidable \cite{Berger66}, it follows directly from
Proposition \ref{Prop:TilingReduction} that satisfiability and entailment  under the ERDF stable model semantics are in general undecidable.

The previous reduction shows that both problems remain undecidable for an ERDF ontology $O=\<G, P\>$, even if (i)
the body of each rule in $P$ has the form $t_1, ..., t_k, \wneg t_{k+1}, ..., \wneg t_n$, where $t_i$
is an ERDF triple and (ii) 
the terms $\erdf\?\mathit{TotalClass}$ and
 $\erdf\?\mathit{TotalProperty}$ do not appear in $O$, that is, $(V_G \cup V_P) \cap \V_{\ERDF}=\emptyset$. 
Note that since each constraint $\mathit{false} \leftarrow F$ that
appears in an ERDF ontology $O$ can be replaced by the rule $\sneg t \leftarrow F$, 
where $t$ is an RDF, RDFS, or ERDF axiomatic triple, the presence of constraints in $O$ does not affect 
decidability.

Future work concerns  the identification of syntactic restrictions for an ERDF ontology
$O$ such that ERDF stable model entailment is decidable.

\section {ERDF Model Theory as Tarski-style Model Theory} \label{sec:ERDF-Tarski}

Tarski-style model theory is not limited to classical first-order
models, as employed in the semantics of OWL. It allows various
extensions, such as relaxing the bivalence assumption (e.g.,
allowing for partial models) or allowing higher-order models. It
is also compatible with the idea of non-monotonic inference, simply
by not considering all models of a rule as being intended, but
only those models that satisfy certain criteria. Thus, the
stable model semantics for normal and (generalized) extended logic
programs \cite{gelfond88:sms,gelfond90:asets,HW97,HJW96}
can be viewed as a Tarski-style model-theoretic semantics for
non-monotonic derivation rules.

A Tarski-style model theory is a triple $\langle \LL, \I,
\models \rangle$ such that:
\begin{itemize}
  \item $\LL$ is a set of formulas, called \emph{language},
  \item $\I$ is a set of interpretations, and
  \item $\models$ is a relation between interpretations and
  formulas, called \emph{model relation}.
\end{itemize}
For each Tarski-style model theory $\langle \LL, \I, \models
\rangle$, we can define:
\begin{itemize}
  \item a notion of {\em derivation rule} $G \leftarrow F$, where $F \in \LL$ is
called {\em condition} and $G \in \LL$ is called {\em conclusion},

  \item  a set of
  derivation rules  $\cal{DR}$$_{\LL} = \{ G \leftarrow F \;|\;  F,G \in \LL \}$,
  
  \item an extension of the model relation $\models$ to include also 
  pairs of interpretations and derivation rules, and

  \item a standard model operator $\M(\mathit{KB}) = \{ I \in \I \; | \; I \models X,  \; \forall X \in \mathit{KB} \}$,
  where $\mathit{KB} \subseteq \LL \cup \cal{DR}_{\LL}$ is a set of formulas
and/or derivation rules, called a \emph{knowledge base}.
\end{itemize}

Notice that in this way we can define rules also for logics which
do not contain an implication connective. This shows that the
concept of a rule is independent of the concept of implication.

Typically, in knowledge representation theories, not all models of
a knowledge base are \emph{intended} models. Except from the
standard model operator $\M$, there are also non-standard model
operators, which do not provide all models of a knowledge base,
but only a special subset that is supposed to capture its intended
models according to some semantics.

A particularly important type of such an ``intended model
semantics" is obtained on the basis of some \emph{information
ordering} $\leq$, which allows to compare the information content
of two interpretations $I_1,I_2 \in \I$. Whenever $I_1 \leq I_2$,
we say that $I_1$ \emph{is less informative} than $I_2$. An
\emph{information model theory} $\< \LL, \; \I, \; \models, \; \leq \>$ is a
Tarski-style model theory, extended by an information ordering $\leq$.

For any information model theory, we can define a number of
natural non-standard model operators, such as the \emph{minimal}
model operator:
\[
\M^{min}(\mathit{KB}) = minimal_{\leq}(\M(\mathit{KB}))
\]
and various refinements of it, like the {\em stable generated}
models \cite{gelfond88:sms,gelfond90:asets,HW97,HJW96}.

For any given model operator $\M^{x}: \POW(\LL \cup \cal{DR}_{\LL})
\rightarrow \POW(\I)$, knowledge base $\mathit{KB} \subseteq \LL \cup
\cal{DR}_{\LL}$, and $F \in \LL$, we can define an entailment relation:
\[
\mathit{KB} \models^{x} F \quad\mbox{iff}\quad \forall I \in
\M^x(\mathit{KB}), \;  I \models F
\]

For non-standard model operators, like minimal and stable models,
this entailment relation is typically \emph{non-monotonic}, in the
sense that for an extension $\mathit{KB}' \supseteq  \mathit{KB}$ it may be the case
that $\mathit{KB}$ entails $F$, but $\mathit{KB}'$ does not entail $F$.

Our (ERDF) stable model theory can be seen as a Tarski-style model
theory, where $\LL=L(\URI \cup \Lit)$, $\I$ is the set of
ERDF interpretations over any vocabulary $V \subseteq \URI \cup
\Lit$, and the model relation $\models$ is as defined in
Definitions \ref{def:modelRelation} and \ref{def:modelsRuleOntology}. In our theory, the intended
model operator ($\M^{st}$) assigns to each ERDF ontology a (possibly empty) set of stable models
(Definition \ref{def:stableModel}).

\section{Related Work} \label{sec:RelatedWork}

In this section, we briefly review extensions of web ontology languages with rules.

Ter Horst  \citeyear{HorstJWS05,Horst04} generalizes RDF graphs to {\em generalized RDF graphs}, by allowing variables
in the property position of RDF triples. Additionally, the author extends the RDFS semantics with datatypes and part of the OWL
vocabulary, defining the $pD*$ semantics, which extends the ``if-semantics" of RDFS and is weaker than the 
``iff-semantics" of D-entailment \cite{rdfsemantics} and OWL Full \cite{owlsemantics}.
A sound and complete set of entailment rules for $pD*$ entailment is also presented.

In a subsequent work, ter Horst  \citeyear{HorstISWC05} considers the extension of the previous framework with the inclusion of rules of
the form ``if $G$ then $G'$", where $G$ is an RDF graph without blank nodes but possibly with variables 
and $G'$  is a generalized RDF graph, possibly
with  both blank nodes and variables. Intuitively, rule variables are universally quantified in the front of the rule 
(like the free variables of our rules) and blank nodes in the head of the rule correspond to existentially quantified variables
(this feature is not supported in our model). Based on a set of rules $R$ and a datatype map $D$, 
$R$-entailment\footnote{The symbol $D$ does not appear explicitly in the notation of $R$-entailement, for reasons of simplification.} is defined between two 
generalized RDF graphs $G$ and $G'$ ($G \models_R G'$),
and a set of sound and complete rules for $R$-entailment is presented. To relate our work with that of ter Horst \citeyear{HorstISWC05},
we state the following proposition: 

\vspace{0.1cm}
{\small
\noindent
Let $D$ be a datatype map, containing only $\rdfXMLLiteral$,
and let $R$ be a set of rules of the form ``if $G$ then $G'$" with the constraints: (i)
all terms appearing in property position are URIs, (ii)
if $G \not = \{\}$ then no blank node appears in $G'$,
and (iii) $V_R \cap (\V_{\mathit{pOWL}} \cup \V_{\ERDF})=\emptyset$, where $\V_{\mathit{pOWL}}$ denotes the part of the OWL vocabulary, 
included in the $pD*$ semantics. Let $G,G'$ be RDF graphs such that $(V_G \cup V_{G'}) \cap (\V_{\mathit{pOWL}} \cup \V_{\ERDF})=\emptyset$.
Then based on $G$ and $R$, we can define, by a simple transformation, an ERDF ontology $O$ such that $G \models_R G'$ iff $O \models^{st} G'$.
}
\vspace{0.1cm}

\noindent 
However, in this work, weak and strong negation are not considered. Thus, closed-world reasoning
is not supported. Additionally, in our theory, the condition of a rule is any ERDF formula over a vocabulary $V$, 
(thus, involving any of the logical factors
$\wneg$, $\sneg$,  $\mImpl$, $\AND$, $\OR$, $\forall$, and $\exists$), and not just
a conjunction of positive triples.

TRIPLE \cite{SD02} is a rule language for the Semantic Web that is
especially designed for querying and transforming RDF models (or contexts),
supporting RDF and a subset of OWL Lite.
Its syntax is based on F-Logic \cite{KLW95} and supports an
important fragment of first-order logic. A triple is represented by
a statement of the form $s[p \rightarrow o]$ and sets of statements,
sharing the same subject $s$, can be aggregated using molecules of
the form $s[p_1 \rightarrow o_1; p_2 \rightarrow o_2; ....]$.
All variables must be
explicitly quantified, either existentially or universally.
Arbitrary formulas can be used in the body, while the head of the rules
is restricted to atoms or conjunctions of
molecules. An interesting and relevant feature of TRIPLE is the use
of models to collect sets of related sentences. In particular,
part of the semantics of the RDF(S) vocabulary is represented as pre-defined rules (and not as semantic conditions on interpretations), which are
grouped together in a module.
TRIPLE provides other features like path
expressions, skolem model terms, as well as model intersection and
difference. Finally, it should be mentioned that the queries and
models are compiled into XSB Prolog. TRIPLE uses the Lloyd-Topor transformations \cite{LlTo84} to take care of
the first-order connectives in the sentences and supports weak negation under the
well-founded semantics \cite{gelder90:wfs}. Strong negation is not used.

Flora-2 \cite{YKZ03} is a rule-based object-oriented knowledge base system for reasoning with semantic information on the Web.
It is based on F-logic \cite{KLW95} and supports metaprogramming, non-monotonic multiple inheritance, logical database updates, encapsulation, dynamic modules, and two kinds of weak negation. Specifically, it supports Prolog negation and well-founded negation \cite{gelder90:wfs}, through invocation of the corresponding operators $\setminus +$ and $tnot$ of the XSB system \cite{RSSWF97}.
The formal semantics for non-monotonic multiple inheritance is defined by Yang \& Kifer \citeyear{YaKi03}.
In addition, Flora-2 supports reification and anonymous resources \cite{YaKi03b}. In particular, in Flora-2,
reified statements $\$\{s(p  \rightarrow o)\}\$$ are themselves objects. In contrast, in RDF(S), they are referred to by
a URI or a blank node $x$, and are associated with the following RDF triples: $\rdf\?type(x, \rdf\?Statement)$,
$\rdf\?subject(x, s)$, $\rdf\?predicate(x, p)$, and $\rdf\?object(x, o)$. In RDF(S) model theory (and thus, in our theory), no special semantics are given to reified statements. In Flora-2, anonymous resources
are handled through skolemization (similarly to our theory).

Notation 3 (N3)  \cite{BCKSH08}  provides a more human readable syntax for RDF and also
extends RDF by adding numerous pre-defined constructs (``built-ins") for
being able to express rules conveniently. Remarkably, N3 contains
a built-in ({\tt log:definitiveDocument}) for making restricted
completeness assumptions and another built-in ({\tt log:notIncludes}) for
expressing simple negation-as-failure tests. The addition of these
constructs was motivated by use cases. However, N3 does not provide strong negation and closed-world reasoning
is not fully supported.
N3 is supported by the CWM system\footnote{http://www.w3.org/2000/10/swap/doc/cwm.html.}, a forward engine especially designed for the Semantic Web, and the Euler system\footnote{http://www.agfa.com/w3c/euler/.},
a backward engine relying on loop checking techniques to guarantee termination.

Alferes et al. \citeyear{ADP03} propose the paraconsistent well-founded semantics with explicit negation ($\mathit{WFSX_P}$)\footnote{$\mathit{WFSX_P}$ \cite{adp95:jarsi}
is an extension of the well-founded semantics with explicit negation (WFSX)  on extended logic programs \cite{pa92:wfsx} and, thus, also of the well-founded semantics (WFS)  on normal logic programs \cite{gelder90:wfs}.},
as the appropriate semantics for reasoning with (possibly, contradictory) information in the Semantic Web.
Supporting arguments include: (i) possible reasoning, even in the presence of contradiction,
(ii) program transformation into WFS, and (iii) polynomial time inference procedures.
No formal model theory has been explicitly
provided for the integrated logic.

DR-Prolog \cite{ABW04b} and DR-DEVICE \cite{BAV04} are two systems that integrate RDFS ontologies with rules (strict or defeasible),
that are partially ordered through a superiority relation,
based on the semantics of defeasible logic \cite{ABGM01,Maher02}. Defeasible logic contains
only one kind of negation (strong negation) in the object language\footnote{However, in defeasible logic,
negation-as-failure
can be easily simulated by other language ingredients.} and allows to reason in the presence of
contradiction and  incomplete information. It supports monotonic and non-monotonic rules,
exceptions, default inheritance, and preferences. No formal
model theory has been explicitly
provided for the integrated logic.

OWL-DL \cite{owloverview} is an ontology representation language for the Semantic Web, that is
a syntactic variant of the $\cal{SHOIN}$({\bf D}) description logic and a decidable fragment of first-order logic \cite{HoPaISWC03}.
However, the need for extending the expressive power of OWL-DL with rules has initiated several studies, including the SWRL (Semantic Web Rule Language) proposal \cite{SWRL}. Horrocks \& Patel-Schneider \citeyear{HP04}
show that this extension is in general undecidable.
$\cal{AL}$-log \cite{DLNS98} was one of the first efforts to integrate Description Logics with (safe) datalog rules, while achieving decidability.
It considers the basic description logic $\cal{ALC}$ and imposes  the constraint that only concept DL-atoms are allowed to appear in the body of the rules, whereas the heads of the rules are always non DL-atoms. Additionally, each variable appearing in a concept DL atom in the body of a rule has also to appear in a non DL-atom in the body or head of the rule.
CARIN \cite{LR98} provides a framework for studying the effects of combining the description logic $\cal{ALCNR}$ with (safe) datalog rules. In CARIN, both concept and role DL-atoms are allowed in the body of the rules. It is shown that
the integration is decidable if rules are non-recursive, or certain combinations of constructors are not allowed in the DL component, or rules are {\em role-safe} (imposing a constraint on the variables of role DL atoms in the body of the rules)\footnote{A rule is {\em role-safe} if at least one of the variables $x,y$ of each role DL atom $R(x,y)$ in the body of the rule, appears in some body atom of a {\em base} predicate, where a {\em base predicate} is an ordinary predicate that appears only in facts or in rule bodies.}.
Motik et al. \citeyear{MSS04} show that the integration of a $\cal{SHIQ}$({\bf D}) knowledge base $L$ with a disjunctive datalog program $P$ is decidable, if $P$ is DL-safe, that is, all variables in a rule occur in at least one non DL-atom in the body of the rule.
In this work, in contrast to $\cal{AL}$-log and CARIN, no tableaux algorithm is employed for query answering but $L$ is translated to a disjunctive logic program $\mathit{DD}(L)$ which is combined with $P$ for answering ground queries.

In this category of works, entailment on the  DL, that has been extended with rules, is based on first-order logic. This means that both the DL component and the logic program are viewed as a set of first-order logic statements. Thus, negation-as-failure, closed-world-assumptions, and non-monotonic reasoning cannot be supported. In contrast, our work supports both weak and strong negation, and allows closed-world and open-world reasoning on a selective basis.

A different kind of integration is achieved by Eiter et al. \citeyear{ELST04(a)}.
In this work, a $\cal{SHOIN}$({\bf D}) knowledge base $L$ communicates with an extended logic program $P$ (possibly with weak and strong negation), only through  DL-query atoms in the body of the rules. In particular, the description logic component $L$ is used for answering the augmented, with input from the logic program, queries appearing in the (possibly weakly negated) DL-query atoms, thus allowing flow of knowledge from $P$ to $L$ and vice-versa. The answer set semantics of $\<L,P\>$ are defined, as a generalization of the answer set semantics \cite{gelfond90:asets} on ordinary extended logic programs. A similar kind of integration is achieved by Eiter et al. \citeyear{ELST04(b)}.
In this work, a $\cal{SHOIN}$({\bf D}) knowledge base $L$ communicates with a normal logic program $P$ (possibly with weak negation), through DL-query atoms in the body of the rules. The well-founded semantics of $\<L,P\>$ are defined, as a generalization of the well-founded semantics  \cite{gelder90:wfs} of ordinary normal logic programs. 
Obviously, in both of these works, 
derived information concerns only non DL-atoms (that can be possibly used as input to DL-query atoms). 
Thus, rule-based reasoning is supported only for non DL-atoms. In contrast, in our work, properties and classes appearing in the 
ERDF graphs can freely appear in the heads and bodies of the rules, allowing even the derivation of metalevel statements such as subclass and subproperty relationships, property transitivity, property and class totalness.

Rosati \citeyear{Rosati99} defines the semantics of a {\em disjunctive} $\cal{AL}$-log knowledge base, based on the stable model semantics
for disjunctive databases \cite{GeLiNGC91},
extending $\cal{AL}$-log \cite{DLNS98}.
A disjunctive $\cal{AL}$-log knowledge base is the integration of an $\cal{ALC}$ knowledge base $T$ with a (safe) disjunctive logic program
$P$ that allows concept and role DL-atoms in the body of the rules (along with weak negation on non DL-atoms). The safety condition
enforces that each variable in the head of a rule should also appear in the body of the rule. 
Additionally, all constants in $P$ should be DL-individuals. Similarly to our case,
in defining the disjunctive $\cal{AL}$-log semantics, only the grounded versions of the rules are considered
(by instantiating variables with DL individuals). However
rule-based reasoning is supported only for non DL-atoms,
and DL-atoms in the body of the rules mainly express constraints.

In a subsequent work, Rosati \citeyear{RosatiJWS05} defines the $r$-hybrid knowledge bases.
In $r$-hybrid knowledge bases, DL-atoms are allowed in the head of the rules and the DL component $T$
is a $\cal{SHOIN}$({\bf D}) knowledge base. Additionally, constants in $P$ are not necessarily DL-individuals.
However, a stronger safety condition is imposed, as each rule variable should appear in a (positive) non DL-atom in the body of the rule. Additionally, weak negation is allowed only for non DL-atoms and rule-based meta-reasoning is not supported. In general, we can say that for non DL-atoms, a closed-world assumption is made, while 
DL-atoms conform to the open-world assumption, as $\cal{SHOIN}$({\bf D}) is a fragment of first-order logic.

\section{Conclusions} \label{sec:Conclusions}

In this paper, we have extended RDF graphs to ERDF graphs by allowing negative triples for representing explicit negative information.
Then, we proceeded by defining an ERDF ontology as an ERDF graph complemented by a set 
of derivation rules with all connectives $\wneg$ (weak negation), $\sneg$ (strong negation),  $\mImpl$ (material implication), $\AND$, $\OR$, $\forall$,
$\exists$ in the body of a rule, and with strong negation $\sneg$ in the head of a rule.
Moreover, we have extended the RDF(S) vocabulary by adding the predefined vocabulary elements 
$\erdf\?\mathit{TotalProperty}$ and $\erdf\?\mathit{TotalClass}$, for 
representing the metaclasses of total properties and total classes, 
on which the open-world assumption applies.

We have defined ERDF formulas, ERDF interpretations, and ERDF entailment on ERDF formulas,
showing that it conservatively extends RDFS entailment on RDF graphs.
We have developed the model-theoretic semantics of ERDF ontologies, called {\em ERDF stable model semantics}, showing that stable model entailment extends ERDF entailment on ERDF graphs, and thus it also extends
RDFS entailment on RDF graphs. The ERDF stable model semantics is based on Partial Logic and, in particular, on
its generalized definition of stable models \cite{HW97,HJW96} (which extends answer set semantics on extended logic programs). We have shown
that classical (boolean) Herbrand model reasoning is a special case of our semantics, when all properties are total. In this case, similarly to classical logic, an open-world assumption is made for all properties and classes
and the two negations (weak and strong negation) collapse. Allowing (a) the totality of properties and classes to be declared on a selective basis and (b) the explicit representation of  closed-world assumptions (as derivation rules) enables the combination of open-world and closed-world reasoning in the same framework.

In particular, for a total property $p$, the open-world assumption applies, since each considered Herbrand 
interpretation $I$, in the computation of ERDF stable models, satisfies $p(x,y) \OR \sneg p(x,y)$, for each pair 
$(x,y)$ of ontology vocabulary terms. For a closed property $p$, a default closure rule
of the form $\sneg p(?x,?y) \leftarrow \wneg p(?x,?y)$ is added, which allows to infer the falsity of 
$p(x,y)$, if there is no evidence that $p(x,y)$ holds.
However, this method only works for partial properties. For a total property $p$, it may happen that 
there is a stable model, where $p(x,y)$ holds, even though there is no evidence for it 
(see the example in Section \ref{sec:ERDF_stableModel}, above 
Proposition \ref{prop:stableImpliesHerbrand}). In fact, if $p$ is a total property, the existence or not of the corresponding default closure rule does not affect the ontology semantics.

\godown
The main advantages of ERDF are summarized as follows:

\begin{itemize}

\item It has a Tarski-style model theory, which is a desirable feature for logic languages for the Semantic Web \cite{BryM05}.

\item It is based on Partial Logic \cite{HJW96}, which is the simplest conservative extension of classical
logic that supports both weak and strong negation.
Partial logic also extends Answer Set Programming (ASP)\footnote{ASP is a well-known and accepted knowledge representation formalism
that allows (through credulous reasoning) the definition of concepts ranging over a space of choices.  
This feature enables the compact representation of search and optimization problems \cite{EiterIPS06}.} \cite{gelfond90:asets}, by allowing all logical factors 
$\wneg$, $\sneg$, $\mImpl$, $\AND$, $\OR$, $\forall$,
$\exists$ in the body of a rule. 

\item It enables the combination of open-world (monotonic) and closed-world (non-monotonic) reasoning, 
in the same framework. 

\item It extends RDFS ontologies with derivation rules and integrity constraints.
\end{itemize}
Satisfiability and entailment under the ERDF stable model semantics are in general undecidable.
In a subsequent paper, we plan to  identify syntactic restrictions for the ERDF ontologies that guarantee decidability of reasoning
and to elaborate on the ERDF computability and complexity issues.

 In this work, we consider only coherent ERDF interpretations. However, 
due to the Semantic Web's decentralized and distributed nature, 
contradictory information is frequent \cite{SchaffertBBDDEH05}. Though Partial Logic
allows for truth-value clashes,  
handling inconsistency
in the Semantic Web is a topic that deserves extended treatment, which is 
outside the scope of this paper.
It is in our future plans to consider general ERDF interpretations and 
extend the vocabulary of ERDF with the terms
$\mathit{\erdf\?CoherentProperty}$ and $\mathit{\erdf\?CoherentClass}$, whose
instances are properties and classes that satisfy {\em coherence}.
Thus, coherence will be decided on a per property and per class basis. 
Admitting incoherent models will only be
interesting in combination with a second preference criterion
of ``minimal incoherence" \cite{HJW96}.

Our future work also concerns the support of datatypes, including XSD datatypes, and the extension of the 
predefined ERDF vocabulary by adding other useful constructs, possibly in accordance with the extensions of ter Horst \citeyear{HorstJWS05}.
We also plan to formally define the ERDF/XML syntax, briefly presented in Section \ref{sec:ERDF/XMLsyntax}.
Moreover, we plan
to implement an ERDF inference engine.

Finally, we would like to mention that the success of
the Semantic Web is impossible without support for modularity, encapsulation,
information hiding, and access control.  
Modularity mechanisms and syntactic restrictions for merging knowledge bases in the Semantic
Web are explored by Dam{\'a}sio et al. \citeyear{DAAW_PPSWR06}.
However, in this work, knowledge bases are expressed by extended logic programs.
Our future plans include the extension of ERDF with mechanisms
allowing sharing of knowledge between different ERDF ontologies, 
along the lines proposed  by Dam{\'a}sio et al. \citeyear{DAAW_PPSWR06}.

\acks{The authors would like to thank the reviewers for their valuable comments.
This research has been partially funded by the
European Commission and by the Swiss Federal Office for Education
and Science within the 6th~Framework Programme project REWERSE
num.~506779 (www.rewerse.net).}

\appendix
\section*{Appendix A: RDF(S) Semantics}

  \setcounter{section}{1}
For self-containment, in this Appendix, we review the definitions of simple, RDF, and RDFS interpretations, as
well as the definitions of satisfaction of an RDF graph and RDFS entailment. For details, see the W3C Recommendation of RDF semantics \cite{rdfsemantics}.

Let $\URI$ denote the set of URI references,
$\PL$ denote the set of plain literals, and $\TL$ denote the set of
typed literals, respectively. A vocabulary $V$ is a subset of $\URI \cup \PL \cup \TL$.
The vocabulary of RDF, $\V_{RDF}$, and the vocabulary of RDFS,
$\V_{RDFS}$, are shown in Table
\ref{table:Vocabulary_RDF_RDFS} (Section \ref{sec:ERDF_interpretations}).

\begin{definition} [Simple interpretation]
{\em A {\em simple interpretation} $I$ of a vocabulary $V$
consists of:

{\small
 \begin{itemize}
\item A non-empty set of resources $Res_I$, called the {\em domain} or {\em universe} of $I$.

\item A set of properties $Prop_I$.

\item A vocabulary interpretation mapping $I_V$$:V \cap \URI \rightarrow
Res_I \cup Prop_I$.

\item A property extension mapping $\PT_I: Prop_I \rightarrow
\POW(Res_I \times Res_I)$.

\item A mapping $\IL_I: V \cap \TL \rightarrow Res_I$.

\item A set of literal values $\LV_I \subseteq Res_I$, which contains
$V \cap \PL$.
\end{itemize}
}

\noindent  We define the mapping: $\;I: V \rightarrow Res_I \cup Prop_I$ such that:
{\small
 \begin{itemize}
 \item $I(x)=I_V(x)$,  $\; \forall x \in V \cap \URI$.
\item $I(x)=x$, $\; \forall \; x \in V \cap \PL.$

 \item $I(x)=\IL_I(x)$, $\; \forall \; x \in V \cap \TL.$ $\Box$
 \end{itemize}
 }
 }
\end{definition}

\begin{definition} [Satisfaction of an RDF graph w.r.t. a simple interpretation]
\label{def:RDFmodelRelation} {\em Let $G$ be an RDF graph and
let $I$ be a simple interpretation of a vocabulary $V$. Let $v$ be a mapping $v:\Var(G) \rightarrow Res_I$.
If $x \in  \Var(G)$, we define $[I+v](x)=v(x)$. If $x \in V$, we define $[I+v](x)=I(x)$.
We define:
\begin{itemize}
\item $I, v \models G$ iff $\forall \; p(s,o) \in G$, it holds that: $p \in V, \; s,o \in V \cup \Var, \; I(p) \in Prop_I$, and
$\<[I+v](s),[I+v](o)\> \in \PT_I(I(p))$.

\item $I$ {\em satisfies} the RDF graph $G$, denoted by $I \models G$, iff there exists a mapping
$v:\Var(G) \rightarrow Res_I$ such that $I, v \models G$. $\Box$
\end{itemize}}
\end{definition}

\puttableSV{table:RDFaxiomaticTriples}{The RDF axiomatic triples}{}{\small { \btbl{l}
$\rdf\?type(\rdf\?type, \rdf\?\mathit{Property})$ \\
$\rdf\?type(\rdf\?subject, \rdf\?\mathit{Property})$ \\
$\rdf\?type(\rdf\?predicate, \rdf\?\mathit{Property})$ \\
$\rdf\?type(\rdf\?object, \rdf\?\mathit{Property})$ \\
$\rdf\?type(\rdf\?first, \rdf\?\mathit{Property})$ \\
$\rdf\?type(\rdf\?rest, \rdf\?\mathit{Property})$ \\
$\rdf\?type(\rdf\?value, \rdf\?\mathit{Property})$ \\
$\rdf\?type(\rdf\?\_i, \rdf\?\mathit{Property})$, $\;\;\forall i \in \{1,2,...\}$ \\
$\rdf\?type(\rdf\?nil, \rdf\?List)$ 
 \etbl }}

 \begin{definition} [RDF interpretation]  \label{def:RDFInterpretation}
 {\em An {\em RDF interpretation} $I$ of a
vocabulary $V$ is a simple interpretation of $V \cup \V_{RDF}$, which satisfies the following
semantic conditions:

{\small
\begin{enumerate}
\item $x \in Prop_I$ iff $\<x, I(\rdf\?\mathit{Property})\> \in \PT_I(I(\rdf\?type))$.

 \item If $``s"\h\h \rdfXMLLiteral \in V$  and $s$ is a
well-typed XML literal string, then\\
 $\IL_I$($``s"\h\h \rdfXMLLiteral$) is the XML value of $s$, \\
 $\IL_I(``s"\h\h \rdfXMLLiteral)\in \LV_I$, and\\
 $\<\IL_I(``s"\h\h \rdfXMLLiteral), I(\rdfXMLLiteral)\> \in
 \PT_I(I(\rdf\?type))$.

 \item If $``s"\h\h \rdfXMLLiteral \in V$  and $s$ is an
ill-typed XML literal string then\\
 $\IL_I(``s"\h\h \rdfXMLLiteral)\in Res_I-\LV_I$, and\\
$\<\IL_I(``s"\h\h \rdfXMLLiteral), I(\rdfXMLLiteral)\> \not \in
 \PT_I(I(\rdf\?type))$.

 \item $I$ satisfies the RDF axiomatic triples, shown in Table \ref{table:RDFaxiomaticTriples}. $\Box$
\end{enumerate}
}
}
\end{definition}

{\small 
 \puttableSV{table:RDFSaxiomaticTriples}{The RDFS axiomatic triples}{}{\small { \btbl{l}
$\mathit{rdfs}\?domain(\rdf\?type, \mathit{rdfs}\?\mathit{Resource})$ \\
$\mathit{rdfs}\?domain(\mathit{rdfs}\?domain, \rdf\?\mathit{Property})$ \\
$\mathit{rdfs}\?domain(\mathit{rdfs}\?range, \rdf\?\mathit{Property})$  \\
$\mathit{rdfs}\?domain(\mathit{rdfs}\?\mathit{subPropertyOf}, \rdf\?\mathit{Property})$ \\
$\mathit{rdfs}\?domain(\mathit{rdfs}\?\mathit{subClassOf}, \mathit{rdfs}\?\mathit{Class})$ \\
$\mathit{rdfs}\?domain(\rdf\?subject, \rdf\?Statement)$  \\
$\mathit{rdfs}\?domain(\rdf\?predicate, \rdf\?Statement)$ \\
$\mathit{rdfs}\?domain(\rdf\?object, \rdf\?Statement)$ \\
$\mathit{rdfs}\?domain(\mathit{rdfs}\?member, \mathit{rdfs}\?\mathit{Resource})$ \\
$\mathit{rdfs}\?domain(\rdf\?first, \rdf\?List)$  \\
$\mathit{rdfs}\?domain(\rdf\?rest, \rdf\?List)$  \\
$\mathit{rdfs}\?domain(\mathit{rdfs}\?seeAlso, \mathit{rdfs}\?\mathit{Resource})$  \\
$\mathit{rdfs}\?domain(\mathit{rdfs}\?isDefinedBy, \mathit{rdfs}\?\mathit{Resource})$  \\
$\mathit{rdfs}\?domain(\mathit{rdfs}\?comment, \mathit{rdfs}\?\mathit{Resource})$  \\
$\mathit{rdfs}\?domain(\mathit{rdfs}\?label, \mathit{rdfs}\?\mathit{Resource})$  \\
$\mathit{rdfs}\?domain(\mathit{rdfs}\?value, \mathit{rdfs}\?\mathit{Resource})$  \\

$\mathit{rdfs}\?range(\rdf\?type, \mathit{rdfs}\?\mathit{Class})$  \\
$\mathit{rdfs}\?range(\mathit{rdfs}\?domain, \mathit{rdfs}\?\mathit{Class})$  \\
$\mathit{rdfs}\?range(\mathit{rdfs}\?range, \mathit{rdfs}\?\mathit{Class})$  \\
$\mathit{rdfs}\?range(\mathit{rdfs}\?\mathit{subPropertyOf}, \rdf\?\mathit{Property})$  \\
$\mathit{rdfs}\?range(\mathit{rdfs}\?\mathit{subClassOf}, \mathit{rdfs}\?\mathit{Class})$  \\
$\mathit{rdfs}\?range(\rdf\?subject, \mathit{rdfs}\?\mathit{Resource})$  \\
$\mathit{rdfs}\?range(\rdf\?predicate, \mathit{rdfs}\?\mathit{Resource})$  \\
$\mathit{rdfs}\?range(\rdf\?object, \mathit{rdfs}\?\mathit{Resource})$  \\
$\mathit{rdfs}\?range(\mathit{rdfs}\?member, \mathit{rdfs}\?\mathit{Resource})$  \\
$\mathit{rdfs}\?range(\rdf\?first, \mathit{rdfs}\?\mathit{Resource})$  \\
$\mathit{rdfs}\?range(\rdf\?rest, \rdf\?List)$  \\
$\mathit{rdfs}\?range(\mathit{rdfs}\?seeAlso, \mathit{rdfs}\?\mathit{Resource})$  \\
$\mathit{rdfs}\?range(\mathit{rdfs}\?isDefinedBy, \mathit{rdfs}\?\mathit{Resource})$  \\
$\mathit{rdfs}\?range(\mathit{rdfs}\?comment, \mathit{rdfs}\?Literal)$  \\
$\mathit{rdfs}\?range(\mathit{rdfs}\?label, \mathit{rdfs}\?Literal)$  \\
$\mathit{rdfs}\?range(\rdf\?value, \mathit{rdfs}\?\mathit{Resource})$  \\

$\mathit{rdfs}\?\mathit{subClassOf}(\rdf\?Alt, \mathit{rdfs}\?Container)$  \\
$\mathit{rdfs}\?\mathit{subClassOf}(\rdf\?Bag, \mathit{rdfs}\?Container)$  \\
$\mathit{rdfs}\?\mathit{subClassOf}(\rdf\?Seq, \mathit{rdfs}\?Container)$  \\
$\mathit{rdfs}\?\mathit{subClassOf}(\mathit{rdfs}\?\mathit{ContainerMembershipProperty}, \rdf\?\mathit{Property})$  \\
$\mathit{rdfs}\?\mathit{subPropertyOf}(\mathit{rdfs}\?isDefinedBy, \mathit{rdfs}\?seeAlso)$  \\
$\rdf\?type(\rdfXMLLiteral, \mathit{rdfs}\?Datatype)$  \\
$\mathit{rdfs}\?\mathit{subClassOf}(\rdfXMLLiteral, \mathit{rdfs}\?Literal)$  \\
$\mathit{rdfs}\?\mathit{subClassOf}(\mathit{rdfs}\?Datatype, \mathit{rdfs}\?\mathit{Class})$  \\
$\rdf\?type(\rdf\?\_i, \mathit{rdfs}\?\mathit{ContainerMembershipProperty})$, $\;\;\forall i \in \{1,2,...\}$  \\
$\mathit{rdfs}\?domain(\rdf\?\_i, \mathit{rdfs}\?\mathit{Resource})$, $\;\;\forall i \in \{1,2,...\}$  \\
$\mathit{rdfs}\?range(\rdf\?\_i, \mathit{rdfs}\?\mathit{Resource})$, $\;\;\forall i \in \{1,2,...\}$  

\etbl   }}} 
 
\begin{definition} [RDF entailment]
{\em Let $G,G'$ be RDF graphs. We say that $G$ {\em
RDF-entails} $G'$ ($G \models^{RDF} G'$) iff for every RDF
interpretation $I$, if $I \models G$ then $I \models G'$. $\Box$
}
\end{definition}

\begin{definition} [RDFS interpretation] \label{def:RDFSInterpretation}
 {\em An {\em RDFS interpretation} $I$ of a
vocabulary $V$ is an RDF interpretation of $V \cup \V_{RDF}
\cup \V_{RDFS}$, extended by the new ontological
category $Cls_I \subseteq Res_I$ for classes, as well as the class extension mapping
$\CT_I:Cls_I \rightarrow \POW(Res_I)$, such that:

{\small
\begin{enumerate}

\item $x \in \CT_I(y) $ iff $\<x,y\> \in \PT_I(I(\rdf\?type))$.

\item The ontological categories are defined as follows:\\
$Cls_I=\CT_I(I(\mathit{rdfs}\?\mathit{Class}))$, \\
$Res_I=\CT_I(I(\mathit{rdfs}\?\mathit{Resource}))$, and \\
$\LV_I=\CT_I(I(\mathit{rdfs}\?Literal))$.

\item If $\<x,y\> \; \in \; \PT_I(I(\mathit{rdfs}\?domain))$ and
$\<z,w\> \; \in\; \PT_I(x)$ then $z \in \CT_I(y)$.

\item If $\<x,y\> \; \in \; \PT_I(I(\mathit{rdfs}\?range))$ and
$\<z,w\> \; \in\; \PT_I(x)$ then $w \in \CT_I(y)$.

\item If $x \in Cls_I$ then $\<x, I(\mathit{rdfs}\?\mathit{Resource})\> \;
\in \; \PT_I(I(\mathit{rdfs}\?\mathit{subClassOf}))$.

\item If $\<x,y\> \in \PT_I(I(\mathit{rdfs}\?\mathit{subClassOf}))$ then
$x,y \in Cls_I$, $\CT_I(x) \subseteq \CT_I(y)$.

\item $\PT_I(I(\mathit{rdfs}\?\mathit{subClassOf}))$ is a reflexive and
transitive relation on $Cls_I$.

\item If $\<x,y\> \in \PT_I(I(\mathit{rdfs}\?\mathit{subPropertyOf}))$ then
$x,y \in Prop_I$, $\PT_I(x) \subseteq \PT_I(y)$.

\item $\PT_I(I(\mathit{rdfs}\?\mathit{subPropertyOf}))$ is a reflexive and
transitive relation on $Prop_I$.

\item If $x \in \CT_I(I(\mathit{rdfs}\?Datatype))$ then $\<x,
I(\mathit{rdfs}\?Literal)\> \; \in \;
\PT_I(I(\mathit{rdfs}\?\mathit{subClassOf}))$.

\item If $x \in
\CT_I(I(\mathit{rdfs}\?ContainerMembershipProperty))$ then \\
$\<x,
I(\mathit{rdfs}\?member)\> \in
\PT_I(I(\mathit{rdfs}\?\mathit{subPropertyOf}))$.

 \item $I$
satisfies the RDFS axiomatic triples, shown in Table  \ref{table:RDFSaxiomaticTriples}. $\Box$
\end{enumerate}
}
}
\end{definition}

\begin{definition} [RDFS entailment]
{\em Let $G,G'$ be RDF graphs. We say that $G$ {\em
RDFS-entails} $G'$ ($G \models^{RDFS} G'$) iff for every RDFS
interpretation $I$, if $I \models G$ then $I \models G'$. $\Box$
}
\end{definition}

\appendix
\section*{Appendix B: Proofs}
In this Appendix, we prove the lemmas and propositions presented in the main paper.
In addition, we provide Lemma B.1, which is used in some of the proofs.
To reduce the size of the proofs, we have eliminated the namespace from the URIs in $\V_{RDF} \cup
\V_{RDFS} \cup \V_{ERDF}$.

\vspace{0.2cm}
\noindent
{\bf Lemma B.1 }
Let $F$ be an ERDF formula and let $I$ be a partial interpretation of a vocabulary $V$.
Let $u, u'$ be mappings $u, u': \Var(F) \rightarrow Res_I$ such that $u(x)=u'(x)$, $\;\; \forall x \in \FVar(F)$.
It holds: $I,u \models F$ iff $I, u' \models F$.

\noindent
{\bf Proof}:
We prove the proposition by induction.
Without loss of generality, we assume that $\sneg$ appears only in front of positive ERDF triples. Otherwise we apply the transformation rules of Definition \ref{def:satisfiesValuation}, to get an equivalent formula that
satisfies the assumption.

Let $F=p(s,o)$. It holds: $I,u \models F$ iff $p \in V$, $s,o \in V \cup \Var$, $I(p) \in Prop_I$, and
$\<[I+u](s), [I+u](o)\> \in \PT_I(I(p))$ iff $p \in V$, $s,o \in V \cup \Var$, $I(p) \in Prop_I$, and
$\<[I+u'](s), [I+u'](o)\> \in \PT_I(I(p))$ iff $I,u' \models p(s,o)$.

Let $F=\sneg p(s,o)$. It holds: $I,u \models F$ iff $p \in V$, $s,o \in V \cup \Var$, $I(p) \in Prop_I$, and
$\<[I+u](s), [I+u](o)\> \in \PF_I(I(p))$ iff $p \in V$, $s,o \in V \cup \Var$, $I(p) \in Prop_I$, and
$\<[I+u'](s), [I+u'](o)\> \in \PF_I(I(p))$ iff $I,u' \models \sneg p(s,o)$.

\noindent
{\em Assumption:} Assume that the lemma holds for the subformulas of $F$.

We will show that the lemma holds also for $F$.

Let $F= \wneg G$. It holds:
$I,u \models F$ iff $I,u \models \wneg G$ iff
$V_G \subseteq V$ and $I,u \not \models G$ iff
$V_G \subseteq V$ and $I,u' \not \models G$ iff $I,u' \models \wneg G$
iff $I,u' \models F$.

Let $F=F_1 \AND F_2$. It holds:
$I,u \models F$ iff $I,u \models F_1 \AND F_2$ iff
$I,u \models F_1$ and  $I,u \models F_2$ iff
$I,u' \models F_1$ and  $I,u' \models F_2$ iff $I,u' \models F_1 \AND F_2$
iff $I,u' \models F$.

Let $F= \exists x \; G$. We will show that (i) if $I,u \models F$ then $I,u' \models F$
and (ii) if $I,u' \models F$ then $I,u \models F$.\\
(i) Let $I,u \models F$. Then, $I,u \models \exists x G$. Thus,
  there exists a mapping $u_1: \Var(G) \rightarrow Res_I$ s.t. $u_1(y)=u(y),$ $\;\forall y \in \Var(G)-\{x\}$,
  and $I,u_1 \models G$. Let $u_2$ be the mapping $u_2: \Var(G) \rightarrow Res_I$ s.t.
  $u_2(y)=u'(y),$ $\;\forall y \in \Var(G)-\{x\}$, and $u_2(x)=u_1(x)$. Since $u(z)=u'(z), \; \forall z \in \FVar(F)$   	and $x \in \FVar(G)$,
  it follows that $u_1(z)=u_2(z)$, $\; \forall z \in \FVar(G)$. Thus, $I,u_2 \models G$.
  Therefore,  there exists a mapping $u_2: \Var(G) \rightarrow Res_I$ s.t.
  $u_2(y)=u'(y),$ $\;\forall y \in \Var(G)-\{x\}$, and
  $I,u_2 \models G$. Thus, $I,u' \models \exists x \; G$, which implies that $I,u' \models F$.\\
(ii) We prove this statement similarly to (i)  by exchanging $u$ and $u'$.

Let $F=F_1 \OR F_2$ or $F=F_1 \mImpl F_2$ or $F= \forall x G$.  We can prove, similarly to the above cases,
that
$I,u \models F$ iff $I,u' \models F$. $\Box$

\vspace{0.2cm}
\noindent
{\bf Lemma \ref{lem:ERDFInterpretation_ERDF_graph}.}
Let $G$ be an ERDF graph and let $I$ be a partial interpretation of a vocabulary $V$.
It holds:
$I \models_{\tt GRAPH} G$ iff $I \models \formula(G)$.

\noindent
{\bf Proof}: Let $G=\{t_1, ..., t_n\}$ and $F=\formula(G)$. \\
$\Rightarrow)$ Assume that $I \models_{\tt GRAPH} G$, we will show that $I \models F$.
Since $I \models_{\tt GRAPH} G$, it follows that
$\exists v:\Var(G) \rightarrow Res_I$ such that $I,v \models t_i$, $\forall i=1,...,n$.
Thus,
$\exists v:\Var(G) \rightarrow Res_I$ such that $I,v \models t_1 \AND ... \AND t_n$.
This implies that
$\exists u:\Var(G) \rightarrow Res_I$ such that $I,u \models F$.
Since $\FVar(F)=\emptyset$, it follows from Lemma B.1 that
$\forall u':\Var(G) \rightarrow Res_I$, it holds that $I,u' \models F$. Thus, $I \models F$.\\

\noindent
$\Leftarrow)$ Assume that $I \models F$, we will show that $I \models_{\tt GRAPH} G$.
Since $I \models F$, it follows that
$\forall v:\Var(G) \rightarrow Res_I$ it holds that $I,v \models F$.
Thus, $\exists v:\Var(G) \rightarrow Res_I$ such that $I,v \models F$.
This implies that $\exists u:\Var(G) \rightarrow Res_I$ such that $I,u \models t_1 \AND ... \AND t_n$.
Thus,
$\exists u:\Var(G) \rightarrow Res_I$ such that $I,u \models t_i$, $\forall i=1, ..., n$.
Therefore, $I \models_{\tt GRAPH} G$. $\Box$\\

\vspace{0.2cm}
\noindent
{\bf Proposition \ref{prop:classicalNegation}.}
Let $I$ be an ERDF interpretation of a vocabulary $V$ and let $V'= V \cup \V_{RDF} \cup \V_{RDFS} \cup \V_{ERDF}$.
Then,
\begin{enumerate}

\item For all $p,s,o \in V'$ such that $I(p) \in \TProp_I$, it holds:\\
   $I \models \wneg p(s,o)$ iff $I \models \sneg p(s,o)$ (equivalently, $I \models p(s,o) \vee \sneg p(s,o)$).

\item For all $x,c \in V'$ such that $I(c) \in \TCls_I$, it holds:\\
 $I \models \wneg \rdf\?type(x, c)$ iff $I \models \sneg \rdf\?type(x, c)$ \\
 (equivalently, $I \models \rdf\?type(x, c) \vee \sneg \rdf\?type(x, c)$).

\end{enumerate}

\noindent
{\bf Proof:}\\
\noindent
1) It holds: $I \models \wneg p(s,o)$ iff $I \not \models p(s,o)$ iff $\<I(s), I(o)\> \not \in \PT_I(p)$
iff (since $p \in \TProp_I$) $\<I(s), I(o)\> \in \PF_I(p)$ iff $I \models \sneg p(s,o)$.
Therefore, $I \models \wneg p(s,o)$ iff $I \models \sneg p(s,o)$.

We will also show that $I \models p(s,o) \vee \sneg p(s,o)$.
It holds $I \models p(s,o)$ or $I \models \wneg p(s,o)$. This implies that
$I \models p(s,o)$ or $I \models \sneg p(s,o)$, and thus, $I \models p(s,o) \vee \sneg p(s,o)$.

\noindent
2) The proof is similar to the proof of 1) after replacing $p(s,o)$ by $type(x, c)$
and $\TProp_I$ by $\TCls_I$. $\Box$

\vspace{0.2cm}
\noindent
{\bf Proposition \ref{ERDFextendsRDFS}.}
Let $G,G'$ be RDF graphs such that $V_G \cap
\V_{ERDF}=\emptyset$ and $V_{G'} \cap \V_{ERDF}=\emptyset$. Then,
$G \models^{RDFS} G'$ iff $G \models^{ERDF} G'$.

\noindent
{\bf Proof:}\\
$\Leftarrow$) Let $G \models^{ERDF} G'$. We will show that $G \models^{RDFS} G'$.
In particular, let $I$ be an RDFS interpretation of a vocabulary $V$ s.t. $I \models G$,
we will show that $I \models G'$.

Since $I \models G$, it holds that $\exists v: \Var(G) \rightarrow Res_I$ s.t. $I, v \models G$.
Our goal is to construct an ERDF interpretation $J$ of $V$ s.t. $J \models G$.
We consider an 1-1 mapping $res:\V_{ERDF} \rightarrow R$, where $R$ is a set disjoint from $Res_I$.
Additionally, let $V'=V \cup \V_{RDF} \cup \V_{RDFS} \cup \V_{ERDF}$.
Based on $I$ and the mapping $res$, we construct a partial interpretation $J$ of $V$ as follows:
\begin{itemize}
\item $Res_J=Res_I \cup res(\V_{ERDF})$.

\item $J_V(x)=I_V(x)$, $\; \forall x \in (V' - \V_{ERDF}) \cap \URI$ and $J_V(x)=res(x)$, $\; \forall x \in \V_{ERDF}$.

\item We define the mapping: $\IL_J: V' \cap \TL \rightarrow Res_J$ such that:
$\IL_J(x)= \IL_I(x)$.

\item  We define the mapping: $\;J: V' \rightarrow Res_J$ such that:

 \begin{itemize} \item
 $J(x)=J_V(x)$,  $\; \forall x \in V' \cap \URI$.

 \item $J(x)=x$, $\; \forall \; x \in V' \cap \PL.$

 \item $J(x)=\IL_J(x)$, $\; \forall \; x \in V' \cap \TL.$
 \end{itemize}

\item We define the mapping $\PT_J': Res_J \rightarrow \POW(Res_J \times Res_J)$  as follows:

\begin{tabbing}
 aaaaaaa\= aaaaaaa \= aaaaa \= aaaaaa \=  \kill
(PT1) \> if $x,y,z \in Res_I$ and $\L x, y\R \in \PT_I(z)$  then
$\L x, y\R \in \PT_J'(z)$. \\
(PT2) \> $\L res(\mathit{TotalClass}), J(\mathit{Class})\R \in \PT_J'(J(\mathit{subClassOf}))$. \\
(PT3) \> $\L res(\mathit{TotalProperty}), J(\mathit{Property}) \R \in \PT_J'(J(\mathit{subClassOf}))$.
\end{tabbing}

Starting from the derivations of (PT1), (PT2), and (PT3), the following rules are applied recursively, until a fixpoint is reached:

\begin{tabbing}
 aaaaaaa\= aaaaaaa \= aaaaa \= aaaaaa \=  \kill
(PT4) \> if $\L x,y \R \in \PT_J'(J(domain))$ and $\L z,w \R \in \PT_J'(x)$ then \\
 \> $\L z,y \R \in \PT_J'(J(type))$.\\
(PT5) \> if $\L x,y \R \in \PT_J'(J(range))$ and $\L z,w \R \in \PT_J'(x)$ then \\
 \> $\L w,y \R \in \PT_J'(J(type))$.\\
(PT6) \> if $\L x,J(\mathit{Class}) \R \in \PT_J'(J(type))$ then \\
\> $\L x,J(\mathit{Resource}) \R \in \PT_J'(J(\mathit{subClassOf}))$.\\
(PT7) \> if $\L x,y \R \in \PT_J'(J(\mathit{subClassOf}))$  then $\L x,J(\mathit{Class}) \R \in \PT_J'(J(type))$.\\
(PT8) \> if $\L x,y \R \in \PT_J'(J(\mathit{subClassOf}))$  then $\L y,J(\mathit{Class}) \R \in \PT_J'(J(type))$.\\
(PT9) \> if $\L x,y \R \in \PT_J'(J(\mathit{subClassOf}))$ and $\L z,x \R \in \PT_J'(J(type))$ then \\
\>  $\L z,y \R \in \PT_J'(J(type))$.\\
(PT10) \> if $\L x,J(\mathit{Class}) \R \in \PT_J'(J(type))$  then $\L x,x \R \in \PT_J'(J(\mathit{subClassOf}))$.\\
(PT11) \> if $\L x,y \R \in \PT_J'(J(\mathit{subClassOf}))$ and $\L y,z \R \in \PT_J'(J(\mathit{subClassOf}))$ then  \\
\> $\L x,z \R \in \PT_J'(J(\mathit{subClassOf}))$.\\
(PT12) \> if $\L x,y \R \in \PT_J'(J(\mathit{subPropertyOf}))$  then $\L x,J(\mathit{Property}) \R \in \PT_J'(J(type))$.\\
(PT13) \> if $\L x,y \R \in \PT_J'(J(\mathit{subPropertyOf}))$  then $\L y,J(\mathit{Property}) \R \in \PT_J'(J(type))$.\\
(PT14) \> if $\L x,y \R \in \PT_J'(J(\mathit{subPropertyOf}))$ and $\L z,w \R \in \PT_J'(x)$ then \\
 \> $\L z,w \R \in \PT_J'(y)$.\\
(PT15) \> if $\L x,J(\mathit{Property}) \R \in \PT_J'(J(type))$  then $\L x,x \R \in \PT_J'(J(\mathit{subPropertyOf}))$.\\
(PT16) \> if $\L x,y \R \in \PT_J'(J(\mathit{subPropertyOf}))$ and $\L y,z \R \in \PT_J'(J(\mathit{subPropertyOf}))$ \\
\> then $\L x,z \R \in \PT_J'(J(\mathit{subPropertyOf}))$.\\
(PT17) \>if $\L x,J(Datatype)\R \in \PT_J'(J(type))$  then \\
\> $\L x,J(Literal)\R \in \PT_J'(J(\mathit{subClassOf}))$.\\
(PT18) \> if $\L x,J(ContainerMembershipProperty)\R \in \PT_J'(J(type))$  then \\
\> $\L x,J(member)\R \in \PT_J'(J(\mathit{subPropertyOf}))$.

\end{tabbing}

After reaching fixpoint, nothing else is contained in $\PT_J'(x), \; \forall x \in Res_J$.

\item $Prop_J=\{x \in Res_J \;|\; \<x, J(\mathit{Property})\} \in \PT_J'(J(type))\}$.

\item The mapping $\PT_J: Prop_J \rightarrow \POW(Res_J \times Res_J)$ is defined as follows: \\
$\PT_J(x)=  \PT_J'(x)$, $\; \forall x \in Prop_J$.

\item $\LV_J= \{x \in Res_J \;|\; \<x, J(Literal)\> \in \PT_J(J(type))\}$.

\item The mapping $\PF_J: Prop_J \rightarrow \POW(Res_J \times Res_J)$ is defined as follows:

\begin{tabbing}
 aaaaaaa\= aaaaaaa \= aaaaa \= aaaaaa \=  \kill
(PF1) \> if $``s"\h\h \rdfXMLLiteral \in V$ is an ill-typed XML-Literal then \\
\> $\L \IL_J(``s"\h\h \rdfXMLLiteral),J(Literal) \R \in \PF_J(J(type))$.\\
(PF2) \> if $\L J(\mathit{TotalClass}), J(\mathit{TotalClass}) \R \in \PT_J(J(type))$ then\\
\> $\forall x \in Res_J-\{J(\mathit{TotalClass})\}, \;\; \L x, J(\mathit{TotalClass}) \R \in \PF_J(J(type))$.\\
(PF3) \> if $\L J(\mathit{TotalProperty}), J(\mathit{TotalProperty}) \R \in \PT_J(J(type))$ then\\
\> $\forall x,y \in Res_J, \;\; \L x, y \R \in \PF_J(J(\mathit{TotalProperty}))$.
\end{tabbing}
\vspace{-0.1cm}

Starting from the derivations of (PF1), (PF2), and (PF3), the following rules are applied recursively, until a fixpoint is reached:

\begin{tabbing}
 aaaaaaa\= aaaaaaa \= aaaaa \= aaaaaa \=  \kill
(PF4) if $\L x,y \R \in \PT_J(J(\mathit{subClassOf}))$ and $\L z,y \R \in \PF_J(J(type))$ then \\
\> $\L z,x \R \in \PF_J(type)$.\\
(PF5) if $\L x,y \R \in \PT_J(J(\mathit{subPropertyOf}))$ and $\L z,w \R \in \PF_J(y)$ then \\
\> $\L z,w \R \in \PF_J(x)$.
\end{tabbing}
\vspace{-0.1cm}

After reaching fixpoint, nothing else is contained in $\PF_J(x), \; \forall x \in Prop_J$.
\end{itemize}

Before we continue, we prove the following lemma:\\\\
{\em Lemma}: For all $x,y,x \in Res_J$, $\;\; \<x,y\> \in \PT_J'(z)$ iff $\<x,y\> \in \PT_J(z)$.\\
{\em Proof} :\\
$\Leftarrow$) if $\<x,y\> \in \PT_J(z)$, then from the definition of $\PT_J$,
it follows immediately that $\<x,y\> \in \PT_J'(z)$.\\
$\Rightarrow$) Let $\<x,y\> \in \PT_J'(z)$. Then, from the definition of $\PT_J'$,
it follows that it holds (i) $z \in Prop_I$ or (ii) $\exists w \in Res_J$, s.t.
$\<w,z\> \in \PT_J'(J(\mathit{subPropertyOf}))$.

(i) Assume that $z \in Prop_I$. Then, $\<z, I(\mathit{Property})\> \in \PT_I(I(type))$.
This implies that $\<z, J(\mathit{Property})\> \in \PT_I(J(type))$. From (PT1), it now follows that 
$\<z, J(v)\> \in \PT_J'(J(type))$.
Therefore, $z \in Prop_J$. From the definition of $\PT_J$, it now
follows that $\<x,y\> \in \PT_J(z)$.

(ii) Assume that $\exists w \in Res_J$ s.t.
$\<w,z\> \in \PT_J'(J(\mathit{subPropertyOf}))$. Then, from (PT13), it follows that
$\<z, J(\mathit{Property})\> \in \PT_J'(J(type))$.
Therefore, $z \in Prop_J$. From the definition of $\PT_J$, it now
follows that $\<x,y\> \in \PT_J(z)$.\\
{\em End of Lemma}\\

Though not mentioned explicitly, the above Lemma is used throughout the rest of the proof.

\vspace{0.1cm}
To show that $J$ is a partial interpretation of $V'$, it is enough to show that $V' \cap \PL \subseteq \LV_J$.
Let $x \in V' \cap \PL$. Then, $x \in \LV_I$. Thus,
$\<x, I(Literal)\> \in \PT_I(I(type))$. Due to (PT1), this implies that $\<x, J(Literal)\> \in \PT_J(J(type))$.
Thus, $x \in \LV_J$.

\vspace{0.1cm}
Now, we extend $J$ with the ontological categories:\\
 $Cls_J = \{x \in Res_J \;|\; \<x, J(\mathit{Class})\> \in \PT_J(J(type))\}$,\\
 $\TCls_J=\{x \in Res_J \;|\; \<x, J(\mathit{TotalClass})\> \in \PT_J(J(type))\}$, and\\
 $\TProp_J=\{x \in Res_J \;|\; \<x, J(\mathit{TotalProperty})\> \in \PT_J(J(type))\}$. \\
 We define $\CT_J,\CF_J: Cls_J \rightarrow \POW(Res_J)$ as follows:\\
$x \in \CT_J(y) $ iff $\<x,y\> \in \PT_J(J(type))$, and\\
 $x \in \CF_J(y)$ iff $\<x,y\> \in \PF_J(J(type))$.

We will now show that $J$ is an ERDF interpretation of $V$.
Specifically, we will show that $J$ satisfies the semantic conditions of Definition \ref{def:ERDFInterpretation}
(ERDF interpretation) and Definition \ref{def:CoherentERDFinterpretation} (Coherent ERDF interpretation).\\

First, we will show that $J$ satisfies semantic condition 2 of Definition \ref{def:ERDFInterpretation}.
We will start by proving that $Res_J = \CT_J(J(\mathit{Resource}))$.
Obviously, \\
$\CT_J(J(\mathit{Resource})) \subseteq Res_J$. Thus, it is enough to prove that
$Res_J \subseteq \CT_J(J(\mathit{Resource}))$. Let $x \in Res_J$. Then, we distinguish the following cases:

Case 1) $x \in Res_I$. Since $I$ is an RDFS interpretation, it holds that $\<x, I(\mathit{Resource})\> \in $ $\PT_I(I(type))$. Thus, it holds $\<x, J(\mathit{Resource})\> \in \PT_J(J(type))$, which implies
that $x \in $ $\CT_J(J(\mathit{Resource}))$.

Case 2) $x \in res(\V_{ERDF})$. From the definition of $\PT_J'$, it follows that\\
$\<x, J(\mathit{Resource})\> \in \PT_J'(J(type))$.
Thus, $\<x, J(\mathit{Resource})\> \in \PT_J(J(type))$, which implies that
 $x \in \CT_J(J(\mathit{Resource}))$.

Thus, $Res_J = \CT_J(J(\mathit{Resource}))$.

\vspace{0.1cm}
Additionally, it is easy to see that it holds $Prop_J = \CT_J(J(\mathit{Property}))$, $Cls_J = \CT_J(J(\mathit{Class}))$, $\LV_J = \CT_J(J(Literal))$,
$\TCls_J = \CT_J(J(\mathit{TotalClass}))$, and \\
$\TProp_J = \CT_J(J(\mathit{TotalProperty}))$. 

\vspace{0.1cm}
We will now show that $J$ satisfies semantic condition 3 of Definition \ref{def:ERDFInterpretation}.
Let $\<x,y\> \; \in \; \PT_J(J(domain))$ and $\<z,w\> \; \in \; \PT_J(x)$. Then,
from (PT4) and the definition of $\CT_J$, it follows that $z \in \CT_J(y)$.

\vspace{0.1cm}
We will now show that $J$ satisfies semantic condition 4 of Definition \ref{def:ERDFInterpretation}.
Let $\<x,y\> \; \in \; \PT_J(J(range))$ and $\<z,w\> \; \in \; \PT_J(x)$. Then,
from (PT5) and the definition of $\CT_J$, it follows that $w \in \CT_J(y)$.

\vspace{0.1cm}
We will now show that $J$ satisfies semantic condition 5 of Definition \ref{def:ERDFInterpretation}.
Let $x \in Cls_J$.
Thus, it holds: $\<x, J(\mathit{Class})\> \in \PT_J(J(type))$. From (PT6), it now follows
that $\< x, J(\mathit{Resource})\> \in \PT_J(J(\mathit{subClassOf}))$.

\vspace{0.1cm}
We will now show that $J$ satisfies semantic condition 6 of Definition \ref{def:ERDFInterpretation}.
Let $\<x,y\> \in \PT_J(J(\mathit{subClassOf}))$. Then,
from (PT7), (PT8), and the definition of $\CT_J$, it follows that $x,y \in Cls_J$.

Let $\<x,y\> \in \PT_J(J(\mathit{subClassOf}))$. We will show that $\CT_J(x) \subseteq \CT_J(y)$.
In particular, let $z \in \CT_J(x)$. Then,
from (PT9) and the definition of $\CT_J$, it follows that $z \in \CT_J(y)$.

Let $\<x,y\> \in \PT_J(J(\mathit{subClassOf}))$. We will show that $\CF_J(y) \subseteq \CF_J(x)$.
In particular, let $z \in \CF_J(y)$. Then,
from (PF4) and the definition of $\CF_J$, it follows that $z \in \CF_J(x)$. 

\vspace{0.1cm}
In a similar manner, we can prove that $J$ also satisfies the semantic conditions 7, 8, 9, 10, and 11
of Definition \ref{def:ERDFInterpretation}. 

\vspace{0.1cm}
To continue the rest of the proof, we need to make a few observations.

Consider the mapping $h: Res_J \rightarrow Res_I$, which is defined as follows:

\[ h(x)= \left\{ \begin{array}{ll}
x & \mbox{if $x \in Res_I$ }\\
I(\mathit{Class}) & \mbox{if $x = res(\mathit{TotalClass})$}\\
I(\mathit{Property}) & \mbox{if $x = res(\mathit{TotalProperty})$}\\
\end{array} \right. \]

\noindent
{\em Observation 1}: If $\<x,y\> \in \PT_J(z)$ and $y \in res(\V_{ERDF})$ then $x=y$.\\
{\em Observation 2}: If $x \in res(\V_{ERDF})$ and $x \in Prop_J$ then $\PT_J(x)=\emptyset$.\\
{\em Observation 3}:  If $\<x,y\> \in \PT_J(z)$ then $\<h(x),h(y)\> \in \PT_I(h(z))$.\\
{\em Observation 4}: If $x,y,z \in Res_I$ and $\<x,y\> \in \PT_J(z)$ then $\<x,y\> \in \PT_I(z)$\footnote
{Note that {\em Observation 3} implies {\em Observation 4}.}.\\
The proof of these observations is made by induction. It is easy to see that all observations hold
for the derivations of (PT1), (PT2), and (PT3).  Assume now that the observations hold
for the derivations obtained at a step $k$ of the application of the fixpoint operator for $\PT_J$. Then, the observations also hold
for the derivations  obtained at step $k+1$.

\vspace{0.1cm}
We will now show that $J$ satisfies semantic condition 12 of Definition \ref{def:ERDFInterpretation}.
Let $x \in \TCls_J$. Thus, $\<x, J(\mathit{TotalClass})\> \in \PT_J(J(type))$. From {\em Observation 1},
it follows that $x=J(\mathit{TotalClass})$. From (PF2), it now follows that $\CT_J(J(\mathit{TotalClass})) \cup
\CF_J(J(\mathit{TotalClass}))=Res_J$. Thus, $\CT_J(x) \cup \CF_J(x)=Res_J$.

\vspace{0.1cm}
We will now show that $J$ satisfies semantic condition 13 of Definition \ref{def:ERDFInterpretation}.
Let $x \in \TProp_J$. Thus, $\<x, J(\mathit{TotalProperty})\> \in \PT_J(J(type))$. From {\em Observation 1},
it follows that $x=J(\mathit{TotalProperty})$. From (PF3), it now follows that $\PT_J(J(\mathit{TotalProperty})) \cup
\PF_J(J(\mathit{TotalProperty}))=Res_J \times Res_J $. Thus, $\PT_J(x) \cup \PF_J(x)=Res_J \times Res_J$.

\vspace{0.1cm}
We will now show that $J$ satisfies semantic condition 14 of Definition \ref{def:ERDFInterpretation}. \\
Let 
$``s"\h\h \rdfXMLLiteral$ be a well-typed XML-Literal in $V$ then
$\IL_J(``s"\h\h \rdfXMLLiteral)$
$= \IL_I(``s"\h\h \rdfXMLLiteral)$ is the XML value of $s$.
Additionally, since $I$ is an RDFS interpretation of $V$, it holds:
$\<\IL_I(``s"\h\h \rdfXMLLiteral),I(\XMLLiteral)\> \in \PT_I(I(type))$.
Therefore, from (PT1), it follows that 
$\<\IL_J(``s"\h\h \rdfXMLLiteral),J(\XMLLiteral)\>$ 
$\in$ \\
$ \PT_J(J(type))$.

\vspace{0.1cm}
We will now show that $J$ satisfies semantic condition 15 of Definition \ref{def:ERDFInterpretation}. 
Let \\
$``s"\h\h \rdfXMLLiteral \in V$ s.t. $s$ is not a well-typed XML literal string.
Assume now that $\IL_J(``s"\h\h \rdfXMLLiteral) \in $
$\LV_J$. Then,
$\<\IL_J(``s"\h\h \rdfXMLLiteral), J(Literal)\> $ $\in $
$\PT_J(J(type))$. From {\em Observation 4}, 
it follows that $\<\IL_J(``s"\h\h \rdfXMLLiteral), $
$J(Literal)\> $ $\in $
$ \PT_I(J(type))$.
Therefore, it follows that
$\<\IL_I(``s"\h\h \rdfXMLLiteral), $
$I(Literal)\> \in $\\
$\PT_I(I(type))$. Thus, $\IL_I(``s"\h\h \rdfXMLLiteral) \in \LV_I$,
which is impossible since $I$ is an RDFS interpretation of $V$.
Therefore,  $\IL_J(``s"\h\h \rdfXMLLiteral) \in Res_J- \LV_J$.

Additionally, from (PF1), it follows that $\<\IL_J(``s"\h\h \rdfXMLLiteral), J(Literal)\> \in $\\
$\PF_J(J(type))$.

\vspace{0.1cm}
$J$ also satisfies semantic condition 16 of Definition \ref{def:ERDFInterpretation},
due to (PT1). Finally, $J$ satisfies semantic condition 17, due to (PT2) and (PT3).

Thus, $J$ is an ERDF interpretation of $V$.

Now, we will show that $J$ is a coherent ERDF interpretation (Definition \ref{def:CoherentERDFinterpretation}).
Assume that
this is not the case. Thus, there is $z \in Prop_J$ s.t. $\PT_J(z) \cap \PF_J(z) \not = \emptyset$.
Assume that $\<x,y\> \in \PT_J(z) \cap \PF_J(z)$, for such a $z$.
We distinguish the following cases:

{\em Case 1}) $\;z \in res(\V_{ERDF})$. Then, from {\em Observation 2},
it follows that $\PT_J(z)=\emptyset$, which is a contradiction.

{\em Case 2}) $\;y \in res(\V_{ERDF})$ and $z \in Res_I$. Then, it holds:\\
 (i) $\< z, res(\mathit{TotalProperty})\> \in \PT_J(J(\mathit{subPropertyOf}))$, or \\
 (ii) $\< z, J(type)\>
\in \PT_J(J(\mathit{subPropertyOf}))$ and $\<x,y\> \in \PF_J(J(type))$.

\noindent
Now, from {\em Observation 1} and since $z \in Res_I$,  (i) is impossible. Thus,
$\< z, J(type)\>
\in \PT_J(J(\mathit{subPropertyOf}))$ and $\<x,y\> \in \PF_J(J(type))$. This implies that \\
$y = res(\mathit{TotalClass})$. From {\em Observation 1},
it follows that $x=res(\mathit{TotalClass})$, which is impossible since, due to (PF2),
$\<res(\mathit{TotalClass}),res(\mathit{TotalClass})\> \not \in \PF_J(J(type))$.

{\em Case 3}) $x \in res(\V_{ERDF})$ and $y, z \in Res_I$.  Then, it holds:\\
 (i) $\< z, res(\mathit{TotalProperty})\> \in \PT_J(J(\mathit{subPropertyOf}))$, or \\
 (ii) $\< z, J(type)\>
\in \PT_J(J(\mathit{subPropertyOf}))$ and $\<x,y\> \in \PF_J(J(type))$.

\noindent
Now, from {\em Observation 1} and since $z \in Res_I$,  (i) is impossible. Thus,
$\< z, J(type)\>
\in \PT_J(J(\mathit{subPropertyOf}))$ and $\<x,y\> \in \PF_J(J(type))$. This implies that \\
$y = res(\mathit{TotalClass})$, which is impossible, since
$y \in Res_I$.

{\em Case 4}) $x,y,z \in Res_I$.
Then, $x=\IL_J(s)$, where $s$ is an ill-typed XML-Literal in $V$,
$\<z,J(type)\> \in \PT_J(J(\mathit{subPropertyOf}))$ and $\<y,J(Literal)\> \in \PT_J(J(\mathit{subClassOf}))$.
Since $\<x,y\> \in \PT_J(z)$, it follows that $\<x,y\> \in \PT_J(J(type))$.
Since $\<y,J(Literal)\> \in \PT_J(J(\mathit{subClassOf}))$, it follows that $\<x, J(Literal)\> \in \PT_J(J(type))$. From {\em Observation 4}, 
it follows that $\<\IL_J(s), J(Literal)\> \in \PT_I(J(type))$.
Therefore,\\
$\<\IL_I(s), I(Literal)\> \in \PT_I(I(type))$. But this implies that $\IL_I(s) \in \LV_I$,
which is impossible since $I$ is an RDFS interpretation of $V$.

Since all cases lead to contradiction, it follows that:
\begin{center}
$\forall z \in Prop_J$, $\;\; \PT_J(z) \cap \PF_J(z) = \emptyset$.
\end{center}

We will now show that $J, v \models G$. Let $p(s,o) \in G$. Since $I,v \models G$, it holds
that $p \in V'$, $s,o \in V' \cup \Var$.  Note that, due to (PT1), it holds $Prop_I \subseteq Prop_J$.
Since $p \not \in \V_{ERDF}$, it holds $J(p)=I(p) \in Prop_I \subseteq Prop_J$. Since $s,o \not \in \V_{ERDF}$, it holds that $[I+v](s)=[J+v](s)$ and $[I+v](o)=[J+v](o)$.
Since $I,v \models G$, it holds $\<[I+v](s), [I+v](o)\> \in \PT_I(I(p))$.
Thus,  $\<[J+v](s), [J+v](o)\> \in \PT_I(J(p))$. From (PT1), it follows that
$\<[J+v](s), [J+v](o)\> \in \PT_J(J(p))$.
Thus, $J, v \models G$, which implies that $J \models G$.
Since $J$ is an ERDF interpretation and $G \models^{ERDF} G'$, it follows
that $J \models G'$. Thus, there is $u:\Var(G') \rightarrow Res_J=Res_I \cup res(\V_{ERDF})$ s.t. $J, u \models G'$.
We define a mapping $u':\Var(G') \rightarrow Res_I$ as follows:

\[ u'(x)= \left\{ \begin{array}{ll}
u(x) & \mbox{if $u(x) \in Res_I$ }\\
I(\mathit{Class}) & \mbox{if $u(x) = res(\mathit{TotalClass})$}\\
I(\mathit{Property}) & \mbox{if $u(x)= res(\mathit{TotalProperty})$}\\
\end{array} \right. \]

We will show that $I,u' \models G'$. Let $p(s,o) \in G'$.
Since $J \models G'$ and $V_{G'} \cap \V_{ERDF}=\emptyset$, it follows that $p \in V \cup \V_{RDF} \cup \V_{RDFS}$,
$s,o \in V \cup \V_{RDF} \cup \V_{RDFS} \cup \Var$, and $J(p) \in Prop_J$. Thus, $\<J(p), J(type)\> \in \PT_J(J(\mathit{Property}))$, which implies
(since $p \not \in \V_{ERDF}$) that
$\<I(p), I(type)\> \in \PT_J(I(\mathit{Property})$. Due to {\em Observation 4},
it follows that $\<I(p), I(type)\> \in \PT_I(I(\mathit{Property})$. Thus, $I(p) \in Prop_I$.
Additionally, it holds: $\<[J+u](s), [J+u](o)\> \in \PT_J(J(p))$.
We want to show that $\<[I+u'](s), [I+u'](o)\> \in \PT_I(I(p))$.

{\em Case 1}) It holds: (i) if $s \in \Var(G')$ then $u(s) \not \in res(\V_{ERDF})$ and
 (ii) if $o \in \Var(G')$ then $u(o) \not \in res(\V_{ERDF})$. \\
 Then, $[J+u](s)=[J+u'](s)=[I+u'](s) \in Res_I$,
 $[J+u](o)=[J+u'](o)=[I+u'](o) \in Res_I$, and $J(p)=I(p) \in Res_I$.
 Thus, $\<[J+u](s), [J+u](o)\> \in \PT_J(J(p))$ implies that $\<[I+u'](s), [I+u'](o)\> \in \PT_J(I(p))$.
 From {\em Observation 4}, the latter implies that  $\<[I+u'](s), [I+u'](o)\> \in \PT_I(I(p))$.

{\em Case 2}) It holds: (i) $s \in \Var(G')$ and $u(s) \in res(\V_{ERDF})$ and
(ii) if $o \in \Var(G')$ then $u(o) \not \in res(\V_{ERDF})$. \\
Assume that $u(s)= res(\mathit{TotalClass})$, $[J+u](o)=y$, and $J(p)=z$.
Then $y,z \in Res_I$. Additionally, $I(p)=J(p)=z$ and $[I+u'](o)=[J+u](o)= y$.
Thus, $\<[I+u'](s), [I+u'](o)\>=\<I(\mathit{Class}), y\>$.
It holds $\<res(\mathit{TotalClass}), y\> \in \PT_J(z)$. Due to {\em Observation 3}, it holds
  $\<I(\mathit{Class}), y\> \in \PT_I(z).$ Thus,
 $\<[I+u'](s), [I+u'](o)\>=\<I(\mathit{Class}), y\> \in \PT_I(z)=\PT_I(I(p))$.

Similarly, if $u(s)= res(\mathit{TotalProperty})$, we prove that  $\<[I+u'](s), [I+u'](o)\> \in \PT_I(I(p))$.

{\em Case 3}) It holds:  $o \in \Var(G')$ and $u(o) \in res(\V_{ERDF})$.
Then, {\em Observation 1}, it follows that $s \in \Var(G')$ and $u(s) = u(o)$.
Assume that $u(o)= res(\mathit{TotalClass})$, and $J(p)=z$.
Then, $z \in Res_I$ and $I(p)=J(p)=z$.
Additionally, $\<[I+u'](s), [I+u'](o)\>=\<I(\mathit{Class}), I(\mathit{Class})\>$.
It holds $\<res(\mathit{TotalClass}), res(\mathit{TotalClass})\> \in \PT_J(z)$.
Due to {\em Observation 3}, it follows
that $\<I(\mathit{Class}), I(v)\> \in \PT_I(z).$ Thus,
 $\<[I+u'](s), [I+u'](o)\>=\<I(\mathit{Class}), I(\mathit{Class})\> \in \PT_I(z)=\PT_I(I(p))$.

Similarly, if $u(o) = res(\mathit{TotalProperty})$, we prove that  $\<[I+u'](s), [I+u'](o)\> \in \PT_I(I(p))$.

\vspace{0.1cm}
\noindent
As in all cases,  it holds
$\<[I+u'](s), [I+u'](o)\>=\PT_I(I(p))$, it follows that  $I, u' \models G'$, which implies that $I \models G'$.

\vspace{0.1cm}
\noindent
$\Rightarrow)$ Let $G \models^{RDFS} G'$. We will show that $G \models^{ERDF} G'$.
Let $I$ be an ERDF interpretation of a vocabulary $V$, such that $I \models G$.
Thus, there is $u: \Var(G) \rightarrow Res_I$ s.t.  $I,u \models G$.
We will show  that $I \models G'$.

We define $V'=V \cup \V_{RDF} \cup \V_{RDFS} \cup \V_{ERDF}$.
Based on $I$, we construct an RDFS interpretation $J$ of $V'$ such that:
$Res_J=Res_I, \; Prop_J=Prop_I, \; \LV_J=\LV_I, \; Cls_J=Cls_I, \;  J_V(x)=I_V(x), \forall x \in V' \cap \URI, \; \PT_J(x)=\PT_I(x),  \forall x \in Prop_J,
\; \IL_J(x)=\IL_I(x),  \forall x \in V' \cap \TL, \; \CT_J(x)=\CT_I(x),  \forall x \in Cls_J.$

We will now show that $J$ is indeed an RDFS interpretation of $V'$.

First, we will show that $J$ satisfies semantic condition 1 of Definition  \ref{def:RDFInterpretation} (Appendix A, RDF interpretation).
It holds: $x \in Prop_J$ iff $x \in Prop_I$ iff
$x \in \CT_I(I(\mathit{Property}))$ iff $\<x, I(\mathit{Property})\> \in \PT_I(I(type))$ iff $\<x, J(\mathit{Property})\> \in \PT_J(J(type))$.

We will now show that $J$ satisfies semantic condition 2 of Definition \ref{def:RDFInterpretation}.\\
Let 
$``s"\h\h \rdfXMLLiteral \in V$  such that $s$ is a
well-typed XML literal string. Then, it follows from the definition of $J$
and the fact that $I$ is an ERDF interpretation of $V$ that
 $\IL_J$($``s"\h\h \rdfXMLLiteral$) is the XML value of $s$,  and
$\IL_J(``s"\h\h \rdfXMLLiteral) \in $ $\CT_J(J(\XMLLiteral))$.
 We will show that $\IL_J(``s"\h\h \rdfXMLLiteral)\in \LV_J$.
Since $I$ is an ERDF interpretation, $\IL_I(``s"\h\h \rdfXMLLiteral) \in \CT_I(I(\XMLLiteral))$.
Additionally, $\<I(\XMLLiteral), I(Literal)\> \in \PT_I(I(\mathit{subClassOf}))$.
Therefore, 
$\IL_I(``s"\h\h \rdfXMLLiteral) \in \CT_I(I(Literal))$, and thus,
$\IL_I(``s"\h\h \rdfXMLLiteral) \in \LV_I$. The last statement implies that
 $\IL_J(``s"\h\h \rdfXMLLiteral)\in \LV_J$.

 We will now show that $J$ satisfies semantic condition 3 of Definition \ref{def:RDFInterpretation}.\\
Let 
$``s"\h\h \rdfXMLLiteral \in V$  such that $s$ is an
ill-typed XML literal string. Then, it follows from the definition of $J$
and the fact that $I$ is an ERDF interpretation of $V$ that
 $\IL_J(``s"\h\h \rdfXMLLiteral)\in Res_J-\LV_J$.
 We will show that \\
$\<\IL_J(``s"\h\h \rdfXMLLiteral), J(\XMLLiteral)\> \not \in
 \PT_J(J(type))$. Assume that \\
 $\<\IL_J(``s"\h\h \rdfXMLLiteral), J(\XMLLiteral)\> \in
 \PT_J(J(type))$. Then, \\
 $\<\IL_I(``s"\h\h \rdfXMLLiteral), I(\XMLLiteral)\> \in
 \PT_I(I(type))$. Thus, \\
 $\IL_I(``s"\h\h \rdfXMLLiteral) \in \CT_I(I(\XMLLiteral))$.
 Since it holds \\
 $\<I(\XMLLiteral), I(Literal)\> \in \PT_I(I(\mathit{subClassOf}))$, it follows that\\
 $\IL_I(``s"\h\h \rdfXMLLiteral) \in \CT_I(I(Literal))$. Thus,
 $\IL_I(``s"\h\h \rdfXMLLiteral) \in \LV_I$,
 which is impossible since $I$ is an ERDF interpretation of $V$.
 Therefore, \\
 $\<\IL_J(``s"\h\h \rdfXMLLiteral), J(\XMLLiteral)\> \not \in
 \PT_J(J(type))$. 

\vspace{0.1cm}
 It is easy to see that $J$ satisfies semantic condition 4 of Definition \ref{def:RDFInterpretation}
 and all the semantic conditions of Definition \ref{def:RDFSInterpretation} (Appendix A, RDFS Interpretation).
 Therefore, $J$ is an RDFS interpretation of $V'$.

\vspace{0.1cm}
We will now show that
$J,u \models G$. Let $p(s,o) \in G$.
Since $I \models G$, it holds that
$p \in V'$, $s,o \in V' \cup \Var$, and
$J(p)=I(p)  \in Prop_I=Prop_J$.
It holds:
$\<[J+u](s), [J+u](o)\> \in \PT_J(J(p))$ iff
$\<[I+u](s)), [I+u](o)\> \in Prop_I(I(p))$, which is true, since
$I,u \models G$. Thus, $J, u \models G$, which implies that  $J \models G$.
Since $G \models^{RDFS} G'$, it follows that $J \models G'$. Thus, there is $v:\Var(G') \rightarrow Res_J$ s.t.
$J,v \models G'$.

We will now show that  $I \models G'$. Let $p(s,o) \in G'$. Since $J, v \models G'$, it holds that $p \in V'$,
$s, o \in V' \cup \Var$, and $I(p)=J(p) \in Prop_J=Prop_I$.
It holds:
$\<[I+v](s), [I+v](o)\> \in \PT_I(I(p))$ iff
$\<[J+v](s), [J+v](o)\> \in \PT_J(J(p))$, which is true, since $J,v \models G'$.
Thus, $I,v \models G'$, which implies that $I \models G'$. $\Box$

\vspace{0.2cm}
\noindent
{\bf Proposition \ref{ERDFentailmentEquivalence1}.}
Let $G$ be an ERDF graph and let $F$ be an ERDF formula such that $V_{F} \cap sk_G(\Var(G)) =\emptyset$. It holds:
$G \models^{ERDF} F$  iff $sk(G) \models^{ERDF} F$.

\noindent
{\bf Proof:}

\noindent
$\Rightarrow$) Let $G \models^{ERDF} F$. We will show that $sk(G) \models^{ERDF} F$.
Let $I$ be an ERDF interpretation over a vocabulary $V$ s.t. $I \models sk(G)$.
We will show that $I \models G$. We define $V'=V \cup \V_{RDF} \cup \V_{RDFS} \cup \V_{ERDF}$.
Additionally, we define a total function $u:\Var(G) \rightarrow Res_I$ s.t.
$u(x)=I_V(sk_G(x)), \forall x \in \Var(G)$. Moreover, we define a total function $u':V' \cup \Var(G) \rightarrow
V'$ s.t. $u'(x)=sk_G(x),$ if $x \in \Var(G)$ and $u'(x)=x$, otherwise.

Let $p(s,o) \in G$. Then, $p \in V'$, $s,o \in V' \cup \Var$, and
$I(p)  \in Prop_I$.
It holds:
$\<[I+u](s), [I+u](o)\> \in \PT_I(I(p))$ iff
$\<I(u'(s)), I(u'(o))\> \in \PT_I(I(p))$, which is true, since $p(u'(s), u'(o)) \in sk(G)$ and
$I \models sk(G)$. Thus, $I, u \models p(s,o)$. 

Let $\sneg p(s,o) \in G$. Then, $p \in V'$, $s,o \in V' \cup \Var$, and
$I(p)  \in Prop_I$.
It holds:
$\<[I+u](s), [I+u](o)\> \in \PF_I(I(p))$ iff
$\<I(u'(s)), I(u'(o))\> \in \PF_I(I(p))$, which is true, since $\sneg p(u'(s), u'(o)) \in sk(G)$ and
$I \models sk(G)$. Thus, $I, u \models \sneg p(s,o)$. 

Therefore,  $I \models G$.
Since  $G \models^{ERDF} F$, it follows that $I \models F$. \\

\noindent
$\Leftarrow)$
 Let $sk(G) \models^{ERDF} F$. We will show that $G \models^{ERDF} F$.
Let $I$ be an ERDF interpretation of a vocabulary $V$ such that $I \models G$.
We will show that $I \models F$.
Since $I \models G$, there is a total function $u:\Var(G) \rightarrow Res_I$ s.t. $I, u \models G$.
We define $V'=V \cup \V_{RDF} \cup \V_{RDFS} \cup \V_{RDFS}$. We construct an ERDF interpretation $J$ of $V \cup sk_G(\Var(G))$ as follows:
$Res_J=Res_I, \; Prop_J=Prop_I, \; \LV_J=\LV_I, \; Cls_J=Cls_I$.
We define $J_V: (V' \cup sk_G(\Var(G))) \cap \URI \rightarrow Res_J$, as follows:
 $J_V(x)=I_V(x), \forall x \in V' \cap \URI$ and
$J_V(x)=u(sk^{-1}_G(x)), \forall x \in sk_G(\Var(G))$. Moreover, $\PT_J(x)=\PT_I(x),  \forall x \in Prop_J,$
$\PF_J(x)=\PF_I(x),  \forall x \in Prop_J,$
$\; \IL_J(x)=\IL_I(x),  \forall x \in V' \cap \TL,$ $\; \CT_J(x)=\CT_I(x),  \forall x \in Cls_J,$ and
$\; \CF_J(x)=\CF_I(x),  \forall x \in Cls_J.$

Since $I$ is an ERDF interpretation of $V$, it is easy to see that $J$ is indeed an ERDF interpretation of
$V \cup sk_G(\Var(G))$. We will show that $J \models sk(G)$.
First, we define a total function $g:V' \cup sk_G(\Var(G)) \rightarrow V' \cup \Var(G)$ as follows: $g(x)= sk^{-1}_G(x), \;\;\forall x \in sk_G(\Var(G))$ and $g(x)=x$, otherwise.
Let $p(s,o) \in sk(G)$. Since $I \models G$, it follows that $p \in V'$, $s,o \in V' \cup \Var$, and
$J(p)=I(p) \in Prop_I=Prop_J$. It holds $J(s)=[I+u](g(s))$, $J(o)=[I+u](g(o))$, and $J(p)=I(p)$.
Therefore, it holds: $\<J(s), J(o)\> \in \PT_J(J(p))$ iff
$\<[I+u](g(s)), [I+u](g(o))\> \in \PT_I(I(p))$, which holds since $p(g(s),g(o)) \in G$ and $I,u \models G$.
Let $v: \{\} \rightarrow Res_J$. It follows that $J, v \models p(s,o)$. Let $\sneg p(s,o) \in sk(G)$. We can show that
 $J, v \models \sneg p(s,o)$, in a similar manner.
Therefore, $J \models sk(G)$.\\

Since $sk(G) \models^{ERDF} F$,
it follows that $J \models F$. We will show that $I \models F$.
We define $V'= V \cup \V_{RDF} \cup \V_{RDFS} \cup \V_{ERDF}$.
Note that $Res_J=Res_I$.\\\\
{\em Lemma:} For every mapping $u:\Var(F) \rightarrow Res_J$, it holds $\; J,u \models F$ iff $I,u \models F$.\\
{\em Proof:} We will prove the Lemma by induction. Without loss of generality, we assume that $\sneg$ appears only in front of positive ERDF triples. Otherwise we apply the transformation rules of Definition \ref{def:satisfiesValuation}, to get an equivalent formula that
satisfies the assumption.

Let $F=p(s,o)$. Assume that $J, u \models F$. Since $V_{F} \cap sk_G(\Var(G)) =\emptyset$,
it follows that $p \in V'$, $s,o \in V' \cup \Var$, and
$J(p)=I(p) \in Prop_I=Prop_J$. Since $\<[J+u](s), [J+u](o)\> \in \PT_J(J(p))$,
it follows that $\<[I+u](s), [I+u](o)\> \in \PT_I(I(p))$.
Therefore, $I, u \models F$.

Assume that $I, u \models F$. It follows that $p \in V'$, $s,o \in V' \cup \Var$, and
$J(p)=I(p) \in Prop_I=Prop_J$. Since $\<[I+u](s), [I+u](o)\> \in \PT_I(I(p))$, it follows that
$\<[J+u](s), [J+u](o)\> \in \PT_J(J(p))$.
Therefore, $J, u \models F$.

Let $F= \sneg p(s,o)$. Similarly, we prove that $J,u \models F$ iff $I,u \models F$.

\noindent
{\em Assumption:} Assume that the lemma holds for the subformulas of $F$.

\noindent
We will show that the lemma holds also for $F$.

Let $F= \wneg G$. It holds:
$I,u \models F$ iff $V_G \subseteq V'$ and $I,u \not \models G$ iff $V_G \subseteq V'$ and
$J,u \not \models G$ iff $J,u \models F$.

Let $F=F_1 \AND F_2$. It holds:
$I,u \models F$ iff $I,u \models F_1$ and $I,u \models F_2$ iff
$J,u \models F_1$ and $J,u \models F_2$ iff $J,u \models F$.

Let $F= \exists x \; G$. It holds:
$I,u \models F$ iff $I,u \models \exists x \; G$ iff there is $v:\Var(G) \rightarrow  Res_I$ s.t.
$v(y)=u(y)$, $\;\forall y \in \Var(G)- \{x\}$ and $I,v \models G$ iff
there is $v:\Var(G) \rightarrow  Res_J$ s.t.
$v(y)=u(y)$, $\;\forall y \in \Var(G)- \{x\}$ and $J,v \models G$ iff $J,u \models \exists x \; G$ iff
$J,u \models F$.

Let $F=F_1 \OR F_2$ or $F=F_1 \mImpl F_2$ or $F= \forall x G$.  We can prove, similarly to the above cases,
that
$I,u \models F$ iff $J,u \models F$. \\
{\em End of lemma}\\

Since $J \models F$, it follows that for every mapping
$u:\Var(F) \rightarrow Res_J$, $\;\;J, u \models F$. Therefore, it follows from Lemma 
and the fact that $Res_J=Res_I$ that
for every mapping
$u:\Var(F) \rightarrow Res_I$, $\;\;I, u \models F$. Thus, $I \models F$. $\Box$

\vspace{0.2cm}
\noindent
{\bf Proposition \ref{prop:incomparable}.}
Let $O=\<G,P\>$ be an ERDF ontology and let $I, J \in \I^H(O)$.
Let $p \in \TProp_I \cap \TProp_J$. If $\PT_I(p) \not = \PT_J(p)$ or 
$\PF_I(p) \not = \PF_J(p)$ then $I \not \leq J$ and $J \not \leq I$.

\noindent
{\bf Proof:} Assume $\PT_I(p) \not = \PT_J(p)$. Now, assume $I \leq J$.
Then, $\PT_I(p) \subset \PT_J(p)$ and $\PF_I(p) \subseteq \PF_J(p)$.
Since  $I,J \in \I^H(O)$ and $p \in \TProp_I \cap \TProp_J$, it holds that $\PF_I(p)=Res^H_O-\PT_I(p)$
and $\PF_J(p)=Res^H_O-\PT_J(p)$. Thus, $\PF_I(p) \supset \PF_J(p)$, which is a contradiction.
Thus, $I \not \leq J$. Similarly, we can prove that $J \not \leq I$.

Assume now that $\PF_I(p) \not = \PF_J(p)$. Then, we can prove that 
$I \not \leq J$ and $J \not \leq I$, in a similar manner. $\Box$

\vspace{0.2cm}
\noindent
{\bf Proposition \ref{prop:stableImpliesHerbrand}.}
Let $O=\<G,P\>$ be an ERDF ontology and let $M \in \M^{st}(O)$. It
holds $M \in \M^H(O)$.

\noindent
{\bf Proof:}
Let $M \in \M^{st}(O)$. Obviously, $M \in \I^H(O)$ and $M \models sk(G)$.
We will show that $M \models r$, $\; \forall r \in P$.
Let $r \in P$. Let $v$ be a mapping $v :\Var(r) \rightarrow Res^H_O$ s.t. $M,v \models Cond(r)$.
It is enough to show that $M, v \models Concl(r)$.

For any mapping $u: X \rightarrow Res^H(O)$, where $X \subseteq \Var$, we  define 
the mapping $u^*: X \rightarrow V_O$ as follows:
\[ u^*(x)= \left\{ \begin{array}{ll}
u(x) & \mbox{if $u(x)$ is not the xml value of a well-typed XML literal in $V_O$}\\
t   & \mbox{if $u(x)$ is the xml value of a well-typed XML literal $t$ in $V_O$}\\
\end{array} \right. \]

Let $x \in V_O$, we define $x^{u^*}=x$. Let $x \in X$, we define $x^{u^*}=u^*(x)$. Let $F\in L(V_O)\cup \{true,false\}$ such that $\FVar(F) \subseteq X$, 
we define $F^{u^*}$ to be the formula that results from $F$ after replacing each free variable of $F$ by
$u^*(x)$. It is easy to see that it holds:
 $\; Concl(r)^{v^*} \leftarrow Concl(r)^{v^*} \in [r]_{V_O} \subseteq [P]_{V_O}$.

\vspace{0.2cm}
\noindent
{\em Lemma:} Let $F$ be an ERDF formula over $V_O$ and let 
$u$ be a mapping $u: \Var(F) \rightarrow Res^H_O$. 
It holds: $M, u \models F$ iff $M,u \models F^{u^*}$.

\noindent
{\em Proof:} We prove the lemma by induction.
Without loss of generality, we assume that $\sneg$ appears only in front of positive ERDF triples. Otherwise we apply the transformation rules of Definition \ref{def:satisfiesValuation}, to get an equivalent formula that
satisfies the assumption.

Let $F=p(s,o)$. It holds: $M,u \models F$ iff $M,u \models p(s,o)$ iff
$\<[M+u](s), [M+u](o)\> \in \PT_M(M(p))$ iff
$\<[M+u](s^{u^*}), [M+u](o^{u^*})\> \in \PT_M(M(p))$ iff $M,u \models p(s,o)^{u^*}$.

Let $F=\sneg p(s,o)$. It holds: $M,u \models F$ iff $M,u \models p(s,o)$ iff
$\<[M+u](s), [M+u](o)\> \in \PF_M(M(p))$ iff
$\<[M+u](s^{u^*}), [M+u](o^{u^*})\> \in \PF_M(M(p))$ iff $M,u \models (\sneg p(s,o))^{u^*}$.

\noindent
{\em Assumption:} Assume that the lemma holds for the subformulas of $F$.

We will show that the lemma holds also for $F$.

Let $F= \wneg G$. It holds:
$M,u \models F$ iff $M,u \models \wneg G$ iff
$M,u \not \models G$ iff
$M,u \not \models G^{u^*}$ iff $M,u \models \wneg G^{u^*}$
iff $M,u \models F^{u^*}$.

Let $F=F_1 \AND F_2$. It holds:
$M,u \models F$ iff $M,u \models F_1 \AND F_2$ iff
$M,u \models F_1$ and  $M,u \models F_2$ iff
$M,u \models F_1^{u^*}$ and  $M,u \models F_2^{u^*}$ iff $M,u \models (F_1 \AND F_2)^{u^*}$
iff $M,u \models F^{u^*}$.

Let $F= \exists x G$. It holds: $M,u \models F$ iff there 
exists a mapping $u_1: \Var(G) \rightarrow Res^H_O$ s.t. 
$u_1(y)=u(y),$ $\;\forall y \in \Var(G)-\{x\}$  s.t. $M,u_1 \models G$ iff 
there 
exists a mapping $u_1: \Var(G) \rightarrow Res^H_O$ s.t. 
$u_1(y)=u(y),$ $\;\forall y \in \Var(G)-\{x\}$  s.t. $M,u_1 \models G^{u_1^*}$ iff
there 
exists a mapping $u_1: \Var(G) \rightarrow Res^H_O$ s.t. 
$u_1(y)=u(y),$ $\;\forall y \in \Var(G)-\{x\}$  s.t. $M,u_1 \models (\exists x G)^{u_1^*}$
iff (since $u_1^*(y)=u^*(y)$, $\; \forall  y \in \FVar(\exists x G)$) $M,u \models (\exists x G)^{u^*}$ iff $M,u \models F^{u^*}$.

Let $F=F_1 \OR F_2$ or $F=F_1 \mImpl F_2$ or $F= \forall x G$.  We can prove, similarly to the above cases,
that
$M,u \models F$ iff $M,u \models F^{u^*}$.

\noindent
{\em End of Lemma}

\vspace{0.2cm}
\noindent
First assume that $Cond(r) \not = true$. Then, $Cond(r) \in L(V_O)$ and thus, $Cond(r)$ is 
an ERDF formula over $V_O$.
Since $M,v \models Cond(r)$, it follows from Lemma that
$M,v \models Cond(r)^{v^*}$. Now since $\FVar(Cond(r)^{v^*})=\emptyset$, it follows from 
Lemma B.1 that $M \models Cond(r)^{v^*}$.
Since $M \in \M^{st}(O)$, it follows that $M \models Concl(r)^{v^*}$.
Thus, $Concl(r) \not = false$ and $Concl(r) \in L(V_O|\{\sneg\})$.
Now since 
$\FVar(Concl(r)^{v^*})=\emptyset$, it follows from lemma B.1
that $M, v \models Concl(r)^{v^*}$.
Since $Concl(r)$ is an  ERDF formula over $V_O$, it follows from Lemma
that $M, v \models Concl(r)$.

Assume now that  $Cond(r) = true$. Then, $M \models Cond(r)^{v^*}$. Since $M \in \M^{st}(O)$, 
it follows that $M \models Concl(r)^{v^*}$.
Therefore, $Concl(r) \not = false$, and we can prove as above that $M, v \models Concl(r)$.

Therefore, $M \models r$, $\forall r \in P$. $\Box$

\vspace{0.2cm}
\noindent
{\bf Proposition \ref{TotalStableHerbrand}.}
Let $O=\<G,P\>$ be an ERDF ontology, such that \\
$\mathit{rdfs\?subClassOf}(\rdf\?\mathit{Property},
 \erdf\?\mathit{TotalProperty}) \in G$.
Then, $\M^{st}(O)=\M^H(O)$.

\noindent
{\bf Proof:} From Proposition \ref{prop:stableImpliesHerbrand}, it follows that $\M^{st}(O) \subseteq \M^H(O)$.
We will show that $\M^H(O) \subseteq \M^{st}(O)$.
Let $M \in \M^H(O)$. It follows that $M \models sk(G)$.
We will show that $M \in minimal(\{I \in \I^H(O) \; | \; I \models sk(G)\})$.

Let $J \in I^H(O)$ s.t. $J \models sk(G)$ and $J \leq M$. We will show that $J=M$.
Since $J \leq M$, it follows that $Prop_J \subseteq Prop_M$ and for all $p \in Prop_J$, it holds
$\PT_J(p) \subseteq \PT_M(p)$ and $\PF_J(p) \subseteq \PF_M(p)$.
Let $p \in Prop_J$. Since $J \models sk(G)$, it follows that $Prop_J \subseteq \TProp_J$.
Thus, $p \in \TProp_J$.
Assume that $\PT_J(p) \not = \PT_M(p)$.
Then, there is $\<x,y\> \in \PT_M(p)$ s.t. $\<x,y\> \not \in \PT_J(p)$.
Then,   $\<x,y\> \in \PF_J(p)$. Thus,  $\<x,y\> \in \PF_M(p)$, which is impossible, since $\<x,y\> \in \PT_M(p)$.
Thus, $\PT_J(p) = \PT_M(p)$. Similarly, we can prove that $\PF_J(p) = \PF_M(p)$.
Therefore, for all $p \in Prop_J$, it holds
$\PT_J(p) = \PT_M(p)$ and $\PF_J(p)= \PF_M(p)$. We will now show that $Prop_J=Prop_M$.
It holds $Prop_J=$$\{x \in Res^H_O \; | \; \<x, \mathit{Property} \> \in \PT_J(type)\}=$
$\{x \in Res^H_O \; | \; \<x, \mathit{Property} \> \in \PT_I(type)\}=$$ Prop_M$.
Based on these results, the fact that $J,M \in \I^H(O)$, 
it follows that  $J=M$.
Therefore, $M \in minimal(\{I \in \I^H(O) \; | \; I \models sk(G)\})$.

We will now show that $M \in minimal(\{I \in \I^H(O) \; | \; I \geq M$ and $ I
\models Concl(r),$ for all $r \in P_{[M,M]}\})$.
Since $M \in \M^H(O)$ it follows that $M \in \{I \in \I^H(O) \; | \; I \geq M$ and $ I
\models Concl(r),$ for all $r \in P_{[M,M]}\}$.
Let $J \in \{I \in \I^H(O) \; | \; I \geq M$ and $ I
\models Concl(r),$ for all $r \in P_{[M,M]}\}$ and $J \leq M$.
Since $J \geq M$, it follows that $Prop_M \subseteq Prop_J$, and for all $p \in Prop_M$, it holds
$\PT_M(p) \subseteq \PT_J(p)$ and $\PF_M(p) \subseteq \PF_J(p)$.
Since $J \leq M$, it follows that $Prop_J \subseteq Prop_M$, and for all $p \in Prop_J$, it holds
$\PT_J(p) \subseteq \PT_M(p)$ and $\PF_J(p) \subseteq \PF_M(p)$.
Therefore, it follows that $Prop_M=Prop_J$, and for all $p \in Prop_M$, it holds
$\PT_M(p) = \PT_J(p)$ and $\PF_M(p) = \PF_J(p)$.
Based on this result, the fact that $J,M \in \I^H(O)$, 
it follows that  $J=M$.

Thus, $M \in minimal(\{I \in \I^H(O) \;|\; I \geq M$ and $ I
\models Concl(r),$ for all $r \in P_{[M,M]}\})$.

Since $M$ satisfies the conditions of Definition \ref{def:stableModel} (Stable Model), it follows that
$M \in \M^{st}(O)$.
Thus, it holds $\M^H(O) \subseteq \M^{st}(O)$. 

Therefore, $\M^H(O) = \M^{st}(O)$. $\Box$

\vspace{0.2cm}
\noindent
{\bf Proposition \ref{ERDFentailmentEquivalence2}. }
Let $G$ be an ERDF graph and let $F$ be an ERDF formula such that $V_{F} \cap sk_G(\Var(G)) =\emptyset$.
It holds:
\begin{enumerate}
\item If $F$ is an ERDF {\em d}-formula and $\<G,\emptyset\> \models^{st} F$ then $G \models^{ERDF} F$.
\item If $G \models^{ERDF} F$  then $\<G,\emptyset\> \models^{st} F$.
\end{enumerate}

\noindent
{\bf Proof:}

\noindent
1) Let $\<G,\emptyset\> \models^{st} F$. We will show that $sk(G) \models^{ERDF} F$.
Let $I$ be an ERDF interpretation of a vocabulary $V$ s.t. $I \models sk(G)$.
We will show that $I \models F$. We define $V'= V \cup \V_{RDF} \cup \V_{RDFS} \cup \V_{ERDF}$.

Let $O=\<G, \emptyset\>$.
Based on $I$, we construct a partial interpretation $J$ of $V_O$ as follows:
\begin{itemize}
\item $Res_J=Res^H_{O}$.

\item $J_V(x)=x$, for all $x \in V_{O} \cap \URI$.

\item We define the mapping: $\IL_J: V_O \cap \TL \rightarrow Res_J$ such that:\\
$\IL_J(x)=x$, if $x$ is a typed literal in $V_{O}$ other than
a well-typed XML literal, and $\IL_I(x)$ is the XML value of $x$,
if $x$ is a well-typed XML literal in $V_{O}$.

\item  We define the mapping: $\;J: V_O \rightarrow Res_J$ such that:

 \begin{itemize} \item
 $J(x)=J_V(x)$,  $\; \forall x \in V_O \cap \URI$.

 \item $J(x)=x$, $\; \forall \; x \in V_O \cap \PL.$

 \item $J(x)=\IL_J(x)$, $\; \forall \; x \in V_O \cap \TL.$
 \end{itemize}

\item $Prop_J=\{x \in Res_J \;|\; \exists x' \in V_O, \; J(x')=x$ and
$I(x') \in Prop_I\}$.

\item The mapping $\PT_J: Prop_J \rightarrow \POW(Res_J \times Res_J)$ is defined as follows:\\
$\forall x,y,z \in V_O$,
it holds:\\
$\<J(x), J(y)\> \in \PT_J(J(z))$  iff
$\<I(x), I(y)\> \in \PT_I(I(z))$.

\item We define the mapping $\PF_J: Prop_J \rightarrow \POW(Res_J \times Res_J)$ as follows:\\
$\forall x,y,z \in V_O$,
it holds:\\
$\<J(x), J(y)\> \in \PF_J(J(z))$  iff
$\<I(x), I(y)\> \in \PF_I(I(z))$.

\item $\LV_J= \{x \in Res_J \;|\; \<x, J(Literal)\> \in \PT_J(J(type))\}$.

\end{itemize}

To show that $J$ is a partial interpretation, it is enough to show that $V_O \cap \PL \subseteq \LV_J$.
Let $x \in V_O \cap \PL$. Then, $x \in \LV_I$. Thus,
$\<x, I(Literal)\> \in \PT_I(I(type))$. This implies that $\<x, J(Literal)\> \in \PT_J(J(type))$.
Thus, $x \in \LV_J$.

Now, we extend $J$ with the ontological categories:\\
 $Cls_J = \{x \in Res_J \;|\; \<x, J(\mathit{Class})\> \in \PT_J(J(type))\}$, \\
 $\TCls_J=\{x \in Res_J \;|\; \<x, J(\mathit{TotalClass})\> \in \PT_J(J(type))\}$, and \\
$\TProp_J=\{x \in Res_J \;|\; \<x, J(\mathit{TotalProperty})\> \in \PT_J(J(type))\}$. \\
We define the mappings $\CT_J, \CF_J: Cls_J \rightarrow \POW(Res_J)$ as follows:\\
$x \in \CT_J(y) $ iff $\<x,y\> \in \PT_J(J(type))$, and\\
$x \in \CF_J(y)$ iff $\<x,y\> \in \PF_J(J(type))$.

We will now show that $J$ is an ERDF interpretation of $V_O$.
First, we will show that $J$ satisfies semantic condition 2 of Definition \ref{def:ERDFInterpretation}
(ERDF Interpretation), in a number of steps:\\
{\em Step 1:} Here, we prove that $Res_J = \CT_J(J(\mathit{Resource}))$.
Obviously,
$\CT_J(J(\mathit{Resource})) $ $\subseteq Res_J$. We will show that
$Res_J \subseteq \CT_J(J(\mathit{Resource}))$. Let $x \in Res_J$.
Then, there is $x' \in V_O$ such that $J(x')=x$.
We want to show that $\<J(x'), J(\mathit{Resource})\> \in \PT_J(J(type))$. It holds:
$\<J(x'), J(\mathit{Resource})\> \in \PT_J(J(type))$ iff
$\<I(x'), I(\mathit{Resource})\> \in \PT_I(I(type))$, which is true, since
$I$ is an ERDF interpretation that satisfies $sk(G)$ and $I(x') \in Res_I$. Thus, $x=J(x') \in \CT_J(J(\mathit{Resource}\mathit{Resource}))$.

\noindent
Therefore, $Res_J = \CT_J(J(\mathit{Resource}))$.

\vspace{0.1cm}
\noindent
{\em Step 2:} Here, we prove that $Prop_J = \CT_J(J(\mathit{Property}))$. We will show that $Prop_J \subseteq \CT_J(J(\mathit{Property}))$. Let $x \in Prop_J$. Then, there is $x' \in V_O$ such that $J(x')=x$ and $I(x') \in Prop_I$.
We want to show that $\<J(x'), J(\mathit{Property})\> \in \PT_J(J(type))$. It holds:
$\<J(x'), J(\mathit{Property})\> \in \PT_J(J(type))$ iff
$\<I(x'), I(\mathit{Property})\> \in \PT_I(I(type))$,\\
 which is true, since
$I(x') \in Prop_I$. Thus, $x=J(x') \in \CT_J(J(\mathit{Property}))$.

\noindent
Therefore,
$Prop_J \subseteq \CT_J(J(\mathit{Property}))$.

We will now show that $\CT_J(J(\mathit{Property})) \subseteq Prop_J $. Let $x \in \CT_J(J(\mathit{Property}))$.
Then, $\exists x' \in V_O$ such that $J(x')=x$.
It holds
$\<J(x'), J(\mathit{Property})\> \in \PT_J(J(type))$, which implies that
$\<I(x'), I(\mathit{Property})\> \in \PT_I(I(type))$. Thus, $I(x') \in Prop_I$ and
$x \in Prop_J$.

\noindent
Therefore, $\CT_J(J(\mathit{Property})) \subseteq Prop_J $.

\vspace{0.1cm}
\noindent
{\em Step 3:} By definition, it holds $Cls_J = \CT_J(J(\mathit{Class}))$, $\LV_J = \CT_J(J(Literal))$, $\TCls_J = \CT_J(J(\mathit{TotalClass}))$ and $\TProp_J = \CT_J(J(\mathit{TotalProperty}))$. 

\vspace{0.1cm}
We will now show that $J$ satisfies semantic condition 3 of Definition \ref{def:ERDFInterpretation}
(ERDF Interpretation).
Let $\<x,y\> \; \in \; \PT_J(J(domain))$ and $\<z,w\> \; \in \; \PT_J(x)$. We will show
that $z \in \CT_J(y)$.
There are $x',y' \in V_O$ such that $J(x')=x, \;J(y')=y$. Thus,
$\<J(x'), J(y')\> \in \PT_J(J(domain))$. Additionally, there are $z',w' \in V_O$ such that $J(z')=z, \;J(w')=w$. Thus,
$\<J(z'), J(w')\> \in \PT_J(J(x'))$. Then,
$\<I(x'), I(y')\> \in \PT_I(I(domain))$ and
$\<I(z'), I(w')\> \in \PT_I(I(x'))$. Since $I$ is an ERDF interpretation,
$\<I(z'), I(y')\> \in $\\
$\PT_I(I(type))$. Thus,
$\<J(z'), J(y')\> \in \PT_J(J(type))$ and $z \in \CT_J(y)$.

\vspace{0.1cm}
In a similar manner, we can prove that $J$ also satisfies the rest of the semantic conditions of Definition \ref{def:ERDFInterpretation}. Thus, $J$ is an ERDF interpretation of $V_O$.

Moreover, we will show that $J$ is a coherent ERDF interpretation (Definition \ref{def:CoherentERDFinterpretation}).
Assume that
this is not the case. Thus, there is $z \in Prop_J$ s.t. $\PT_J(z) \cap \PF_J(z) \not = \emptyset$.
Thus, there are $x,y \in Res_J$ s.t. $\<x,y\> \in \PT_J(z) \cap \PF_J(z)$, for such a $z$.
Then, there are $x',y',z' \in V_O$ s.t. $J(x')=x$, $J(y')=y$, and $J(z')=z$.
It holds: $\<J(x'), J(y') \> \in \PT_J(J(z'))$ and $\<J(x'), J(y') \> \in \PF_J(J(z'))$.
Thus, $\<I(x'), I(y') \> \in \PT_I(I(z'))$ and $\<I(x'), I(y') \> \in \PF_I(I(z'))$.
But this is impossible, since $I$ is a (coherent) ERDF interpretation.
Therefore, $J$ is also a coherent ERDF interpretation.

Thus, $J \in \I^H(O)$.

\vspace{0.1cm}
We will now show that $J \models sk(G)$. Let $p(s,o) \in sk(G)$. It holds $p,s,o \in V_O$.
Since $I \models sk(G)$, it holds $I(p) \in Prop_I$. Thus,
$\<I(p), I(\mathit{Property})\> \in \PT_I(I(type))$, which implies that $\<J(p), J(\mathit{Property})\> \in \PT_J(J(type))$. From this, it follows that $J(p) \in Prop_J$.
It holds: $\<J(s), J(o)\> \in \PT_J(J(p))$ iff
$\<I(s), I(o)\> \in \PT_I(I(p))$.
The last statement is true since $I \models sk(G)$. Let $u: \{\} \rightarrow  Res^H_O$.
Then, $J,u \models p(s,o)$.  
Let $\sneg p(s,o) \in sk(G)$. We can show that $J,u \models \sneg p(s,o)$, in a similar manner.  
Thus, $J \models sk(G)$. 

Now, from Definition \ref{def:stableModel} (Stable Model) and the fact that $J \models sk(G)$,
it follows that $\exists K \in \M^{st}(O)$ s.t. $K \leq J$. From this and 
the fact that $O \models^{st} F$, it follows that $K \models F$.
Since $F$ is an ERDF {\em d}-formula, it holds that 
\begin{center}
$F= (\exists ?x_1,..., \exists ?x_{k_1} \; F_1) \; \OR \; ... \; \OR \; (\exists ?x_1,...,  \exists ?x_{k_n} \; F_n)$, 
\end{center}
where  $F_i=t_1 \; \AND \; ... \; \AND \; t_{m_i}$ and $t_j$, for $j=1, ..., m_i$, is an ERDF triple.
Thus, there is an $i \in \{1, ..., n\}$ and $u:\Var(F_i) \rightarrow  Res^H_O$ s.t. $K,u \models F_i$.

We will show that $J,u \models F_i$. \\
Let $p(s,o) \in \{t_1,...,t_{m_i}\}$.
Since $K$ is an ERDF interpretation of $V_O, \; K,u \models F_i$, and
$Prop_K \subseteq Prop_J$, it follows that $p \in V_O$, $s,o \in V_O \cup \Var$,
and $J(p)=K(p) \in Prop_K \subseteq Prop_J$. Additionally, $\<[K+u](s), [K+u](o)\> \in \PT_K(p)$.
Since $\<[J+u](s), [J+u](o)\>=\<[K+u](s), [K+u](o)\>$ and $\PT_K(p) \subseteq \PT_J(p)$,
it follows that $\<[J+u](s), [J+u](o)\> \in \PT_J(p)$. Thus, $J,u \models p(s,o)$.\\
Let $\sneg p(s,o) \in \{t_1,...,t_{m_i}\}$.
Since $K$ is an ERDF interpretation of $V_O, \; K,u \models F_i$, and
$Prop_K \subseteq Prop_J$, it follows that $p \in V_O$, $s,o \in V_O \cup \Var$,
and $J(p)=K(p) \in Prop_K \subseteq Prop_J$. Additionally, $\<[K+u](s), [K+u](o)\> \in \PF_K(p)$.
Since $\<[J+u](s), [J+u](o)\>=\<[K+u](s), [K+u](o)\>$ and $\PF_K(p) \subseteq \PF_J(p)$,
it follows that $\<[J+u](s), [J+u](o)\> \in \PF_J(p)$. Thus, $J,u \models \neg p(s,o)$.

We  now define a total function $u':V_{F_i} \cup \Var(F_i) \rightarrow V_O$, as follows:
\[ u'(x)= \left\{ \begin{array}{ll}
u(x) & \mbox{if $x \in \Var(F_i)$ and }\\
 &   \mbox{$\;\;\;u(x)$ is not the xml value of a well-typed XML literal in $V_O$}\\
t   & \mbox{if $x \in \Var(F_i)$ and }\\
    & \mbox{$\;\;\;u(x)$ is the xml value of a well-typed XML literal $t$ in $V_O$}\\
x & \mbox{otherwise}
\end{array} \right. \]

Moreover, we define a total function
$u'': \Var(F_i) \rightarrow Res_I$ s.t.
$u''(x)=I(u'(x))$.

We will show that $I, u'' \models F_i$.

\noindent
Let $p(s,o) \in \{t_1,..., t_{m_i}\}$.
Then, $p \in V_{F_i}$ and $s,o \in V_{F_i} \cup \Var$. Since $J,u \models F_i$, it follows that
$V_{F_i} \subseteq V_O$. Therefore,
$V_{F_i} \subseteq V_{sk(G)} \cup \V_{RDF} \cup \V_{RDFS} \cup \V_{ERDF} \subseteq V'$.
Thus, $p \in V'$ and $s,o \in V' \cup \Var$.

We will now show that $I(p) \in Prop_I$. It holds:\\
$\<I(p), I(\mathit{Property})\> \in \PT_I(I(type))$ iff\\
$\<J(p), J(\mathit{Property})\> \in \PT_J(J(type))$, which holds since $J,u \models F_i$.

We want to show that $\<[I+ v''](s), [I+ v''](o)\> \in \PT_I(I(p))$.
Note that $\forall x \in V_{F_i}$, it holds: $[I+u''](x)=I(u'(x))=I(x)$ and $J(u'(x))=[J+u](x)=J(x)$.
Moreover, $\forall x \in \Var(F_i)$, it holds: $[I+u''](x)=I(u'(x))$ and $J(u'(x))=[J+u](x)$  (recall the definition of $J(.)$).
Therefore, it holds:\\
$\<[I + u''](s), [I + u''](o)\> \in \PT_I(I(p))$ iff \\
$\<I(u'(s)), I(u'(o))\> \in \PT_I(I(p))$ iff\\
$\<J(u'(s)), J(u'(o))\> \in \PT_J(J(p))$ iff\\
$\<[J+u](s), [J+u](o)\> \in \PT_J(J(p))$, which is true since $J,u \models F_i$.
Thus, $I,u'' \models p(s,o)$.

\noindent
Let $\neg p(s,o) \in \{t_1,..., t_{m_i}\}$. We can show that $I,u'' \models \neg p(s,o)$, in a similar manner.\\
Thus,   $I, u'' \models F_i$, which implies that $I, u'' \models \exists \; ?x_1, ..., ?x_{k_i}\; F_i$.
Thus, $I, u'' \models F$. Now, it follows from
Lemma B.1  that $I \models F$.

Thus, $sk(G) \models^{ERDF} F$. Now, it follows from Proposition \ref{ERDFentailmentEquivalence1} that $G \models^{ERDF} F$. \\

\vspace{0.1cm}
\noindent
2) Let $G \models^{ERDF} F$. It follows from Proposition \ref{ERDFentailmentEquivalence1} that $sk(G) \models^{ERDF} F$. We will show that $\<G,\emptyset\> \models^{st} F$.
In particular, let $O=\<G, \emptyset\>$ and let $I \in \M^{st}(O)$.
Note that $I$ is an ERDF interpretation of $V_O$, such that $I \models sk(G)$.
Since $sk(G) \models^{ERDF} F$, it follows that $I \models F$. $\Box$

\godown
\noindent
{\bf Proposition \ref{Prop:TilingReduction}}
Let $\D$ be an  instance of the unbounded tiling problem. It holds:
\begin{enumerate}

\item $\D$ has a solution iff $O_{\D} \cup \{\mathit{false} \; \leftarrow \; F_{\D}\}$ has a stable model.
\item $\D$ has a solution iff $O_{\D} \not \models^{st} F_{\D}$.
\end{enumerate}

\noindent
{\bf Proof:} 

\noindent
1) This statement follows easily from statement 2). \\

\noindent
2)
$\Rightarrow$) Let $\tau$ be a solution to $\D$. Since $\NN \times \NN$ is denumerable, there exists a bijective
function $\pi: \;\NN \times \NN \rightarrow \NN$. Consider now a Herbrand interpretation $I$ of $O_{\D}$ such that:
{\small
\begin{enumerate}

\item $CT_I(\mathit{Tile})= CT_I(\mathit{HasRight})=CT_I(\mathit{HasAbove})=\{ \rdf\?\_i \;|\; i \in \NN \}$ and \\
$CF_I(Tile)= CF_I(\mathit{HasRight})=CF_I(\mathit{HasAbove})=\emptyset$.

\item $PT_I(\mathit{id})= \{ \<x, x\> \;|\; x \in V_O\}$ and $PF_I(\mathit{id})= \emptyset$.

\item $PT_I(\mathit{HConstraint})= H$ and $PF_I(\mathit{HConstraint})= \emptyset$.

\item $PT_I(\mathit{VConstraint})= V$ and $PF_I(\mathit{VConstraint})= \emptyset$.

\item $PT_I(\mathit{Type})= \{ \<\rdf\?\_\pi(i,j), \tau(i,j) \> \;|\; i,j \in \NN \}$ and $PF_I(\mathit{Type})= \emptyset$.

\item $PT_I(\mathit{right})= \{ \<\rdf\?\_\pi(i,j), \rdf\?\_\pi(i+1,j) \> \;|\; i,j \in \NN \}$ and\\ 
$PF_I(\mathit{right})= \{\<\rdf\?\_i, \rdf\?\_j\> \;|\; i, j \in \NN$ and $\<\rdf\?\_i, \rdf\?\_j\> \not \in PT_I(\mathit{right})\}$.

\item $PT_I(\mathit{above})= \{ \<\rdf\?\_\pi(i,j), \rdf\?\_\pi(i,j+1) \> \;|\; i,j \in \NN \}$ and\\ 
$PF_I(\mathit{above})= \{\<\rdf\?\_i, \rdf\?\_j\> \;|\; i, j \in \NN$ and $\<\rdf\?\_i, \rdf\?\_j\> \not \in PT_I(\mathit{above})\}$.
\end{enumerate}
}

It is easy to see that $I$ is a stable model of $O_{\D}$ and $I \not \models F_{\D}$.
Thus, $O_{\D} \not \models^{st} F_{\D}$.\\

\noindent
$\Leftarrow)$ Let $\D=\<\T, H, V\>$, where $\T=\{T_1, ..., T_n\}$. Assume that $O_{\D} \not \models^{st} F_{\D}$ and let 
 $I$ be a stable model of $O_{\D}=\<G,P\>$ such that $I \not \models F_{\D}$.
 Obviously, $CT_I(\mathit{Tile})= \{ \rdf\?\_i \;|\; i \in \NN \}$. Due to rule sets (2)-(4) of $P$
 and since $O_{\D} \not \models^{st} F_{\D}$, it holds that 
 starting from tile $\rdf\?\_0$  and placing tiles according to 
$PT_I(\mathit{right})$ and $PT_I(\mathit{above})$ relations, 
 a grid is formed. We define $\pi(i,j)=k$, for $i,j,k \in \NN$, iff the tile $\rdf\?\_k$ has been placed 
 on the $\<i,j\>$ position of the previous grid. Note that $\pi$ is a total function.
 Due to rule set (1) of $P$, each tile is assigned a unique type in $\T=\{T_1, ..., T_n\}$.
 Due to rule set (5) of $P$, this type assignment satisfies the horizontal and vertical adjacency constraints of $\D$.
 Thus, a solution of $\D$ is $\tau: \NN \times \NN \rightarrow \T$, where
 $\tau(i,j)=T$ iff $\<\rdf\?\_\pi(i,j), T\> \in PT_I(Type)$. Since $\pi$ is a total function and, for all $k \in \NN$, 
 tile $\rdf\?\_k$ is assigned a unique type in $\T$, it follows that $\tau$ is a total function.
 
 \noindent

\vskip 0.2in
\bibliography{ERDF_JAIR}
\bibliographystyle{theapa}

\end{document}